\documentclass[12pt,a4paper]{book}
\usepackage{lmodern}

\usepackage[T5]{fontenc}
\usepackage[utf8]{inputenc}
\usepackage[english]{babel}
\usepackage{amsmath}
\usepackage{amssymb}

\usepackage{fancyhdr}
\usepackage[most]{tcolorbox}
\usepackage{tikz}

\definecolor{dykblue}{HTML}{E6F5FF}   
\definecolor{dykframe}{HTML}{3A7BD5}  
\definecolor{dykyellow}{HTML}{FFC857} 
\definecolor{dyktext}{HTML}{1F2933}   

\newtcolorbox{didyouknow}{
  enhanced,
  breakable,
  colback=dykblue,
  colframe=dykframe,
  coltext=dyktext,
  boxrule=0.9pt,
  rounded corners, arc=3mm,
  left=7mm,right=7mm,
  top=6mm,bottom=6mm,
  before skip=\medskipamount,
  after skip=\medskipamount,
  title={\sffamily\bfseries Did you know?},
  coltitle=dyktext,
  fonttitle=\small\sffamily\bfseries,
  attach boxed title to top left={
    yshift=-3mm,
    xshift=4mm
  },
  boxed title style={
    enhanced,
    colback=white,
    colframe=dykframe,
    boxrule=0.7pt,
    rounded corners, arc=2mm,
    top=1mm,bottom=1mm,left=3mm,right=3mm,
  }
}

\definecolor{corebg}{RGB}{245,245,255}   
\definecolor{coreframe}{RGB}{70,70,160}  
\definecolor{coretext}{RGB}{20,20,60}

\newtcolorbox{coreidea}[1][]{
  enhanced,
  breakable,
  colback=corebg,
  colframe=coreframe,
  coltext=coretext,
  boxrule=0.9pt,
  rounded corners, arc=2mm,
  left=7mm,right=7mm,
  top=6mm,bottom=6mm,
  before skip=\medskipamount,
  after skip=\medskipamount,
  title={\sffamily\bfseries Core idea},
  coltitle=corebg,
  fonttitle=\small\sffamily\bfseries,
  attach boxed title to top left={
    yshift=-3mm,
    xshift=4mm
  },
  boxed title style={
    enhanced,
    colback=coreframe,
    colframe=coreframe,
    boxrule=0pt,
    rounded corners, arc=1.6mm,
    top=1.2mm,bottom=1.2mm,
    left=4mm,right=4mm,
  },
  #1
}

\usepackage{tgpagella}
\newcommand{\TitleFont}{\ttfamily}

\usepackage{geometry}
\usepackage{float}
\usepackage{subcaption}
\usepackage{hyperref}
\hypersetup{
    colorlinks=true,
    linkcolor=blue,
    urlcolor=blue
}

\usepackage{xcolor}
\usepackage[numbers]{natbib}
\usetikzlibrary{calc}       

\definecolor{bgdark}{HTML}{050816}    
\definecolor{accentA}{HTML}{2563EB}   
\definecolor{accentB}{HTML}{7C3AED}   
\definecolor{accentC}{HTML}{22C55E}   
\definecolor{textlight}{HTML}{B0B3B8}   
\definecolor{bgA}{HTML}{1A202C}   

\fancypagestyle{plain}{%
  \fancyhf{}%
  \fancyhead[RO]{\makebox[\headwidth][r]{\nouppercase{\rightmark}}}%
  \fancyhead[LE]{\makebox[\headwidth][l]{\nouppercase{\leftmark}}}%
  \fancyfoot[C]{\thepage}%
}

\begin{document}

\begin{titlepage}
  \thispagestyle{empty}

  \begin{tikzpicture}[remember picture,overlay]

    \fill[bgdark]
      (current page.south west) rectangle (current page.north east);

    \fill[accentB!80!black,opacity=0.85]
      ($(current page.south west)+(-1cm,1cm)$) --
      ($(current page.north west)+(-1cm,4cm)$) --
      ($(current page.north east)+(-2cm,2cm)$) --
      ($(current page.south east)+(-2cm,-1cm)$) -- cycle;

    \fill[accentA!80!black,opacity=0.75]
      ($(current page.south west)+(0cm,0cm)$) --
      ($(current page.north west)+(0cm,3cm)$) --
      ($(current page.north east)+(-3cm,1cm)$) --
      ($(current page.south east)+(-3cm,-2cm)$) -- cycle;

    \fill[accentC!70!black,opacity=0.6]
      ($(current page.south west)+(3cm,-0.5cm)$) --
      ($(current page.north west)+(4cm,2cm)$) --
      ($(current page.north east)+(1cm,0cm)$) --
      ($(current page.south east)+(0cm,-2cm)$) -- cycle;


    \draw[accentA!60, line width=0.8pt]
      ($(current page.east)+(-3cm,0.5cm)$) circle (2.1cm);
    \draw[accentB!60, line width=0.6pt]
      ($(current page.east)+(-3cm,0.5cm)$) circle (1.4cm);
    \fill[accentC!80!white,opacity=0.9]
      ($(current page.east)+(-3cm,0.5cm)$) circle (0.12cm);

    \fill[white,opacity=0.9]
      ($(current page.north east)+(-4.2cm,-3.0cm)$) circle (0.06cm);
    \fill[white,opacity=0.7]
      ($(current page.north east)+(-2.6cm,-4.2cm)$) circle (0.06cm);
    \fill[white,opacity=0.6]
      ($(current page.north east)+(-5.0cm,-5.0cm)$) circle (0.06cm);

    \node[anchor=west]
      at ($(current page.west)+(1.8cm,5.5cm)$) {%
        \parbox{0.8 \textwidth}{%
          \color{white}
          {\Large\bfseries From Pixels to Prompts}\\[0.8em]
          {\fontsize{24}{28}\selectfont\bfseries Vision-Language Models}\\[0.8em]
          {\normalsize\color{white!80}
            Foundations, Architectures, and Applications of\\
            Multimodal Intelligence
          }
        }
      };

    \draw[white!40, line width=0.4pt]
      ($(current page.west)+(1.8cm,3.6cm)$) --
      ($(current page.west)+(12cm,3.6cm)$);

    \node[anchor=west]
      at ($(current page.west)+(1.8cm,3.1cm)$) {%
        \small\color{white!80}
        VISION \; $\times$ \; LANGUAGE
      };

\coordinate (safeNW) at ($(current page.north west)+(2cm,-2cm)$);
\node[anchor=west,
      rounded corners=3pt,
      fill=bgA!82!black,      
      draw=textlight!35,      
      line width=0.5pt,
      inner sep=8pt,
      minimum width=0.60\paperwidth] 
  at ($(safeNW)+(0,-220mm)$)
  {\parbox[t]{0.56\paperwidth}{%
    {\TitleFont\large\bfseries\color{textlight} VO HOANG NHAT KHANG}\\[3pt]
    {\TitleFont\small\color{textlight!85}
      Ph.D.\ Student in Natural Language Processing @MBZUAI}\\[3pt]
    {\TitleFont\small\itshape\color{textlight!60}
      Bridging pixels and language for truly multimodal intelligence.}%
  }};

    \node[anchor=east]
      at ($(current page.east)+(-1.8cm,-13cm)$) {%
        \color{white!70}\scriptsize
        Version 0.1 \quad | \quad \today
      };

  \end{tikzpicture}
\end{titlepage}


\frontmatter


\chapter*{Preface}
\thispagestyle{empty}
\addcontentsline{toc}{chapter}{Preface}

\small 
When you read a paper about a new Vision-Language Model today, it can be
easy to forget how strange this idea would have sounded not so long ago.
Teaching machines to see was already hard. Teaching them to read and
generate language was already hard. Asking them to do both at once - and
then to reason, answer questions, follow instructions, and sometimes
even surprise us - still carries a quiet trace of science fiction, even
as it becomes routine.

This book was born from a simple feeling: \emph{it is too easy to get
lost}. The field moves quickly, new model names appear constantly, and
the gap between ``I know the buzzwords'' and ``I actually understand how
this works'' can feel uncomfortably wide. I have felt that gap many
times. If you are holding this book, you probably have too.

My goal is not to provide an exhaustive catalog of every dataset,
benchmark, and new model variant. Instead, I want to offer something
more modest - and, I hope, more durable: a clear mental map of
Vision-Language Models. Enough structure that you can read new papers
with confidence; enough intuition that you can design your own systems
without feeling as if you are assembling LEGO bricks blindly.

There is also a very specific origin behind this project. Much of the
motivation came from the course \emph{Vision-to-Language Generation} that
I took in my first semester in which my supervisor is the lecturer of the course. The lectures,
discussions, and reading list shaped the way I think about this area:
not as a parade of model names, but as a set of recurring design
patterns, trade-offs, and evaluation habits. In that sense, this book is
also a small gift to her: a way to record what I have learned so far, to
organize it into a coherent map, and to carry those lessons forward as I
continue my research.

This book is for students and practitioners who already care about
machine learning, and who are curious about the multimodal frontier. You
might be a student who has seen CNNs and Transformers in class and now
wants to understand how they are brought together in VLMs. You might be
a researcher or engineer who keeps encountering CLIP, Flamingo, BLIP-2,
or ``multimodal LLMs'' and wants a principled way to think about them.
Or you might be someone building practical systems - search, document
understanding, assistive tools, robotics - and wondering what
Vision-Language Models can realistically do for your domain.

You do \emph{not} need to know every detail of every model to benefit
from this book. If you understand basic deep learning, are comfortable
with the idea of an encoder, a decoder, and a loss function, and you are
willing to pause occasionally to digest an equation, you have enough
background to start.

You also do not have to read this book linearly, cover to cover, to get
value from it. In fact, I expect most readers will jump around. If you
want the big picture and motivation, start with
Chapter~\ref{chap:intro-vlm}. If you want the visual side, go to the
Chapter~\ref{chap:visual-encoders} on visual encoders. If your curiosity is about prompting,
instruction following, and how large language models are extended to
images, focus on the Chapter \ref{chap:lms-for-vlms} on language models and multimodal
alignment. If you want to know how Vision-Language Models are trained, as well as what kinds of architecture are existing out there, refer to Chapter \ref{chap:arch-train}. If you are a practitioner looking for guidance on evaluation
and deployment, the later Chapters \ref{chap:datasets-benchmarks} and \ref{chap:applications-future} on datasets, benchmarks, and
applications will be most relevant.

You are encouraged to read actively: sketch architectures, rewrite
equations in your own notation, and compare the descriptions here with
the models you encounter in papers and code. This book is not a
replacement for reading original work; it is meant to make that reading
less intimidating and more rewarding.

It is also important to be honest about the limits of this text. This
book will not always be up to date with the very latest model names. The
field moves faster than any static document can. You will almost
certainly encounter architectures that are not mentioned here. That is
expected. The real test is not whether you have seen a model before, but
whether you can say: \emph{``Ah, this is basically a patch-based encoder
with a learned bridge into a language model''}, or \emph{``This uses
cross-attention and instruction tuning to behave like an assistant''}.

Finally, Vision-Language Models sit uncomfortably close to how we, as
humans, communicate and interpret the world. They caption personal
photos, summarize news images, read charts, inspect medical scans, and
generate text that can easily be mistaken for something a person wrote.
This is powerful, and it is also risky. Understanding how these systems
are built - where their strengths come from, where their blind spots are,
and how biases in the data surface in their behavior - is not just an
academic exercise. It is a form of responsibility.

I am grateful to the people around me who made this work feel possible.
To my friends working in the AI industry: thank you for the practical perspective, the late-night conversations, and the reminders that elegant ideas only matter when they survive contact with real users and real constraints. To my colleagues: thank you for the discussions, paper-sharing, and the quiet encouragement that comes from learning together. And to my family: thank you for the patience, the trust, and the kind of support that does not always look like research, but makes research possible. I am also grateful to Mr. Khang Hoang Phan and Mr. Khang Thanh Doan for their constructive feedback and thoughtful suggestions.

To be honest, writing this book forced me to slow down in a field that
does not like to slow down. It meant reading papers more carefully than
is strictly necessary, admitting when I did not really understand a
detail, and explaining ideas in a way that my past self would have
appreciated. If, at some point while reading, you feel a concept
``click'' in your mind - a model that used to be a blur finally becomes a
simple picture, an equation suddenly makes intuitive sense, or you
realize you could explain an idea to a colleague - then this book has
done its job.

Thank you for giving it a place in your own learning journey.

\vspace{2em}

\begin{flushright}
  VO HOANG NHAT KHANG\\
  Ph.D. Student in Computing \& Mathematical Sciences Division @MBZUAI, UAE\\[0.5em]
\end{flushright}

\normalsize

\tableofcontents
\cleardoublepage

\mainmatter

\chapter{Introduction to Vision-Language Models}
\label{chap:intro-vlm}

\section{Why Vision-Language Models?}

Most of the information humans perceive is not purely textual. We read
books, but we also watch videos, look at photographs, navigate through
maps, and interpret diagrams. In our everyday reasoning, visual and
linguistic information are tightly coupled: we describe what we see,
imagine what is described, and ground abstract concepts in concrete
images.

From a probabilistic point of view, these modalities are also not
independent. When you look at a scene, there is some underlying
\emph{world state} - objects, relations, intentions, dynamics - that we
can call $Z$. Images, spoken descriptions, captions, and sounds are
different \emph{projections} of this same latent cause:
\[
  x^{\text{image}},\; x^{\text{text}},\; x^{\text{audio}}
  \text{ all arise from the same } Z.
\]
Humans are remarkably good at moving between these projections: from an
image we can imagine a sentence; from a sentence we can sketch a scene.

Traditional machine learning systems, however, have largely treated
\emph{vision} and \emph{language} as separate problems. Computer vision
models take images or videos as input and output labels, bounding boxes
or segmentation masks. Natural language processing (NLP) models take
text as input and output text or discrete labels. Each side has made
rapid progress, but their capabilities remained mostly isolated.

Vision-Language Models (VLMs) represent an important step towards
bridging this gap. They are designed to \emph{jointly} process visual
and textual information so that models can:
\begin{itemize}
  \item describe images and videos in natural language,
  \item answer questions about visual content,
  \item retrieve images that match a textual description (and vice versa),
  \item follow natural language instructions that refer to visual scenes,
  \item and, more broadly, reason over multimodal inputs.
\end{itemize}

At a very high level, almost every model in this book is doing three
things, over and over again:

\begin{enumerate}
  \item \textbf{Encode} raw signals (pixels, tokens) into internal
        vectors.
  \item \textbf{Reason} over those vectors as an abstract
        representation of the world state $Z$.
  \item \textbf{Decode} that representation into the desired output
        (a caption, an answer, a bounding box, a JSON structure).
\end{enumerate}

All of the architectures, losses, and training tricks you will see later
can be viewed as different ways of improving one of these three steps.

\begin{coreidea}
A Vision-Language Model is a machine that
learns and reasons over a \emph{shared hidden representation of the
world}. Images, text, and other signals are just different projections
of that latent world state $Z$. Encoders turn each modality into a
representation of $Z$; decoders read from $Z$ to produce the outputs we
care about.
\end{coreidea}

\begin{didyouknow}
Many everyday ``vision-language'' tasks that feel trivial to humans,
such as describing a photo to a friend or following verbal directions
on a map, actually involve several subproblems at once: detecting
objects, understanding spatial relations, tracking events over time,
and mapping all of this to coherent language. VLMs are interesting not
only because they combine vision and text, but because they force us to
tackle these intertwined abilities in a single model.
\end{didyouknow}

\section{What is a Vision-Language Model?}

\paragraph{Definition.}
Intuitively, a \textbf{\textcolor{blue!85}{Vision-Language Model}} is a parameterized function that
takes a combination of visual and textual inputs and produces one or
more outputs, often in the form of text, scores, or structured
predictions. At a high level, we can write:
\begin{equation}
  f_\theta : (\mathcal{X}_{\text{vision}}, \mathcal{X}_{\text{text}}) \rightarrow \mathcal{Y},
\end{equation}
where $\mathcal{X}_{\text{vision}}$ is a space of visual inputs
(e.g.\ images, video frames), $\mathcal{X}_{\text{text}}$ is a space of
text (e.g.\ prompts, questions, instructions), and $\mathcal{Y}$ is a
task-specific output space (e.g.\ natural language, class labels,
retrieval scores).

In practice, modern VLMs follow a modular architecture with three
conceptual components, which mirror the \emph{encode $\rightarrow$
reason $\rightarrow$ decode} story above:

\begin{enumerate}
  \item \textbf{A visual encoder} that maps images (or video frames)
        into a sequence of visual embeddings: a compressed view of the
        world state $Z$ as seen through pixels.
  \item \textbf{A language model} that encodes and generates text, and
        serves as the main ``reasoning engine'' over these embeddings.
  \item \textbf{A fusion or alignment mechanism} that lets the language
        model \emph{attend to} or \emph{condition on} visual embeddings,
        so that information from different modalities interacts in a
        shared latent space.
\end{enumerate}

Different families of models implement these components in different
ways, but this decomposition will reappear throughout the book and is a
useful lens for understanding the design choices across the field.

\paragraph{Modalities and representations.}
Let us denote an image by $I \in \mathbb{R}^{H \times W \times C}$,
where $H$ and $W$ are height and width, and $C$ is the number of color
channels. A textual sequence is typically represented as a series of
tokens:
\begin{equation}
  \mathbf{t} = (t_1, t_2, \dots, t_n),
\end{equation}
where each token $t_i$ belongs to a finite vocabulary
$\mathcal{V}_{\text{text}}$.

A visual encoder transforms $I$ into a sequence of continuous
embeddings:
\begin{equation}
  \mathbf{v} = (v_1, v_2, \dots, v_m), \quad v_j \in \mathbb{R}^d,
\end{equation}
while a text encoder or language model maps $\mathbf{t}$ into
embeddings
\begin{equation}
  \mathbf{e} = (e_1, e_2, \dots, e_n), \quad e_i \in \mathbb{R}^d.
\end{equation}
You can think of $\mathbf{v}$ and $\mathbf{e}$ as two different ways of
encoding the same underlying world state $Z$ into vectors. The fusion
mechanism then combines $\mathbf{v}$ and $\mathbf{e}$ in a
task-dependent way-typically through cross-attention layers inside a
transformer-based language model-and a decoder head maps the final
representation back into $\mathcal{Y}$: a sentence, an answer, a set of
coordinates, or some other structured output.

\section{Historical Context and Evolution}

The idea of combining vision and language is not new. Early work in
\emph{image captioning} trained models to generate short descriptions
from images, typically using convolutional neural networks (CNNs) for
vision and recurrent neural networks (RNNs) for language. Around the
same time, \emph{visual question answering} (VQA) emerged as a benchmark
for multimodal reasoning, where models answer natural language
questions given an image.

These early architectures were often bespoke: they were carefully
engineered for each task and trained on relatively small datasets. In
contrast, modern VLMs are strongly influenced by two major trends:

\begin{itemize}
  \item \textbf{Large-scale pretraining.} Leveraging millions or
        billions of image-text pairs scraped from the web, models learn
        generic multimodal representations that can be adapted to many
        downstream tasks.
  \item \textbf{Transformer-based language models.} Large language
        models (LLMs) provide powerful sequence modeling and
        generation capabilities, making them attractive backbones for
        multimodal systems.
\end{itemize}

A key development is the use of \emph{contrastive pretraining} for
aligning image and text embeddings in a shared space, enabling robust
zero-shot recognition and retrieval. Another is the emergence of
\emph{instruction-tuned} VLMs that can follow natural language commands
about images and act as general-purpose assistants.

\section{Core Building Blocks}

We will unpack the three conceptual components introduced
earlier: visual encoders, language models, and fusion mechanisms.

\subsection{Visual Encoders}

Modern visual encoders are typically based on either convolutional
architectures or vision transformers. Regardless of the specific design,
their goal is to transform pixel-level inputs into a sequence of
semantic features that can be consumed by a downstream
vision-language architecture.

Formally, an input image can be represented as
$I \in \mathbb{R}^{H \times W \times C}$, where $H$ and $W$ denote
height and width, and $C$ is the number of color channels. A visual
encoder maps $I$ to a sequence of feature vectors
\[
  \mathbf{v} = (v_1, v_2, \dots, v_m), \qquad v_j \in \mathbb{R}^d,
\]
which we will refer to as \emph{visual tokens}. Different encoder
families correspond to different ways of defining these tokens.

\paragraph{Patch-based encoders.}

Patch-based encoders treat an image as an ordered sequence of
non-overlapping patches instead of a single dense grid of pixels. This
design became popular with Vision Transformers (ViT)~\cite{dosovitskiy2021image}
and CLIP-style image encoders~\cite{radford2021learning}. The image is
first divided into patches of fixed spatial size (e.g., $16 \times 16$
pixels). Each patch is then flattened and linearly projected into a
vector embedding, producing an initial sequence
\[
  \tilde{\mathbf{v}} = (\tilde{v}_1, \tilde{v}_2, \dots, \tilde{v}_m),
  \qquad \tilde{v}_j \in \mathbb{R}^d.
\]
A transformer encoder processes this sequence using self-attention,
refining the patch embeddings into contextualized visual tokens
$v_1, \dots, v_m$ that capture both local details and global structure.

Patch-based encoders are attractive for VLMs because they produce a
dense, grid-like set of tokens that aligns naturally with the token
sequence used by language models. Moreover, the transformer
architecture allows the model to scale to large input resolutions and
large pretraining datasets without relying heavily on handcrafted
inductive biases commonly used in convolutional networks.

\paragraph{Region-based encoders.}

In many applications, however, it is useful to focus on semantically
meaningful regions, such as individual objects, text segments, or
salient parts of the scene. Region-based encoders first identify
candidate regions and then compute embeddings for each region. This
paradigm is exemplified by object detection architectures such as
Faster R-CNN \cite{ren2015faster} and Mask R-CNN \cite{he2017mask}, which have been widely adopted as visual
front-ends in early vision-language systems for Visual Question
Answering and grounding.

Region-based encoders provide a sparse set of high-level visual tokens,
each corresponding to a putative object or region. This can be
advantageous for tasks that require fine-grained grounding or reasoning
over discrete entities, but may miss global context compared to
patch-based encoders. Modern VLMs therefore often choose between
patch-based and region-based encoders depending on the target tasks and
computational budget.

\subsection{Language Models}
\label{subsec:language-models}

The language component in a VLM typically plays two roles:

\begin{enumerate}
  \item \textbf{Encoding}: mapping textual prompts, questions, or
        instructions into embeddings that can interact with visual
        tokens.
  \item \textbf{Generation}: producing natural language outputs
        conditioned on visual features and previous tokens.
\end{enumerate}

Modern VLMs almost always rely on transformer-based language
models~\citep{vaswani2017attention}. In particular, large
\emph{autoregressive} language models-such as GPT-3~\citep{brown2020language},
instruction-tuned variants~\citep{ouyang2022training}, or open models
like LLaMA~\citep{touvron2023llama}-serve as powerful backbones that
provide rich world knowledge, strong sequence modeling capabilities, and
flexible natural language generation.

Given a sequence of text tokens
$\mathbf{t} = (t_1, \dots, t_n)$ from a vocabulary
$\mathcal{V}_{\text{text}}$, an autoregressive language model defines a
probability distribution
\begin{equation}
  p(t_1, \dots, t_n)
  = \prod_{i=1}^n p(t_i \mid t_1, \dots, t_{i-1}),
\end{equation}
where each conditional factor $p(t_i \mid t_{<i})$ is parameterized by a
stack of self-attention and feed-forward layers. During generation, the
model repeatedly samples (or greedily selects) the next token from this
distribution, appending it to the context.

In a vision-language setting, this formulation is extended so that the
conditional distribution also depends on visual tokens
$\mathbf{v} = (v_1, \dots, v_m)$ produced by a visual encoder:
\begin{equation}
  p(t_1, \dots, t_n \mid \mathbf{v})
  = \prod_{i=1}^n p(t_i \mid t_1, \dots, t_{i-1}, \mathbf{v}),
\end{equation}
where the dependence on $\mathbf{v}$ is implemented via cross-attention
mechanisms or specialized adapter modules (discussed in the next
subsection). In this way, the language model acts as a general-purpose
reasoning and generation engine, while visual tokens provide grounded
context about the external world.

\subsection{Fusion and Alignment Mechanisms}
\label{subsec:fusion-alignment}

The core technical challenge in VLMs is to make visual information
available to the language model in a flexible, scalable way. Several
strategies have emerged, corresponding to different ways of aligning
visual and textual representations.

\paragraph{Joint embedding space.}

One approach is to learn separate encoders for images and text, and
align their outputs using a contrastive objective. This paradigm is
exemplified by CLIP-style models~\cite{radford2021learning} and
ALIGN~\cite{jia2021scaling}, which train image and text encoders so that
paired (image, caption) representations have high similarity, while
mismatched pairs are pushed apart. The resulting joint embedding space
supports a range of tasks such as image-text retrieval and zero-shot
classification, and often serves as a generic visual backbone for
downstream multimodal systems.

\paragraph{Cross-attention.}

Another approach is to feed visual embeddings directly into a
multi-layer transformer via cross-attention, allowing text tokens to
attend to visual tokens. Early vision-language transformers such as
ViLBERT~\cite{lu2019vilbert} and LXMERT~\cite{tan2019lxmert} encode
images and text with separate streams that interact through
cross-attention layers, enabling the model to ground linguistic
representations in visual context. This architecture is particularly
effective for tasks like image captioning, Visual Question Answering,
and referring expression grounding, where fine-grained interactions
between words and regions are essential.

\paragraph{Adapters and projection layers.}

When integrating a large pre-trained language model with a visual
encoder, the dimensionalities and representation formats often do not
match. Projection layers and lightweight adapters are used to bridge
this gap while preserving most of the original weights of the language
model. For example, Flamingo~\cite{alayrac2022flamingo} augments a
frozen language model with gated cross-attention layers that ingest
visual tokens, and BLIP-2~\cite{li2023blip2} introduces a small
``Q-former'' module that converts visual features into a compact set of
tokens aligned with a frozen LLM. These designs highlight a general
trend: instead of retraining the entire language model, VLMs increasingly
use modular adapters to inject visual information in a parameter-efficient
way.

\section{Canonical Tasks for VLMs}
\label{sec:canonical-tasks}

Vision-Language Models are evaluated and trained on a variety of
multimodal tasks that couple visual inputs with natural language. These
benchmarks not only provide quantitative measures of progress but also
shape the design of model architectures and training objectives.

\subsection{Image Captioning}
\label{subsec:image-captioning}

In image captioning, the goal is to generate a natural language
description of an input image. Early neural captioning systems combined
convolutional visual encoders with recurrent language decoders, and were
evaluated on datasets such as MS~COCO~\cite{lin2014microsoft} and
Flickr30k~\cite{karpathy2015deep}. Formally, given an image $I$ and its
visual representation $\mathbf{v}$, we wish to model
\begin{equation}
  p(\mathbf{t} \mid I)
  = p(t_1, \dots, t_n \mid I),
\end{equation}
where $\mathbf{t} = (t_1, \dots, t_n)$ is a sequence of caption tokens.
The model is trained to maximize the likelihood of reference captions,
often with additional objectives or decoding constraints to control
diversity and length~\cite{vinyals2015show}.

Captioning tasks typically evaluate the fluency, relevance, and
completeness of the generated text using automatic metrics such as
BLEU, METEOR, CIDEr, or SPICE, as well as human judgments. For VLMs,
image captioning serves both as a standalone application and as a
diagnostic task for measuring the quality of multimodal representations.

\subsection{Visual Question Answering}
\label{subsec:vqa}

Visual Question Answering (VQA) combines image understanding with
natural language reasoning. The model receives an image $I$ and a
question $\mathbf{q}$ (e.g., ``How many people are on the boat?''), and
must produce an answer $\mathbf{a}$, which may be a word, phrase, or
short sentence:
\begin{equation}
  p(\mathbf{a} \mid I, \mathbf{q}).
\end{equation}
The original VQA dataset~\cite{antol2015vqa} and its improved variant
VQA v2~\cite{goyal2017vqa2} were designed to probe different aspects of
multimodal understanding, including object recognition, counting,
spatial relationships, commonsense reasoning, and reading text in
images.

From the perspective of VLMs, VQA is a natural setting for evaluating
\emph{grounded} language understanding: the model must integrate visual
and textual evidence to reach a correct answer, rather than relying
solely on linguistic priors. Modern VLMs often treat VQA as a
specialized prompting or instruction-following scenario, where the
answer is generated by a language model conditioned on visual tokens.

\subsection{Image-Text Retrieval}
\label{subsec:image-text-retrieval}

In image-text retrieval tasks, the goal is to align images with
descriptive texts in a shared representation space. Given a text query,
the system must rank images by relevance (text-to-image retrieval), and
conversely, given an image query, it must retrieve appropriate captions
or descriptions (image-to-text retrieval). Datasets such as
MS~COCO~\cite{lin2014microsoft} and Flickr30k Entities~\cite{plummer2015flickr30k}
are commonly used to evaluate such models.

A standard approach is to embed images and texts into a joint space using
separate encoders and train them with a contrastive objective, as in
CLIP and related methods. The similarity between an image and a text,
for example via cosine similarity between mean-pooled embeddings, is
then used to rank candidates. Beyond retrieval, the learned embedding
space serves as a foundation for zero-shot classification and other
downstream multimodal tasks.

\subsection{Referring Expressions and Grounding}
\label{subsec:referring-grounding}

Grounding tasks focus on associating textual expressions with specific
regions or objects in an image. Given a phrase like “the small red cup
on the left”, the model must identify and highlight the corresponding
entity. Datasets such as ReferItGame~\cite{kazemzadeh2014referitgame}
and Flickr30k Entities~\cite{plummer2015flickr30k} provide large
collections of region-level annotations linked to natural language
phrases.

Formally, grounding can be modeled as selecting a region $r$ from a set
of candidates $\mathcal{R}$ given an image $I$ and an expression
$\mathbf{e}$:
\begin{equation}
  p(r \mid I, \mathbf{e}), \qquad r \in \mathcal{R}.
\end{equation}
These tasks require fine-grained alignment between linguistic
descriptions and spatial structure, and have been influential in the
design of region-based encoders and cross-modal attention mechanisms.
For VLMs, strong grounding performance is a key indicator that visual
tokens and language representations are truly interacting, rather than
merely co-existing in the same model.

\section{Why Vision-Language Models Matter}
\label{sec:why-vlms-matter}

Vision-Language Models are not merely another specialized model class.
They represent an important step towards more general multimodal
intelligence, in which information from different modalities can be
jointly represented, reasoned over, and used to support downstream
decision-making.

\begin{itemize}
  \item \textbf{Unified representations.}
        By learning shared or aligned representations for images and
        text, VLMs enable knowledge transfer across tasks and
        modalities. A model trained on large-scale image-text data can
        often be adapted to new tasks with relatively modest amounts of
        supervision, for example through fine-tuning, prompting, or
        lightweight adapters. This \emph{foundation model} role mirrors
        the impact of large language models in purely textual domains.

  \item \textbf{Natural, flexible interfaces.}
        Natural language offers an intuitive interface for specifying
        tasks, constraints, and preferences. VLMs allow users to query
        and direct systems using text (e.g., “find images like this”,
        “describe what is happening here”, “highlight all traffic signs
        in this frame”) while grounding responses in visual evidence.
        This reduces the need for task-specific engineering and enables
        non-experts to interact with complex visual pipelines.

  \item \textbf{Broad applicability across domains.}
        Pretrained VLMs serve as powerful backbones for a wide range of
        applications, including visual assistants, document and chart
        understanding, medical image analysis, robotics and
        embodied AI, multimodal search and recommendation, and content
        moderation. In many of these settings, the same underlying model
        can support multiple capabilities-captioning, retrieval,
        question answering, and grounding-through different prompts or
        heads.

  \item \textbf{A testbed for multimodal reasoning.}
        Because VLMs must integrate heterogeneous information, they
        provide a natural testbed for studying compositionality,
        grounding, and reasoning across modalities. Progress in this
        area often informs the design of more general multimodal
        systems, including those that incorporate audio, video, or
        structured data in addition to images and text.
\end{itemize}

At the same time, VLMs inherit and sometimes amplify many of the
challenges associated with large-scale foundation models: they require
substantial computation and data, they can encode social and cultural
biases present in their training corpora, and they may confidently
hallucinate plausible but incorrect statements or descriptions. These
risks are particularly salient when models are used in high-stakes
settings, or when generated text is presented as factual.

\section{Limitations and Open Challenges}
\label{sec:vlm-limitations}

Despite rapid progress, current Vision-Language Models still exhibit
significant limitations.

\begin{description}
  \item[Data quality and coverage.]
        Large-scale web data contain substantial noise, biases, and
        coverage gaps. Certain visual concepts, demographic groups,
        languages, and cultural contexts are underrepresented or
        misrepresented, leading to uneven performance and systematic
        errors. Cleaning and curating multimodal datasets at scale
        remains a major unresolved challenge.

  \item[Compositional generalization.]
        VLMs often struggle to systematically combine known concepts in
        novel ways. For example, they may fail at counting complex
        objects, reasoning about rare attribute combinations, or
        understanding hypothetical or counterfactual descriptions. Such
        failures reveal limitations in the underlying representations
        and training objectives, which tend to favor surface correlations
        over robust compositional structure.

  \item[Grounding and faithfulness.]
        Although VLMs are designed to ground language in visual inputs,
        their outputs do not always faithfully reflect the image
        content. Under ambiguous, adversarial, or underspecified prompts,
        models may hallucinate objects, attributes, or actions that are
        not present in the scene. Ensuring tight and verifiable grounding
        between generated text and visual evidence is an active area of
        research, with implications for safety and trustworthiness.

  \item[Interpretability and controllability.]
        Understanding \emph{why} a VLM produces a particular answer or
        description is challenging. Existing tools for visualizing
        attention patterns, attributing predictions to input regions or
        tokens, and probing internal representations provide only partial
        insight. Moreover, controlling model behavior-for example, to
        enforce style, avoid certain content, or respect safety
        constraints-remains imperfect and often requires additional
        alignment or post-processing steps.

  \item[Efficiency and accessibility.]
        State-of-the-art VLMs typically require large computational
        resources for pretraining and inference, which limits who can
        train, deploy, and scrutinize them. Developing more
        parameter-efficient architectures, distillation techniques, and
        hardware-friendly implementations is essential for making VLMs
        broadly accessible and environmentally sustainable.

\end{description}

Future work must balance the pursuit of stronger capabilities with a
careful treatment of robustness, fairness, privacy, and societal impact,
so that Vision-Language Models can be deployed in ways that are both
technically sound and socially responsible.

\section{Structure of This Book}

This book is organized to follow the modular view of Vision-Language Models introduced above, moving from basic components to full systems, data, and applications:

\begin{itemize}
  \item \textbf{Chapter~\ref{chap:intro-vlm}: Introduction to Vision-Language Models.}
        Defines what we mean by a Vision-Language Model, reviews the historical context, and introduces core building blocks and canonical tasks such as captioning, VQA, retrieval, and grounding. The chapter closes with a discussion of why VLMs matter, open challenges, and an outline of the rest of the book.

  \item \textbf{Chapter~\ref{chap:visual-encoders}: Visual Encoders and Image Understanding.}
        Surveys convolutional networks, vision transformers, and region-based encoders, together with multi-scale feature representations and pretraining objectives. This chapter focuses on how images are converted into token-like visual representations suitable for downstream multimodal models.

  \item \textbf{Chapter~\ref{chap:lms-for-vlms}: Language Models for Vision-Language Systems.}
        Covers neural language modeling from recurrent networks to transformers, and frames large autoregressive LMs as general-purpose reasoning engines. It then analyzes how these models can be conditioned on visual tokens via prefixing, cross-attention, and encoder-decoder architectures.

  \item \textbf{Chapter~\ref{chap:arch-train}: Architectural Design and Training Paradigms.}
        Examines representative VLM architectures in detail-including encoder-decoder captioners, BLIP/BLIP-2, Flamingo, LLaVA, Qwen-VL, and generalist multimodal models-and the multimodal alignment objectives and bridge modules that connect visual encoders to language backbones.

  \item \textbf{Chapter~\ref{chap:datasets-benchmarks}: Datasets \& Evaluation Benchmarks.}
        Reviews major sources of multimodal data for pretraining, task-specific supervised datasets, and evaluation benchmarks across captioning, VQA, retrieval, grounding, hallucination, and holistic multi-task testing. The chapter also highlights biases, robustness issues, and human- and safety-centered evaluation practices.

  \item \textbf{Chapter~\ref{chap:applications-future}: Applications of Vision-Language Models, and Future Directions.}
        Discusses real-world application domains for VLMs, design considerations for deployment (latency, reliability, privacy, governance), and open research directions around richer world models, grounded generation, data curation, and human-AI interaction. The book concludes with an afterword reflecting on the broader trajectory of multimodal AI.
\end{itemize}

Our goal is to provide a coherent narrative that connects fundamental
ideas with practical systems, enabling readers to understand not only
\emph{how} Vision-Language Models work, but also \emph{why} they are
designed in particular ways and \emph{what} challenges remain open.

\newpage
\chapter{Visual Encoders and Image Understanding}
\label{chap:visual-encoders}

\textit{In any Vision-Language Model, the visual encoder is the component that
turns raw pixels into a representation that other modules can reason
with. It is the first step in a long chain: from an image $I \in
\mathbb{R}^{H \times W \times C}$, through a neural network, to a set of visual tokens
\begin{equation}
  \mathbf{v} = (v_1, v_2, \dots, v_m), \qquad v_j \in \mathbb{R}^d,
\end{equation}
which are later consumed by a language model or a multimodal fusion
module.}

\section{From Pixels to Features: Convolutional Encoders}
\label{sec:conv-encoders}

Convolutional neural networks (CNNs) have been the workhorse of modern
computer vision since the success of models such as LeNet,
AlexNet, VGG, and ResNet~\cite{lecun1998gradient,krizhevsky2012imagenet,simonyan2015very,he2016deep}. Their central idea is to exploit the
spatial structure of images through local receptive fields, shared
weights, and hierarchical feature extraction.

\subsection{Convolution as Learned Feature Extraction}

Let $I \in \mathbb{R}^{H \times W \times C}$ denote an input image. A
convolutional layer applies a bank of filters
$K^{(1)}, \dots, K^{(F)}$, each of spatial size $k \times k$, across
the image, producing feature maps
$F^{(1)}, \dots, F^{(F)}$. For a single filter $K$ and position $(i,j)$,
the convolution output can be written as
\begin{equation}
  F(i,j) = \sigma\!\big( (K * I)_{i,j} + b \big),
\end{equation}
where $*$ denotes the discrete convolution, $b$ is a bias term, and
$\sigma$ is a nonlinearity such as ReLU. Stacking convolutional layers
with intermediate pooling operations yields progressively more abstract
features: early layers respond to edges and simple textures, while
deeper layers capture object parts and semantic patterns.

\begin{figure}[ht]
    \centering
    \includegraphics[width=\linewidth]{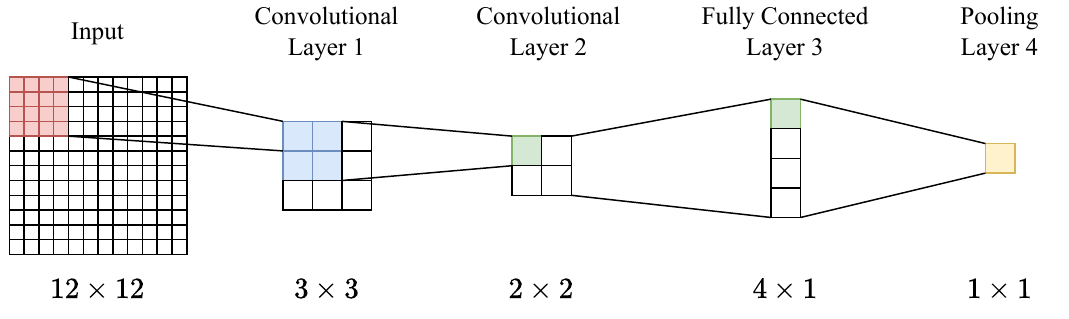}
    \caption{Schematic of a classical convolutional neural network used as a visual encoder. Starting from an input image, a stack of
    convolutional layers extracts local features, a pooling layer
    reduces spatial resolution while retaining salient activations, and
    one or more fully connected layers map the resulting feature vector
    to task-specific outputs such as class scores.}
    \label{fig:cnn-encoder}
\end{figure}

CNN-based visual encoders can be seen as functions
\begin{equation}
  E_{\text{conv}} : \mathbb{R}^{H \times W \times C} \rightarrow
  \mathbb{R}^{h \times w \times d},
\end{equation}
mapping an image to a grid of feature vectors, sometimes referred to as
a \emph{feature map}. The spatial resolution $(h,w)$ is typically much
smaller than $(H,W)$ due to downsampling, while $d$ is the number of
channels in the final layer.

\subsection{Deep Residual Networks and Modern Convnets}

\begin{figure}[ht]
    \centering
    \includegraphics[width=0.5\linewidth]{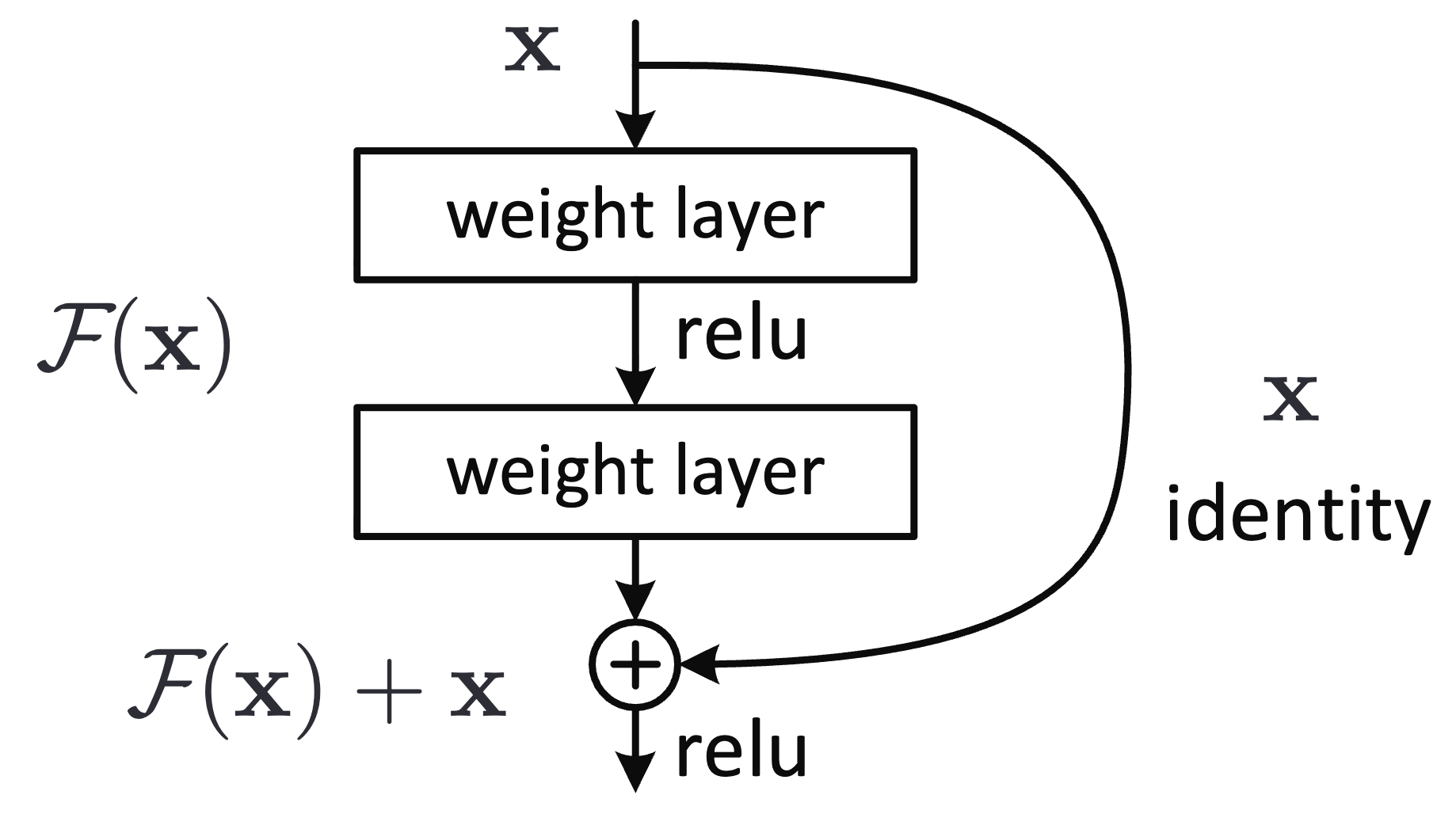}
    \caption{Illustration of a residual block in a deep residual network
    (ResNet)~\cite{he2016deep}. The input activation is propagated
    through a stack of convolution, normalization, and nonlinearity
    layers that implement the residual function $\mathcal{F}(x_{\text{in}})$,
    and is then added back to the original input via an identity (or
    projection) shortcut. This skip connection allows the network to
    learn residual mappings and greatly eases the optimization of very
    deep architectures.}
    \label{fig:resnet-block}
\end{figure}

Deep residual networks (ResNets)~\cite{he2016deep} introduced explicit
skip connections that add the input of a block to its output, enabling
the training of substantially deeper models (Figure \ref{fig:resnet-block}). A residual block computes
\begin{equation}
  x_{\text{out}} = x_{\text{in}} + \mathcal{F}(x_{\text{in}};\,\theta),
\end{equation}
where $\mathcal{F}$ is a small stack of convolution, normalization, and
nonlinearity layers, and $\theta$ denotes its parameters. This
architecture has become a standard backbone for many vision tasks,
including detection, segmentation, and early vision-language systems.

Subsequent work proposed more efficient and scalable convnet families,
such as EfficientNet, ResNeXt, and ConvNeXt, which refine the choice of
kernel sizes, width and depth multipliers, normalization layers, and
activation functions. From the perspective of a VLM, the key design
choice is often not the exact convnet variant, but rather:

\begin{itemize}
  \item the spatial resolution and stride of the final feature map;
  \item whether to use a single global representation or retain a grid
        of local features;
  \item and how to expose these features as visual tokens.
\end{itemize}

\subsection{Global Pooling and Tokenization}

For classification tasks, CNNs typically apply global average pooling
across spatial locations, followed by a linear classifier. In a
Vision-Language setting, however, we seldom want to discard spatial
structure entirely. Two common strategies are:

\begin{enumerate}
  \item \textbf{Global token:} apply global pooling to obtain a single
        feature vector $v_{\text{cls}} \in \mathbb{R}^d$, then treat
        this as a \emph{class} or \emph{image} token when interfacing
        with a language model.
  \item \textbf{Spatial tokens:} flatten the feature map into a sequence
        of $m = h \cdot w$ vectors and use each location as a separate
        visual token. This provides the language model or fusion module
        with finer-grained spatial information at the cost of longer
        sequences.
\end{enumerate}

The choice between these regimes depends on the downstream tasks: global
tokens may suffice for coarse classification or retrieval, while spatial
tokens are usually preferred for grounding and detailed question
answering.

\section{Vision Transformers and Patch-Based Encoders}
\label{sec:vit-encoders}

\begin{figure}[ht]
    \centering
    \includegraphics[width=\linewidth]{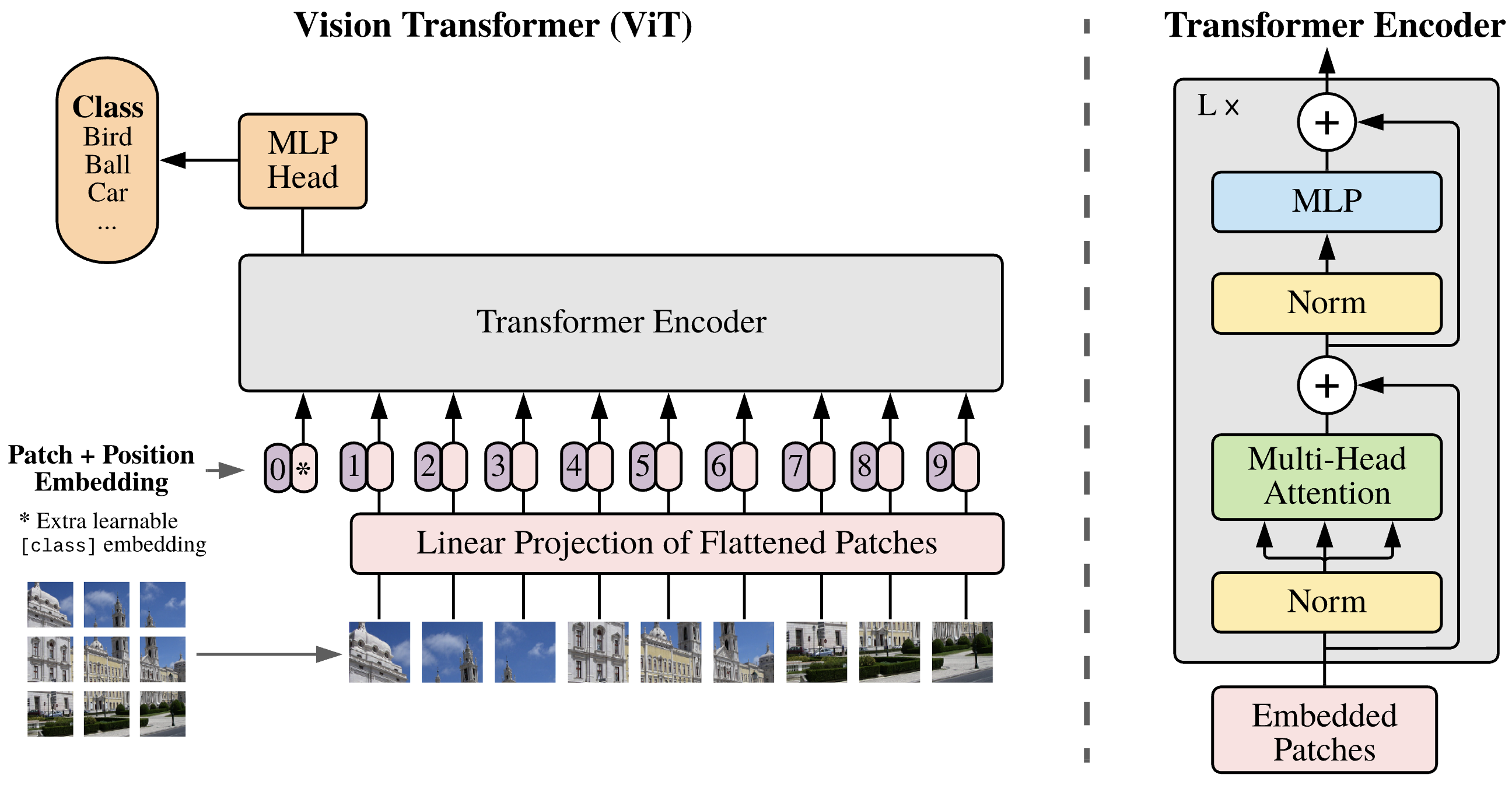}
    \caption{Architecture of the Vision Transformer (ViT)~\cite{dosovitskiy2021image}.
    The image is first divided into non-overlapping patches, which are
    flattened and linearly projected to obtain patch embeddings. A
    learnable \texttt{[CLS]} token is prepended, and positional
    embeddings are added to form the input sequence to a stack of
    transformer encoder layers. The output representation at the
    \texttt{[CLS]} position is passed through an MLP head to produce
    class logits, while all token outputs can be interpreted as
    contextualized patch features. The right panel shows the repeated
    transformer encoder block with multi-head self-attention, MLP, and
    residual connections.}
    \label{fig:vit-arch}
\end{figure}

While convolutional networks encode strong inductive biases for locality
and translation equivariance, they also constrain the way information
flows across the image. Vision Transformers (ViT)~\cite{dosovitskiy2021image}
and related architectures take a different approach: they treat an image
as a sequence of patches and use self-attention to model global
relationships from the outset. Figure~\ref{fig:vit-arch} summarizes the
standard ViT architecture.

\subsection{Patch Embeddings and Positional Information}

In a standard ViT, the input image is partitioned into a grid of
non-overlapping patches of size $P \times P$. Each patch is flattened
and linearly projected to an embedding:
\[
  \tilde{v}_j = W_{\text{patch}} \cdot \mathrm{vec}(I_j) + b_{\text{patch}},
  \quad j = 1, \dots, m,
\]
where $W_{\text{patch}} \in \mathbb{R}^{d \times (P^2 C)}$ and $m$ is
the number of patches. In addition, a learnable \emph{class token}
$v_{\text{cls}}$ (the \texttt{[CLS]} token in
Figure~\ref{fig:vit-arch}) is prepended to this sequence, and positional
embeddings are added to encode patch locations:
\[
  z_0 = [\,v_{\text{cls}}, \tilde{v}_1, \dots, \tilde{v}_m\,] +
        p_{0:m}.
\]

The sequence $z_0$ is then processed by a stack of transformer encoder
layers, each consisting of multi-head self-attention, a position-wise
MLP, and residual connections with layer normalization (right panel of
Figure~\ref{fig:vit-arch}). The output at the class token position can
be used as a global image representation for classification, while the
remaining outputs serve as spatially-aware patch tokens that can be
fed into downstream multimodal fusion modules.

\begin{didyouknow}
One subtle but important aspect of Vision Transformers is that the choice
of patch size implicitly defines the model's ``pixel vocabulary''. Smaller
patches give the encoder more spatial tokens and finer detail, but also
increase sequence length and computation. Larger patches behave more
like high-level ``visual words'', trading granularity for efficiency. Many practical VLMs quietly rely on this trade-off, choosing patch sizes that balance GPU budget, input resolution, and the level of detail needed by the downstream language model.
\end{didyouknow}

\subsection{Advantages and Variants}

Patch-based encoders offer several advantages in the context of
Vision-Language Models:

\begin{itemize}
  \item \textbf{Token-aligned representation.}
        They produce an ordered sequence of visual tokens that
        naturally matches the token-based representation used in
        language models, simplifying the design of multimodal fusion
        mechanisms.
  \item \textbf{Global context via self-attention.}
        Self-attention allows each token to attend directly to every
        other token, enabling flexible modeling of long-range
        dependencies and global image structure without relying on
        hand-crafted receptive-field hierarchies.
  \item \textbf{Scalable architecture.}
        The transformer architecture can be scaled in width, depth, and
        input resolution in a relatively uniform manner, providing a
        clear path from small models suitable for resource-constrained
        settings to large models used in state-of-the-art systems.
\end{itemize}

Building on the original ViT design, a broad family of variants has been
proposed to improve efficiency, robustness, and accuracy. These include
hierarchical transformers that operate at multiple spatial resolutions
(e.g., Swin-type architectures), hybrid backbones that combine
convolutional stages with transformer layers, and models that integrate
convolutional operations directly within transformer blocks. In practice,
Vision-Language Models frequently adopt a pre-trained ViT- or
CLIP-style encoder as their visual backbone, exposing either the global
class token, the full sequence of patch tokens, or a combination of both
to the multimodal fusion module.

\section{Multi-Scale and Region-Level Representations}
\label{sec:multi-scale-region}

Many vision tasks require reasoning at multiple spatial scales or about
specific regions rather than whole images. Object detection, instance
segmentation, and referring expression grounding are prominent examples.
To support such tasks, visual encoders often produce multi-scale feature
maps and region-level embeddings.

\subsection{Feature Pyramids}
\label{subsec:feature-pyramids}

Many visual recognition tasks must cope with objects that appear at
widely varying scales. Relying on a single feature map forces an
undesirable compromise: deep layers provide strong semantics but at
coarse spatial resolution, whereas early layers preserve fine
detail but are less discriminative. Feature Pyramid Networks
(FPNs)~\cite{lin2017feature} explicitly address this trade-off by
constructing a top-down hierarchy of feature maps with rich semantics at
all scales.

\begin{figure}[ht]
    \centering
    \begin{subfigure}[t]{0.48\linewidth}
        \centering
        \includegraphics[width=\linewidth]{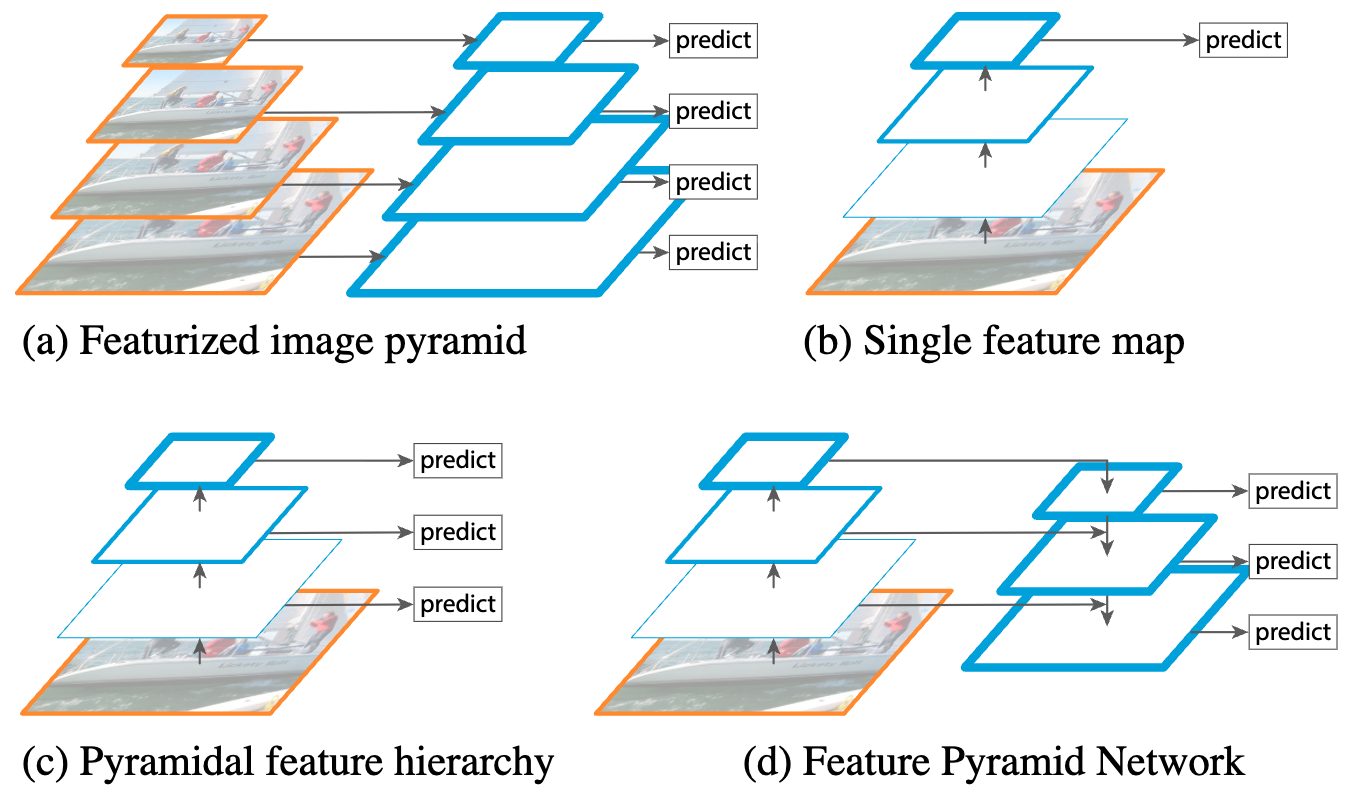}
        \caption{Design space for multi-scale feature representations,
        as discussed in \citet{lin2017feature}. Panel (a)
        illustrates a \emph{featurized image pyramid}, where the
        backbone is applied independently at multiple input resolutions.
        Panel (b) shows prediction from a \emph{single} deep feature
        map, which is semantically strong but spatially coarse. Panel
        (c) depicts a \emph{pyramidal feature hierarchy} obtained
        directly from the backbone, whose lower layers lack strong
        semantics. Panel (d) introduces the Feature Pyramid Network
        (FPN), which aims to combine the benefits of these designs by
        producing multi-scale feature maps that are both semantically
        rich and spatially detailed.}
        \label{fig:fpn-designs}
    \end{subfigure}
    \hfill
    \begin{subfigure}[t]{0.48\linewidth}
        \centering
        \includegraphics[width=\linewidth]{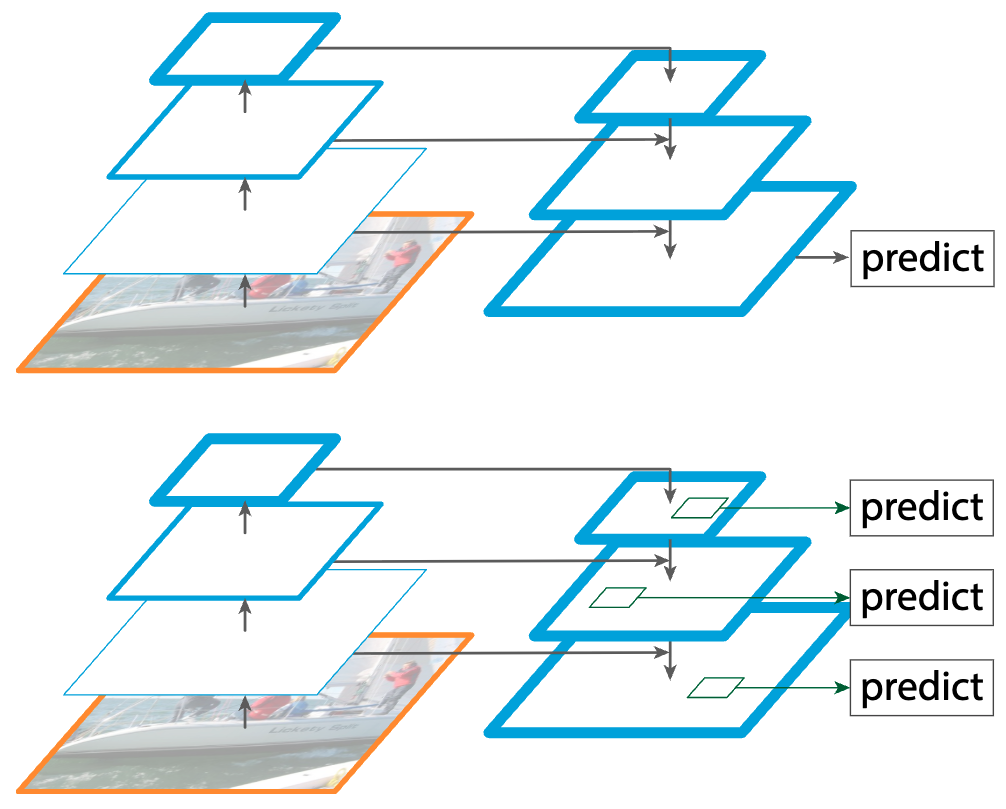}
        \caption{Top–down pathway with lateral connections in a Feature
        Pyramid Network~\cite{lin2017feature}. In the upper row,
        high-level, low-resolution feature maps are progressively
        upsampled and merged with corresponding lower-level features
        using lateral $1{\times}1$ convolutions, yielding a set of
        pyramid maps $\{P_l\}$ that share strong semantics across
        scales. In the lower row, task-specific prediction heads
        operate on each pyramid level, allowing the detector to assign
        anchors or proposals to the feature map that is most appropriate
        for their spatial scale, thereby improving both small-object
        recall and large-object localization.}
        \label{fig:fpn-topdown}
    \end{subfigure}
    \caption{Illustrations of Feature Pyramid Networks (FPNs) for
    multi-scale object detection, adapted from \citet{lin2017feature}. FPNs exploit the inherent feature
    hierarchy of deep convolutional backbones to construct a pyramid of
    semantically strong feature maps at multiple resolutions, each
    equipped with its own prediction head. This design enables accurate
    detection of objects over a wide range of scales while remaining
    computationally efficient.}
    \label{fig:fpn-overview}
\end{figure}

Figure~\ref{fig:fpn-overview} summarizes the design space considered in
the original FPN work. Traditional approaches either build a
\emph{featurized image pyramid}, running the backbone independently at
multiple input resolutions, or attach prediction heads to a single deep
feature map. Both strategies suffer from inefficiencies or poor
small-object performance. FPNs instead leverage the inherent pyramid of
features produced by a deep backbone (e.g., $\{C_2, C_3, C_4, C_5\}$)
and augment it with a top-down pathway and lateral connections to obtain
a set of multi-scale, semantically strong feature maps
$\{P_2, P_3, P_4, P_5\}$.

In the context of Vision-Language Models, such multi-scale
representations provide a richer pool of candidate visual tokens. A
VLM may, for example, derive global tokens from coarse pyramid levels
for high-level scene understanding, while using higher-resolution
features from shallower levels to support fine-grained grounding of
referring expressions or dense captioning. Designing how tokens are
sampled or pooled from the feature pyramid-for instance, via uniform
grid sampling, region proposals, or learned attention over scales-is a
key modeling decision that directly impacts the quality of multimodal
reasoning.

\subsection{Region Features and Proposals}

Region-based encoders, introduced in the context of R-CNN and
Faster R-CNN, focus on semantically meaningful regions such as objects,
text segments, or salient parts of the scene. A typical pipeline
involves:

\begin{enumerate}
  \item Using a convnet backbone (with or without a feature pyramid) to
        compute dense feature maps.
  \item Generating region proposals-candidate bounding boxes-using a
        Region Proposal Network (RPN) or another proposal mechanism.
  \item Applying RoI pooling or RoI align to extract a fixed-size
        feature representation for each region.
\end{enumerate}

Each pooled region feature can be viewed as a visual token corresponding
to a particular object or area. Early vision-language systems for
Visual Question Answering often used dozens or hundreds of such
region-level features as the visual input to an attention-based fusion
module. This sparse, object-centric representation is especially useful
for tasks that require fine-grained reasoning about discrete entities.

\section{Pretraining Objectives for Visual Encoders}
\label{sec:vision-pretraining}

The effectiveness of a visual encoder depends not only on its
architecture, but also on the objectives and data used during
pretraining. In practice, two broad classes of pretraining strategies
have been particularly influential: supervised classification and
self-supervised representation learning.

\subsection{Supervised Classification Pretraining}

A conventional and still widely used approach is to pretrain visual
encoders on large-scale labeled datasets such as ImageNet, using a
multi-class classification loss. In this setting, the network learns to
map each input image $I$ to a distribution over semantic categories
(e.g., $1,000$ ImageNet classes)~\cite{russakovsky2015imagenet}. The
intermediate activations-especially those from higher layers-serve as
generic visual features that can be reused across tasks.

Formally, given an image-label pair $(I, y)$ with $y \in \{1,\dots,K\}$,
a classifier with parameters $\theta$ is trained to minimize the
cross-entropy loss
\begin{equation}
  \mathcal{L}_{\text{cls}}(\theta)
  = - \log p_\theta(y \mid I),
\end{equation}
where $p_\theta(y \mid I)$ is obtained by applying a linear classifier
to a pooled feature vector and normalizing with a softmax. Architectures
such as AlexNet~\cite{krizhevsky2012imagenet}, VGG~\cite{simonyan2015very},
and ResNet~\cite{he2016deep} pre-trained on ImageNet have historically
served as standard backbones for detection, segmentation, and, by
extension, early vision-language systems.

When a supervised encoder is later integrated into a VLM, its features
often provide a strong initialization, particularly for object-centric
tasks where the label space of the pretraining data overlaps with that
of the downstream domain. However, classification labels capture only a
narrow slice of the information present in natural images. They tend to
emphasize object identity while underutilizing fine-grained attributes,
context, and low-level structure. Moreover, domain shifts between the
pretraining corpus and the target data distribution (for example, from
natural images to documents or medical imagery) can significantly limit
transfer performance.

\subsection{Self-Supervised and Masked Image Modeling}
\label{subsec:selfsup-mim}

Self-supervised learning aims to learn rich visual representations
without relying on manually annotated labels. Instead, models are
trained on \emph{pretext tasks} defined directly on raw images, such as
predicting transformations, distinguishing between augmented views of
the same image, or reconstructing missing content.

\begin{figure}[ht]
    \centering
    \includegraphics[width=0.5\linewidth]{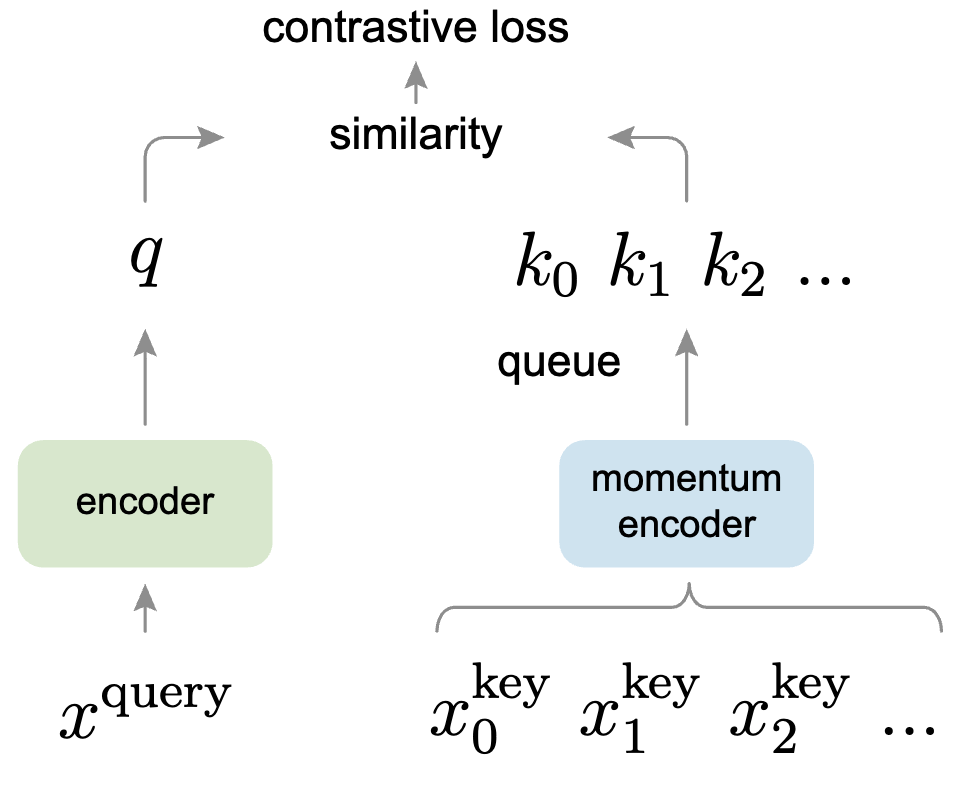}
    \caption{Illustration of the Momentum Contrast framework for
    self-supervised representation learning~\cite{he2020momentum}. An
    encoder processes an augmented \emph{query} image to produce a
    representation $q$, while a separate momentum encoder produces
    \emph{key} representations $\{k_i\}$ for other augmented images that
    are stored in a queue acting as a dynamic dictionary. A contrastive
    loss is applied to maximize the similarity between $q$ and its
    corresponding positive key while minimizing similarity to a large
    set of negative keys in the queue, enabling effective learning with
    many negatives even under modest batch sizes.}
    \label{fig:moco-architecture}
\end{figure}

Contrastive methods, such as SimCLR~\cite{chen2020simple} and Momentum
Contrast (MoCo)~\cite{he2020momentum}, encourage representations of
different augmentations of the same image to be close in embedding space
while pushing apart embeddings of different images. MoCo, in particular,
interprets contrastive learning as building a dynamic dictionary with a
queue and a momentum-updated encoder. As illustrated in
Figure~\ref{fig:moco-architecture}, one encoder processes a \emph{query}
view $x^{\text{query}}$ to produce a representation $q$, while a
momentum encoder processes \emph{key} views
$\{x^{\text{key}}_i\}_{i=0}^{K}$ to produce keys
$\{k_i\}_{i=0}^{K}$ that are stored in a queue. The query is trained to
be similar to its matching key and dissimilar to all other keys using an
InfoNCE-style loss.

\begin{figure}[ht]
    \centering
    \includegraphics[width=\linewidth]{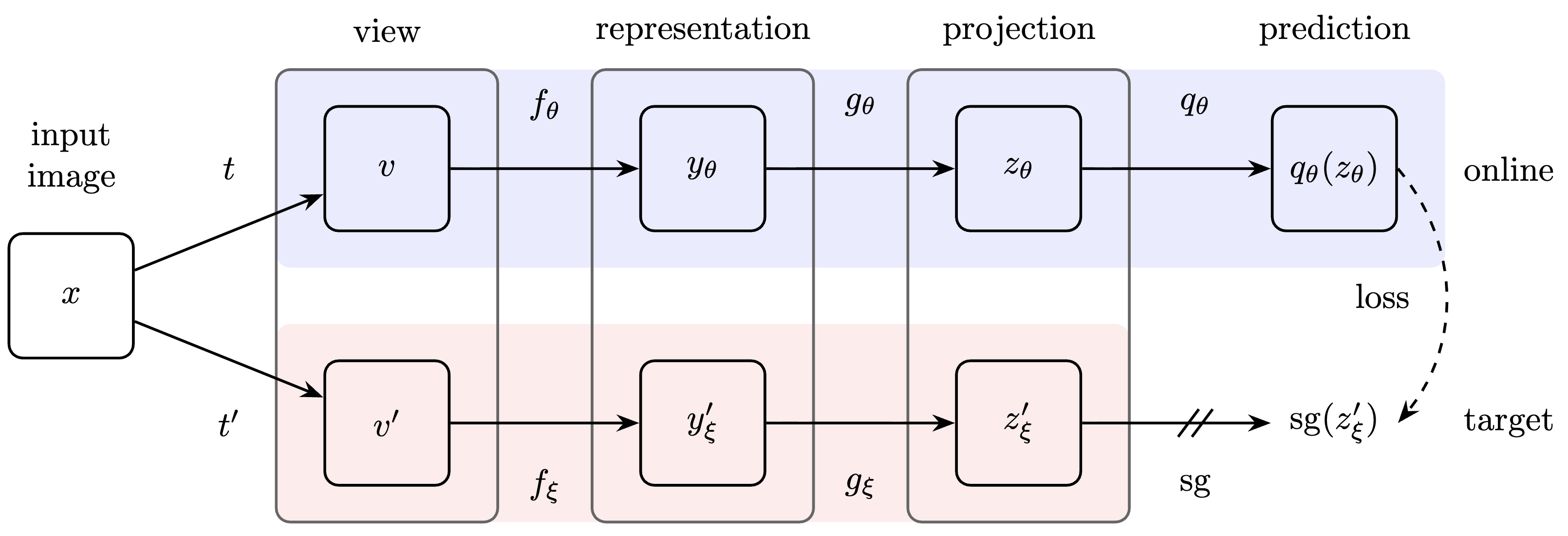}
    \caption{Architecture of Bootstrap Your Own Latent (BYOL) for
    self-supervised visual representation learning~\cite{grill2020bootstrap}.
    Given an input image $x$, two augmented views $t(x)$ and $t'(x)$ are
    generated. The \emph{online} branch (top, parameters $\theta$)
    applies an encoder $f_\theta$ and projector $g_\theta$ to obtain a
    representation $y_\theta$ and projection $z_\theta$, followed by a
    prediction head $q_\theta$ that outputs $q_\theta(z_\theta)$. The
    \emph{target} branch (bottom, parameters $\xi$) applies a separate
    encoder $f_\xi$ and projector $g_\xi$ to the second view, producing
    a target projection $z'_\xi$; a stop-gradient operator
    $\operatorname{sg}(\cdot)$ prevents gradients from flowing into the
    target network. The loss encourages $q_\theta(z_\theta)$ to match
    $\operatorname{sg}(z'_\xi)$ (and vice versa with roles of the views
    swapped), while the target parameters $\xi$ are updated as an
    exponential moving average of $\theta$. This self-distillation
    scheme learns useful representations without explicit negative
    pairs.}
    \label{fig:byol-architecture}
\end{figure}

Clustering- and distillation-based methods, including Bootstrap Your Own
Latent (BYOL)~\cite{grill2020bootstrap} and DINO~\cite{caron2021emerging},
avoid explicit negative pairs and instead learn by predicting slowly
evolving target representations. BYOL, in particular, maintains two
networks: an \emph{online} network that is directly optimized by the
loss and a \emph{target} network whose parameters are updated by an
exponential moving average of the online parameters. As illustrated in
Figure~\ref{fig:byol-architecture}, two augmented views $t(x)$ and
$t'(x)$ of the same image are processed by the online encoder-projector
pair $(f_\theta, g_\theta)$ and the target encoder-projector pair
$(f_\xi, g_\xi)$, respectively. A prediction head $q_\theta$ maps the
online projection $z_\theta$ to a prediction $q_\theta(z_\theta)$, which
is trained to match the stop-gradient target projection
$\text{sg}(z'_\xi)$. Symmetrizing this objective over the two views
yields a self-distillation signal that encourages invariance to data
augmentations without requiring negative examples.

More recently, masked image modeling (MIM) has emerged as a powerful
pretraining paradigm, particularly for transformer-based encoders.
Masked Autoencoders (MAE)~\cite{he2022masked}, for example, randomly
mask a large fraction of image patches and train a model to reconstruct
the missing content from the visible patches, closely mirroring masked
language modeling in NLP.

Self-supervised encoders trained with these objectives often exhibit
improved robustness, better sample efficiency, and stronger performance
under distribution shift compared to purely supervised counterparts.
When combined with large-scale image-text pretraining in a VLM,
self-supervised visual backbones contribute to multimodal
representations that capture both low-level structure and high-level
semantics.

\chapter{Language Models for Vision-Language Systems}
\label{chap:lms-for-vlms}

\textit{In Vision-Language Models, the language component plays a dual
role. On the one hand, it acts as a powerful sequence model that can
encode and generate natural language. On the other hand, it serves as a
general-purpose reasoning engine that integrates information from visual
inputs with prior knowledge about the world. Achieving this dual role
requires both a strong language model (LM) and an effective mechanism
for aligning visual and textual representations.}

\section{Foundations of Neural Language Models}
\label{sec:lm-foundations}

\subsection{Language Modeling as Conditional Probability Estimation}

At a high level, a language model defines a probability distribution
over sequences of tokens. Let
$\mathbf{t} = (t_1, t_2, \dots, t_n)$ denote a sequence of tokens from
a vocabulary $\mathcal{V}_{\text{text}}$. A \emph{left-to-right}
autoregressive language model factorizes the joint distribution as
\begin{equation}
  p(\mathbf{t})
  = p(t_1, \dots, t_n)
  = \prod_{i=1}^n p(t_i \mid t_1, \dots, t_{i-1}).
  \label{eq:lm-factorization}
\end{equation}
Training proceeds by maximizing the log-likelihood of observed text
under this factorization, which is equivalent to minimizing the
cross-entropy loss between the model’s next-token predictions and the
ground-truth tokens.

This formulation is conceptually simple but extremely flexible. Any
conditioning information-including prompts, metadata, or visual
tokens-can be prepended or otherwise incorporated into the context,
thereby influencing the conditional distributions in
(\ref{eq:lm-factorization}). This observation underlies many of the
multimodal alignment strategies discussed later in the chapter.

\subsection{From Recurrent Networks to Transformers}
\label{subsec:rnn-to-transformer}

\begin{figure}[ht]
    \centering
    \includegraphics[width=0.85\linewidth]{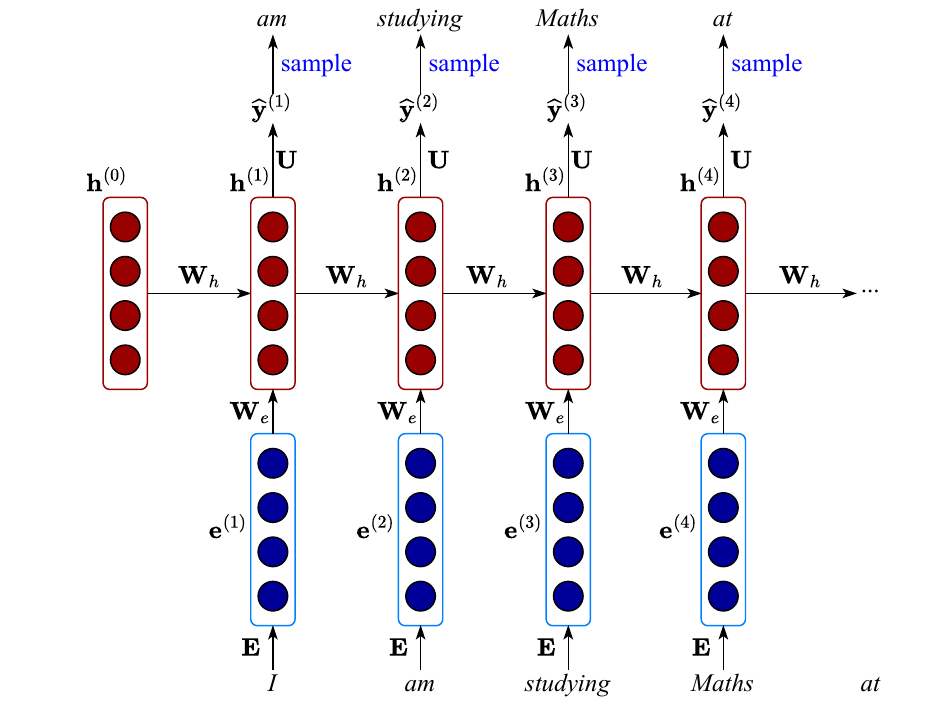}
    \caption{Recurrent neural network for language modeling, unrolled
    over time. At each time step $t$, the current input word $w_t$ is
    mapped to an embedding $e^{(t)}$ via an embedding matrix $E$, then
    combined with the previous hidden state $h^{(t-1)}$ through a
    recurrent transformation parameterized by $W_h$ to produce
    $h^{(t)}$. An output layer with parameters $U$ converts $h^{(t)}$
    into a distribution $\hat{y}^{(t)}$ over the vocabulary, from which
    the next word can be sampled. The same parameters are reused at all
    time steps, and the network is trained so that predictions
    $\hat{y}^{(t)}$ match the ground-truth next tokens in natural
    language sequences.}
    \label{fig:rnn-arch}
\end{figure}

Early neural language models were built using recurrent architectures,
such as Elman networks~\cite{elman1990finding}, Long Short-Term Memory
(LSTM) networks~\cite{hochreiter1997long}, and GRUs~\cite{cho2014learning},
which process tokens sequentially and maintain a hidden state that
summarizes the past. For an input sequence of tokens
$(w_1, \dots, w_T)$, each token is first mapped to an embedding
$e^{(t)} \in \mathbb{R}^d$ through a learned embedding matrix $E$. A
recurrent layer then updates its hidden state according to
\[
  h^{(t)} = f_{\theta}\big(h^{(t-1)}, e^{(t)}\big),
\]
where $f_{\theta}$ denotes the recurrent transition function (e.g., an
RNN or LSTM cell). At each time step, an output layer with parameters
$U$ maps $h^{(t)}$ to a distribution over the vocabulary, from which the
next token can be sampled. During training, the model is typically
optimized with teacher forcing, using the ground-truth token $w_t$ as
input at step $t$ while predicting $w_{t+1}$. Figure~\ref{fig:rnn-arch} illustrates this
recurrent language modeling setup unrolled over time.

\begin{figure}[ht]
    \centering
    \includegraphics[width=0.53\linewidth]{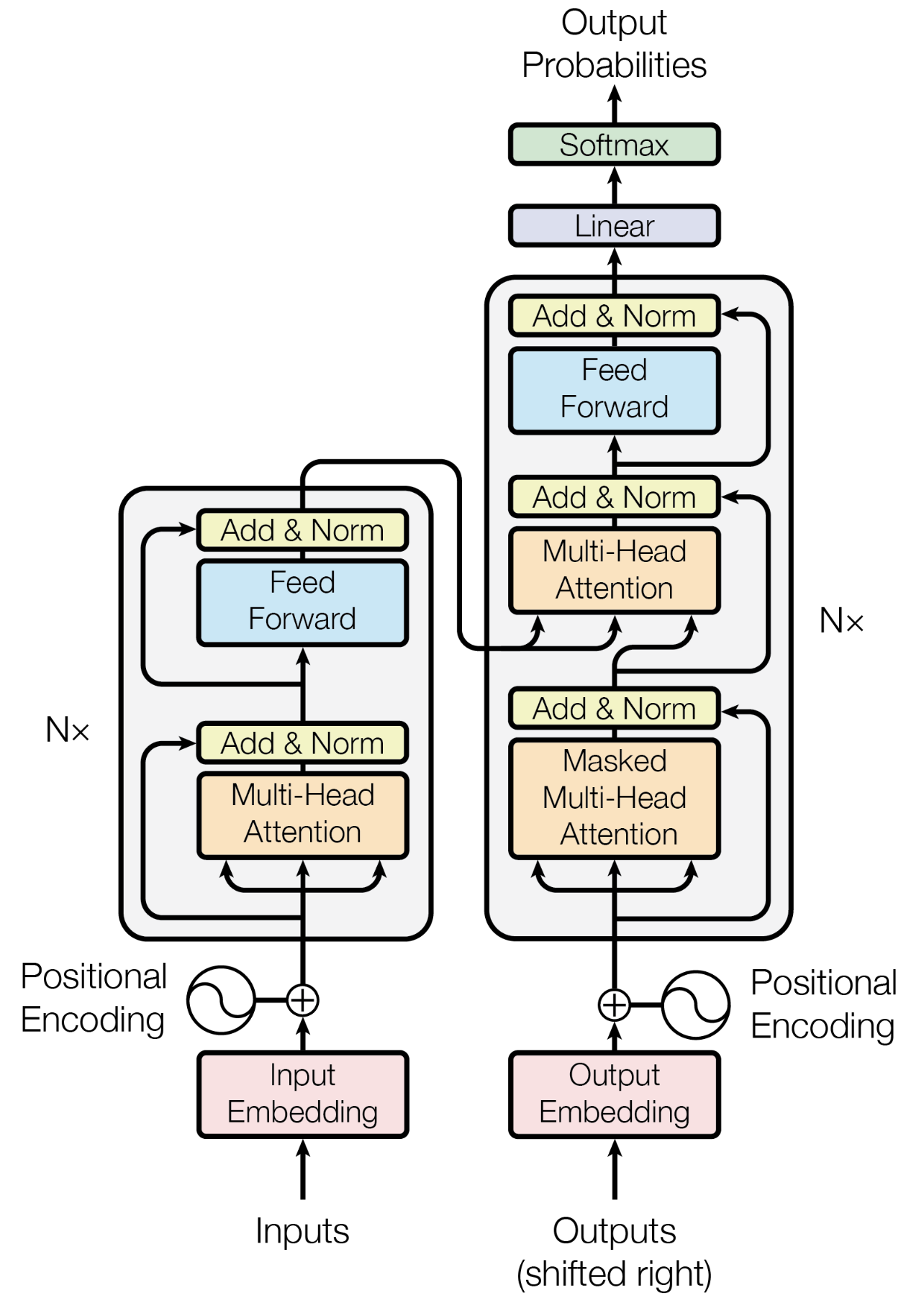}
    \caption{Transformer encoder–decoder architecture for sequence
    modeling~\cite{vaswani2017attention}. On the left, the encoder takes
    input tokens, maps them to embeddings, adds positional encodings,
    and processes the sequence through $N$ stacked blocks, each
    comprising multi-head self-attention and a position-wise
    feed-forward network, wrapped with residual connections and layer
    normalization (``Add \& Norm''). On the right, the decoder receives
    the output tokens (shifted right), adds positional encodings, and
    applies a masked multi-head self-attention layer, followed by
    encoder–decoder (cross) attention over the encoder outputs and a
    feed-forward network. A final linear layer and softmax map the
    decoder representations to output probabilities over the
    vocabulary.}
    \label{fig:transformer-arch}
\end{figure}

Despite their conceptual simplicity, recurrent architectures process
tokens strictly sequentially and rely on a single evolving hidden state
to carry information across many time steps, which limits parallelism
and makes very long-range dependencies difficult to capture in practice. The transformer architecture~\cite{vaswani2017attention} replaces
explicit recurrence with stacks of self-attention and feed-forward
layers that operate on entire sequences in parallel. In its original
formulation, the model follows an encoder-decoder design for
sequence-to-sequence tasks such as machine translation. As shown in
Figure~\ref{fig:transformer-arch}, the encoder (left) maps an input
sequence of token embeddings, augmented with positional encodings, to a
sequence of contextual representations via $N$ repeated blocks of
multi-head self-attention and position-wise feed-forward networks, each
wrapped in residual connections and layer normalization (``Add \& Norm'').

The decoder (right) processes the output sequence, shifted by one
position to preserve autoregressive causality. Each of its $N$ blocks
contains a masked multi-head self-attention layer, an encoder-decoder
(or cross-) attention layer that attends to the encoder outputs, and a
feed-forward sub-layer, again with residual connections and
normalization. A final linear layer and softmax produce a distribution
over the vocabulary at each position. In decoder-only language models commonly used in modern VLMs, the encoder and encoder–decoder attention are omitted, but the core building blocks - multi-head self-attention, feed-forward networks, and residual normalization - remain unchanged.

Formally, given an input sequence of token embeddings
$(e_1, \dots, e_n)$, a self-attention layer computes, for each position
$i$, a weighted combination of all embeddings:
\begin{equation}
  \mathrm{Attn}(e_i)
  = \sum_{j=1}^n \alpha_{ij} \, W_V e_j,
\end{equation}
where the attention weights $\alpha_{ij}$ are obtained by comparing
learned queries $W_Q e_i$ and keys $W_K e_j$ via a scaled dot-product
followed by softmax normalization. Stacking such layers yields a
powerful, scalable sequence model that admits full parallelization over
sequence positions and has become the dominant backbone for large-scale
language models and, by extension, many Vision-Language Models.

\begin{didyouknow}
One practical reason transformers displaced recurrent networks in large
language modeling is not just accuracy, but \emph{hardware fit}. RNNs
and LSTMs process tokens strictly one step at a time, which makes it
hard to fully exploit modern GPUs and TPUs. Self-attention, by contrast,
lets the model see the entire sequence at once, so all positions in a
batch can be processed in parallel. This shift from ``time-step
sequential'' to ``position-parallel'' computation is a big part of why
scaling to hundreds of layers and context lengths of thousands of tokens
became feasible.
\end{didyouknow}

\subsection{Autoregressive Transformers as Foundation Models}
\label{subsec:autoregressive-foundation}

Autoregressive transformer language models trained on large text corpora
have emerged as prototypical \emph{foundation models} for natural
language processing~\cite{brown2020language,bommasani2021opportunities}.
In addition to modeling the distribution in
(\ref{eq:lm-factorization}), these models accumulate substantial world
knowledge and learn intermediate representations that are broadly
useful across tasks. Classic examples range from GPT-2, a large
decoder-only transformer trained as a next-token predictor on web
text~\cite{radford2019language}, to GPT-3 and related models that scale
this paradigm to hundreds of billions of parameters and trillions of
tokens~\cite{brown2020language}. Bidirectional encoders such as
BERT~\cite{devlin2019bert} demonstrate a complementary pattern: a single
pre-trained transformer can be adapted, via either task-specific heads
or prompting, to a wide variety of classification and span-prediction
tasks.

A distinctive property of these models is their flexibility in
\emph{adaptation}. Once an autoregressive transformer has been
pre-trained on generic text, downstream behavior can be specialized
through full fine-tuning, parameter-efficient methods (e.g., adapters or
LoRA-style low-rank updates), prompt engineering, or instruction
tuning. Empirically, a well-trained foundation model often outperforms
task-specific architectures even when only lightly adapted, reflecting
the breadth of linguistic and factual regularities captured during
pretraining.

In the context of Vision-Language Models, this foundation model
perspective has two important consequences. First, the language model
can shoulder much of the burden of reasoning, commonsense inference, and
discourse management, effectively acting as a general-purpose
multimodal ``controller''. Second, it becomes attractive to reuse a
frozen or lightly tuned text-only LM as the core of a VLM, and to focus
training on \emph{alignment} modules that inject visual context into the
model~\cite{alayrac2022flamingo,li2023blip2}. The central challenge is
therefore to provide the LM with visual information in a representation
it can consume efficiently, without destroying the capabilities acquired
during large-scale text pretraining.

\section{Conditioning Language Models on Visual Input}
\label{sec:conditioning-visual}

To transform a text-only language model into a Vision-Language Model,
one must make visual information available to the LM in a way that is
both expressive and computationally efficient. Let
$\mathbf{v} = (v_1, \dots, v_m)$ denote the sequence of visual tokens
produced by a visual encoder. The goal is to model, for example, the
conditional distribution
\begin{equation}
  p(\mathbf{t} \mid \mathbf{v})
  = \prod_{i=1}^n p(t_i \mid t_1, \dots, t_{i-1}, \mathbf{v}),
  \label{eq:vlm-conditional}
\end{equation}
where the dependence on $\mathbf{v}$ is implemented through some
interaction mechanism between visual and textual representations.

Over the past few years, several design patterns have emerged for
injecting visual context into large language models.

\subsection{Prefix and Prompt-Based Conditioning}
\label{subsec:prefix-conditioning}

One simple strategy is to map visual tokens into the same embedding
space as text tokens and insert them into the LM’s input sequence as a
\emph{prefix} or \emph{prompt}. Concretely, a projection module
$\phi_{\text{vis}\rightarrow\text{text}}$ transforms each visual token
$v_j$ into a pseudo-token $\tilde{t}_j$ in the language model’s
embedding space. The LM then receives an extended sequence
\[
  (\tilde{t}_1, \dots, \tilde{t}_m, t_1, \dots, t_n),
\]
and is trained or prompted to generate textual outputs conditioned on
both the pseudo-tokens and the subsequent text. Variants of this idea
appear in models that couple frozen language models with visual
encoders via lightweight projection or adapter layers, such as
Flamingo~\cite{alayrac2022flamingo} and BLIP-2~\cite{li2023blip2}.

Prefix-based conditioning has the advantage of minimal architectural
changes: the language model itself can remain largely unchanged, and all
modality-specific handling is confined to the visual encoder and the
projection module. It also naturally leverages the LM’s existing
abilities in in-context learning and prompt following. However, it may
be less flexible than more explicit cross-attention mechanisms,
particularly when visual and textual tokens need to interact at multiple
layers or at different levels of granularity.

\subsection{Cross-Attention to Visual Tokens}
\label{subsec:cross-attn-conditioning}

A second, more expressive strategy is to introduce \emph{cross-attention}
layers that allow textual representations to attend directly to visual
tokens. In this setup, the language model maintains its own sequence of
hidden states $(h_1, \dots, h_n)$, but each layer (or a subset of
layers) includes a cross-attention module:
\begin{equation}
  \mathrm{CrossAttn}(h_i)
  = \sum_{j=1}^m \beta_{ij} \, W_V^{\text{vis}} v_j,
\end{equation}
where the attention weights $\beta_{ij}$ are obtained by comparing
textual queries $W_Q^{\text{text}} h_i$ against visual keys
$W_K^{\text{vis}} v_j$. The resulting cross-attended representation can
then be combined with $h_i$ through a residual connection and passed to
subsequent layers.

This pattern was popularized by early vision-language transformers such
as ViLBERT~\cite{lu2019vilbert} and LXMERT~\cite{tan2019lxmert}, which
employ dedicated cross-attention modules to fuse image-region features
with textual tokens. More recent systems, including Flamingo, use
cross-attention from a frozen language model into visual tokens while
keeping most LM parameters fixed~\cite{alayrac2022flamingo}. Cross-
attention makes the interaction between modalities explicit and local to
specific layers, allowing different parts of the model to focus on
visual information to varying degrees. It also supports asymmetric
conditioning patterns, such as grounding only certain tokens (e.g.,
nouns or referring expressions) in visual context, while leaving purely
linguistic reasoning largely unchanged.

\subsection{Encoder-Decoder Architectures}
\label{subsec:encdec-conditioning}

A third design, less common in recent large VLMs but conceptually
important, uses an encoder-decoder transformer. In this setting, the
visual encoder (or a joint vision-text encoder) produces a sequence of
context embeddings that serve as the \emph{source} sequence, while a
text decoder generates the \emph{target} sequence via cross-attention to
the encoder outputs. The decoder then generates text autoregressively,
conditioned on both the previously generated tokens and the visual
context, much like in neural machine translation.

This encoder-decoder formulation underlies many early neural image
captioning systems. A canonical example is the Show and Tell model of
\citet{vinyals2015show}, which will be examined further in Chapter \ref{chap:arch-train}.

\begin{didyouknow}
Many modern LVLMs quietly mix more than one conditioning pattern in the
same model. For example, a system may use prefix-style visual tokens at
the input to keep compatibility with an existing text-only LM, while
also adding cross-attention blocks in a subset of layers for stronger
grounding. In practice, the choice between prefixing, cross-attention,
and encoder-decoder designs is often driven less by theory and more by
engineering constraints such as context length, memory budget, and how
much of the original language model you are allowed to modify.
\end{didyouknow}

\section{From Conditioning Mechanisms to Full VLM Architectures}
\label{sec:lm-to-architectures}

The discussion in this chapter has focused on the language side of
Vision-Language Models: how neural language models are formulated, how
transformers displaced recurrent networks as the dominant architecture,
and how large autoregressive transformers can act as general-purpose
foundation models. We have also surveyed three principal mechanisms for
injecting visual information into these models-prefix and
prompt-based conditioning, cross-attention to visual tokens, and
encoder-decoder formulations exemplified by early captioning systems.

Taken together, these ingredients specify \emph{how} a language model
can, in principle, consume visual context. They do not, however, fully
determine the behavior of a Vision-Language Model in practice. The
resulting system still depends critically on:

\begin{itemize}
  \item \textbf{The choice of visual backbone and tokenization scheme},
        which govern what visual information is available to the
        language model (e.g., global image embeddings, patch tokens, or
        region features).
  \item \textbf{The training objectives used to couple vision and
        language}, including contrastive alignment losses, matching and
        classification objectives, generative language modeling
        conditioned on images, and instruction-style supervision.
  \item \textbf{The bridge modules that connect modalities}, such as
        linear projectors, adapter layers, or specialized multimodal
        transformers, which mediate between visual features and the
        internal representation space of the language model.
  \item \textbf{The overall training paradigm}, including whether
        components are trained from scratch or initialized from
        pre-trained models, and whether the visual encoder, language
        model, or both are frozen, partially fine-tuned, or jointly
        optimized.
\end{itemize}

The next chapter will examine the main families of
multimodal alignment objectives, discusses how adapters, projectors, and
bridge transformers are used to connect pre-trained vision and language
backbones, and analyzes the trade-offs between different training
regimes. Together, these considerations complete the picture of how
language models are embedded within end-to-end Vision-Language
architectures.
\chapter{Architectural Design and Training Paradigms}
\label{chap:arch-train}

\textit{From a systems perspective, a VLM is specified by three main decisions: (1) The structure of the vision and language backbones and how they
        interact; (2) The objectives used to align their representations; and (3) The training paradigm used to optimize the combined system. 
        }

\section{Representative Vision-Language Architectures}
\label{sec:rep-architectures}

The abstract design dimensions discussed in
Chapter~\ref{chap:lms-for-vlms} become clearer when instantiated in
specific models.

\subsection{Encoder-Decoder Captioning}
\label{subsec:show-and-tell-arch}

\begin{figure}[t]
    \centering
    \includegraphics[width=0.55\linewidth]{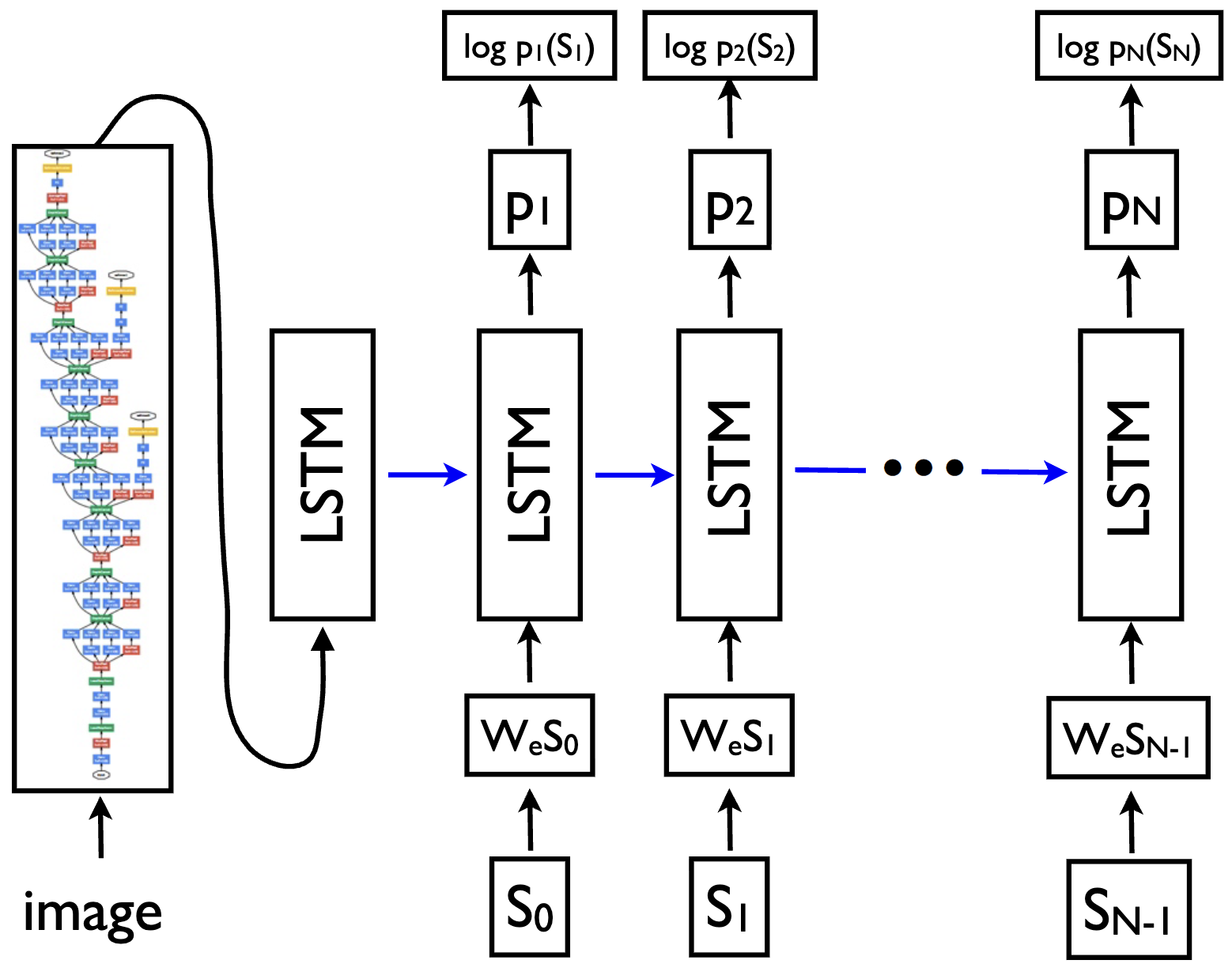}
    \caption{Architecture of the Show and Tell image captioning model
    of \citet{vinyals2015show}. A convolutional network
    encodes the input image into a fixed-dimensional feature vector,
    which is injected into an LSTM-based decoder. Starting from a
    special start-of-sentence token $S_0$, the LSTM consumes the image
    representation and the previously generated tokens
    $(S_0, S_1, \dots, S_{t-1})$ to produce a probability distribution
    over the next token $S_t$ at each time step. The caption is thus
    generated autoregressively, conditioned on the image throughout the
    decoding process.}
    \label{fig:show-and-tell}
\end{figure}

Show and Tell~\cite{vinyals2015show} is an early but influential neural
image captioning system that instantiates the encoder-decoder paradigm
using a convolutional network and an LSTM language model, illustrated in
Figure~\ref{fig:show-and-tell}. A convolutional network first encodes
the input image into a fixed-dimensional feature vector, which is then
used to initialize (or provide the first input to) an LSTM decoder. The
LSTM receives the image representation together with the previously
generated words, and at each step produces a distribution over the next
word in the caption. In this view, image captioning is cast directly as
a conditional language modeling problem, where the conditioning signal
is the image embedding produced by the visual encoder.

\paragraph{Visual encoder.}
The visual encoder is a deep convolutional network pre-trained on
ImageNet-style classification. The final global average pooling layer
produces a single feature vector $\mathbf{v} \in \mathbb{R}^{d_v}$ that
summarizes the entire image. No spatial structure is exposed to the
decoder; the image is compressed into one global embedding.

\paragraph{Language model.}
The decoder is a unidirectional LSTM that models the caption as a
sequence $(w_1,\dots,w_T)$. At each step $t$, the LSTM consumes the
previous word embedding and the recurrent hidden state, and outputs a
distribution over the next word. Training maximizes the conditional
log-likelihood $p(\mathbf{w} \mid I)$ using teacher forcing.

\paragraph{Vision-language interaction.}
The coupling between vision and language is deliberately simple. The
image vector $\mathbf{v}$ is either used to initialize the LSTM hidden
state or injected as an additional “start” embedding. All visual
information must therefore flow through a single vector and a single
recurrent state. This architecture exemplifies \emph{global-feature
conditioning}: a powerful text decoder, a purely unimodal CNN encoder,
and a minimal bridge at initialization time.

Although modern VLMs typically rely on richer visual tokenization, Show
and Tell remains conceptually important: many later models can be seen
as progressively refining this pattern by exposing more structured
visual representations and more expressive fusion mechanisms.

\subsection{Unified Multimodal Pretraining: BLIP and BLIP-2}
\label{subsec:blip-arch}

BLIP~\cite{li2022blip} and BLIP-2~\cite{li2023blip2} illustrate how a
single framework can support both understanding-oriented tasks
(retrieval, VQA) and generative tasks (captioning) while reusing strong
pre-trained vision and language backbones.

\paragraph{BLIP: multimodal mixture of encoder and decoder.}

\begin{figure}[ht]
    \centering
    \includegraphics[width=\linewidth]{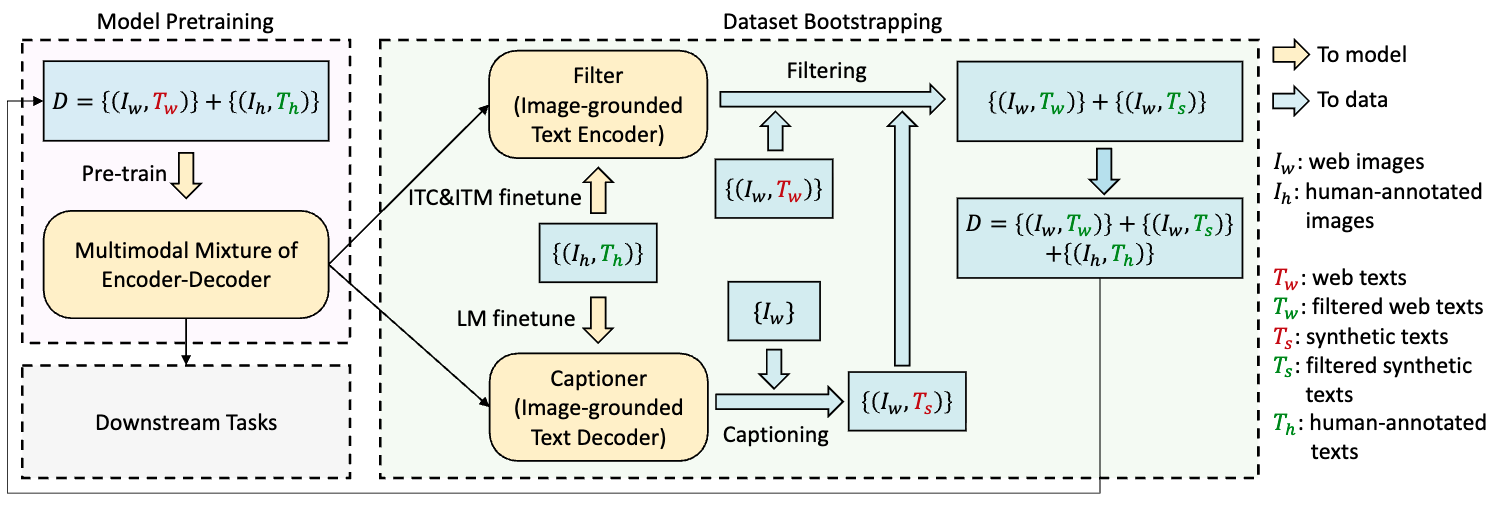}
    \caption{Schematic overview of the BLIP architecture
    \cite{li2022blip}. A pre-trained Vision Transformer encodes the
    input image into a sequence of patch tokens, while a text
    Transformer processes wordpiece token embeddings. BLIP supports
    three main pretraining objectives: (i) \emph{image-text
    contrastive} (ITC), which uses separate image and text encoders to
    map images and captions into a shared embedding space for
    retrieval-style alignment; (ii) \emph{image-text matching} (ITM),
    in which a multimodal encoder with cross-attention predicts whether
    an image-text pair is semantically matched; and (iii) \emph{image-
    conditioned captioning}, where a text decoder generates captions
    autoregressively conditioned on visual tokens. By sharing
    components across these objectives, BLIP learns joint
    representations that are useful for both discriminative and
    generative vision-language tasks.}
    \label{fig:blip-arc}
\end{figure}

BLIP introduces a \emph{Multimodal Mixture of Encoder-Decoder} (MED)
architecture~\cite{li2022blip} that unifies these objectives within a
single model family (Figure \ref{fig:blip-arc}). A Vision Transformer (ViT) encodes each image into
a sequence of patch tokens, and a text Transformer embeds captions or
queries into wordpiece tokens. The core Transformer blocks are used in a
flexible way, operating either as a multimodal encoder or as a
vision-conditioned decoder, depending on the task:

\begin{itemize}
  \item \emph{Encoder mode (understanding).} Image and text tokens are
        fed into a multimodal encoder equipped with cross-attention,
        producing joint representations that capture fine-grained
        correspondences between visual regions and words. This mode is
        used for image-text matching (ITM) and for extracting features
        for downstream classification-style tasks (e.g., VQA heads).

  \item \emph{Dual-encoder mode (contrastive alignment).} For image-text
        contrastive (ITC) pretraining, BLIP uses separate image and text
        encoders to produce global embeddings $f(I)$ and $g(\mathbf{t})$
        in a shared space. A symmetric contrastive loss over similarity
        scores encourages aligned image-caption pairs to be close and
        misaligned pairs to be distant, improving retrieval and
        zero-shot recognition.

  \item \emph{Decoder mode (generation).} For captioning and
        answer-generation tasks, the model acts as a text decoder
        conditioned on visual tokens. Image features from the ViT are
        injected, via cross-attention or prefix-style conditioning, into
        an autoregressive text decoder that predicts the next token
        given the image and previous text. This enables the same
        backbone to perform fluent image-conditioned generation.
\end{itemize}

BLIP is trained with a mixture of these objectives-image-text
contrastive learning, image-text matching, and image-conditioned
caption generation. The contrastive objective enforces global alignment
between image and text embeddings; the matching objective encourages
fine-grained pairwise discrimination; and the captioning objective
teaches the model to express visual content in natural language. This
combination leads to a joint representation space that can be probed
both discriminatively (for retrieval, VQA, or classification) and
generatively (for captioning and open-ended responses), making BLIP a
canonical example of a unified vision-language pretraining framework.

Despite its flexibility, BLIP still exhibits several practical
limitations when viewed against the backdrop of rapidly scaling
large language models. First, the multimodal encoder-decoder backbone
in BLIP is relatively modest in size compared with contemporary
large-scale LMs, and scaling it end-to-end is computationally expensive:
the vision and language components must be trained jointly, which limits
the ability to reuse very large, purely textual foundation models.
Second, BLIP couples the vision and language parameters tightly,
reducing modularity: adapting the system to a new language model
typically requires substantial retraining of the multimodal backbone.
Third, because the same transformer stack must serve as both encoder and
decoder, architectural choices are constrained by the need to balance
representation quality for understanding tasks with fluency and control
for generation.

BLIP-2~\cite{li2023blip2} is explicitly designed to address these
bottlenecks. It decouples the visual encoder and the language model,
keeping both as frozen, pre-trained backbones, and introduces a
lightweight querying transformer (Q-Former) as a bridge. This design
enables BLIP-2 to plug into powerful off-the-shelf LMs without
retraining them, drastically reduces multimodal training cost, and
restores modularity: the same image encoder and Q-Former can, in
principle, be paired with different language models while preserving the
benefits of large-scale text-only pretraining.

\paragraph{BLIP-2: frozen encoders with a querying transformer.}

BLIP-2 adopts an explicitly LM-centric design~\cite{li2023blip2}. Both
the visual encoder and the language backbone are strong, pre-trained,
\emph{frozen} models: the image encoder is typically a ViT or similar
large-scale vision backbone, and the language component is a
high-capacity LM (e.g., a T5- or GPT-style transformer). Rather than
fine-tuning these backbones jointly, BLIP-2 introduces a lightweight
\emph{Querying Transformer} (Q-Former) that serves as the primary
multimodal bridge.

\begin{figure}[ht]
    \centering
    \includegraphics[width=\linewidth]{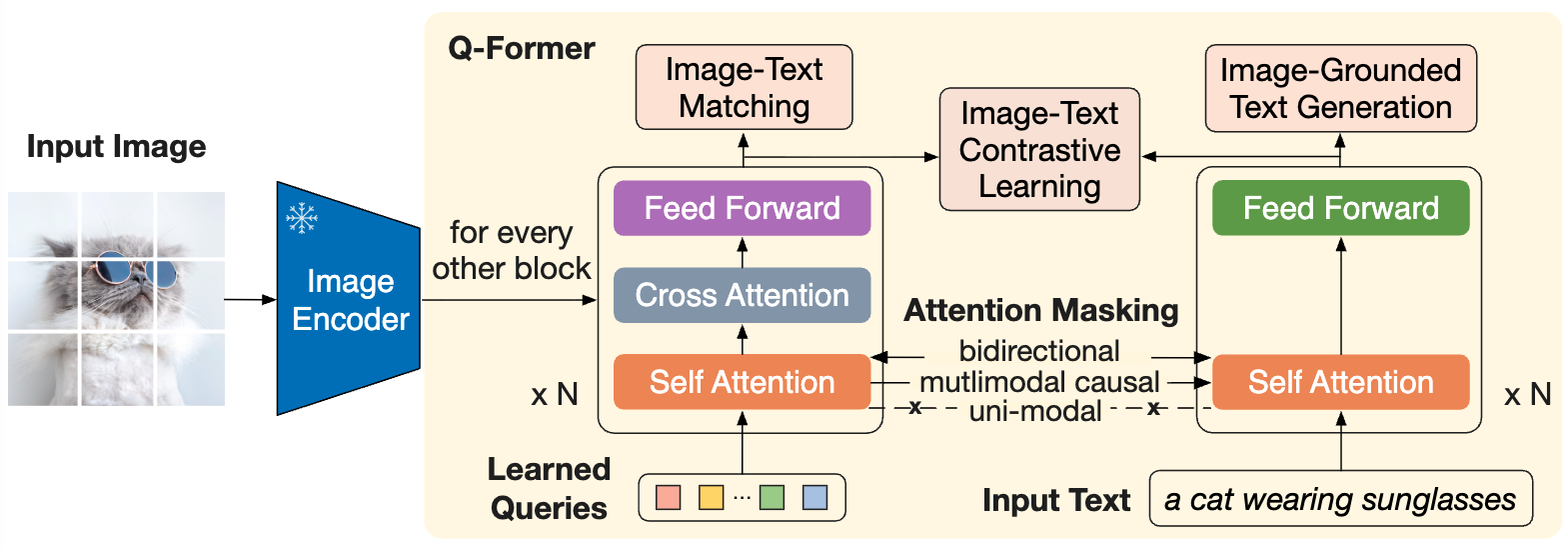}
    \caption{Illustration of the BLIP-2 architecture with a Querying
    Transformer (Q-Former)~\cite{li2023blip2}. A frozen Vision
    Transformer encodes the input image into a grid of patch features.
    A small transformer, the Q-Former, maintains a fixed set of
    learnable query tokens that attend to these frozen visual features
    via cross-attention, producing a compact sequence of
    image-conditioned query outputs. These outputs are then linearly
    projected into the language model’s embedding space and injected as
    prefix tokens to a frozen large language model, which performs
    image-conditioned generation. Only the Q-Former and projection
    layers are trained during multimodal adaptation, while both the
    image encoder and language model remain frozen.}
    \label{fig:qformer-arch}
\end{figure}

The Q-Former plays two key roles (Figure~\ref{fig:qformer-arch}):

\begin{itemize}
  \item \textbf{Query-based visual summarization.} The Q-Former
        maintains a small set of learnable query tokens (e.g., $32$ or
        $64$) that interact with the frozen image features through
        cross-attention. Each query attends to the entire set of visual
        tokens and aggregates information into a single embedding. After
        several transformer layers, the queries become a compact,
        image-conditioned representation that captures salient aspects
        of the scene.

  \item \textbf{Interface to the language model.} The final query
        embeddings are passed through a linear projection to match the
        language model’s hidden dimension. The projected embeddings are
        then inserted as prefix tokens (or special ``image tokens'') at
        the input of the frozen LM. From the LM’s perspective, these
        tokens are simply part of the input sequence, allowing standard
        autoregressive decoding to produce captions or answers
        conditioned on the image.
\end{itemize}

\begin{figure}[ht]
    \centering
    \includegraphics[width=\linewidth]{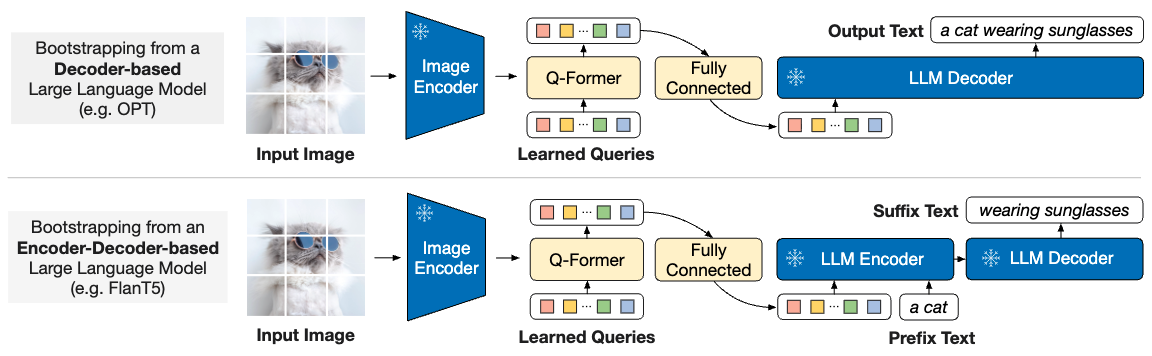}
    \caption{Bootstrapping BLIP-2 from different classes of large
    language models~\cite{li2023blip2}. \textbf{Top:} When using a
    decoder-only LM (e.g., OPT), the Q-Former consumes frozen image
    features and produces query outputs that are linearly projected and
    inserted as visual prefix tokens into the LM decoder, which then
    generates the full caption (e.g., ``a cat wearing sunglasses'').
    \textbf{Bottom:} When using an encoder-decoder LM (e.g., FlanT5),
    the projected query tokens are fed into the LM encoder together with
    textual prefix tokens (e.g., ``a cat''), while the decoder generates
    the suffix text (e.g., ``wearing sunglasses''). In both cases, the
    image encoder and LM remain frozen; only the Q-Former and projection
    layers are trained, enabling BLIP-2 to flexibly reuse a variety of
    large language models as the text backbone.}
    \label{fig:blip2-bootstrapping}
\end{figure}

BLIP-2 supports both decoder-only and encoder-decoder language models,
using the same Q-Former bridge but slightly different wiring, as
illustrated in Figure~\ref{fig:blip2-bootstrapping}.

Training proceeds in two main stages~\cite{li2023blip2}:

\begin{enumerate}
  \item \textbf{Vision-Q-Former pretraining.} In the first stage, the
        image encoder and Q-Former are trained together (with the LM
        absent or replaced by a lightweight text head) on
        image-text contrastive, image-text matching, and
        image-conditioned captioning objectives. The visual encoder is
        typically frozen or only lightly tuned, while the Q-Former
        learns to extract query embeddings that align well with textual
        descriptions.

  \item \textbf{Coupling to a frozen language model.} In the second
        stage, the pre-trained Q-Former is connected to a frozen
        large-scale LM via the projection layer. The combined system is
        then fine-tuned on image-conditioned language tasks (e.g.,
        captioning, VQA-style data, or multimodal instructions),
        updating primarily the Q-Former and projector. The LM itself
        remains unchanged, preserving its general-purpose linguistic and
        reasoning capabilities.
\end{enumerate}

BLIP-2 thus realizes a clear template of \emph{frozen backbones plus a
trainable bridge transformer}: a powerful, reusable vision encoder and
language model are linked by a relatively small Q-Former that distills
visual information into a token sequence that large language models can
consume efficiently.

\subsection{LM-Centric Few-Shot Modeling: Flamingo}
\label{subsec:flamingo-arch}

\begin{figure}[ht]
    \centering
    \includegraphics[width=\linewidth]{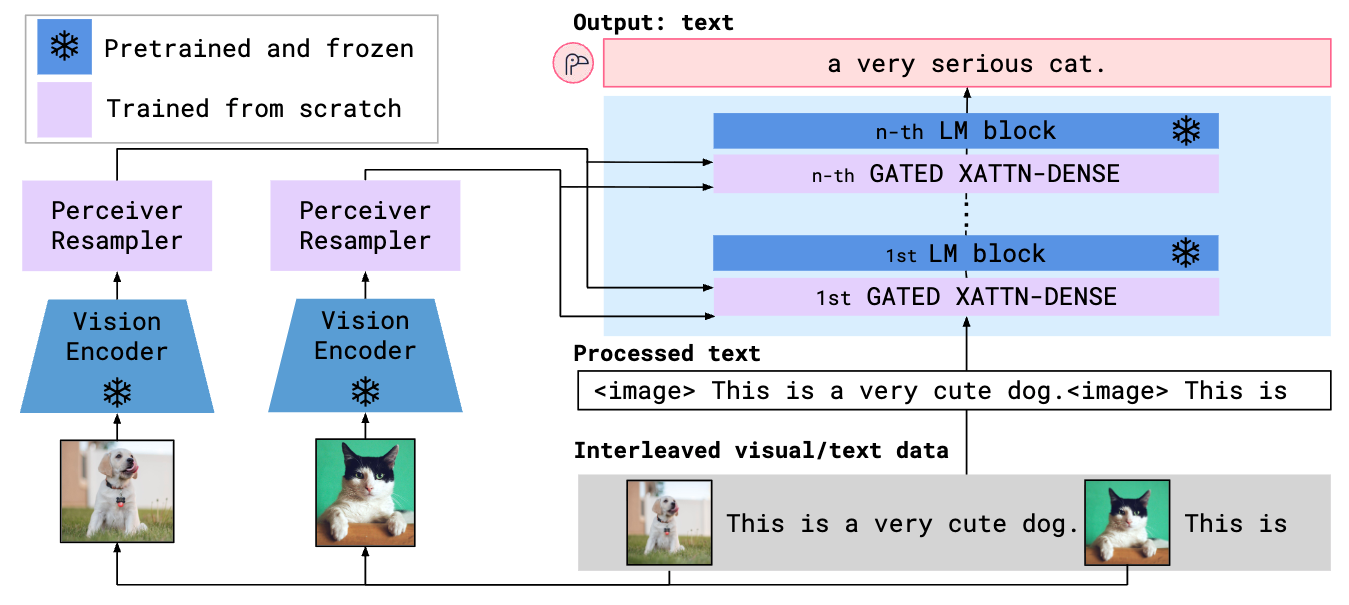}
    \caption{High-level Flamingo architecture~\cite{alayrac2022flamingo}.
    One or more images are first encoded by a frozen vision backbone,
    producing dense feature maps. A Perceiver Resampler-a small
    transformer trained from scratch-consumes these features and
    outputs a fixed-size set of latent visual tokens for each image,
    decoupling the number of visual tokens from the input resolution.
    These visual tokens are injected into selected layers of a frozen
    autoregressive language model via gated cross-attention blocks
    (``GATED XATTN-DENSE''). The LM processes interleaved visual/text
    data, represented as sequences such as
    \texttt{<image> This is a very cute dog. <image> This is ...}, and
    produces text-only outputs (e.g., captions or answers) conditioned
    on the entire multimodal context. Blue components are pre-trained
    and frozen; purple components are trained from scratch during
    Flamingo pretraining.}
    \label{fig:flamingo-arch}
\end{figure}

Flamingo~\cite{alayrac2022flamingo} is a representative LM-centric VLM
designed for few-shot learning on interleaved image-text sequences. It
demonstrates how a large, mostly frozen language model can be extended
to vision with minimal but carefully designed additions, as summarized
in Figure~\ref{fig:flamingo-arch}.

\paragraph{Visual encoder and Perceiver Resampler.}

As illustrated on the left side of
Figure~\ref{fig:flamingo-arch}, Flamingo employs a strong visual
backbone (e.g., a ConvNet or Vision Transformer pre-trained on image or
video data) to produce dense spatial feature maps for each frame or
image. These features can have high resolution and variable size, which
would be prohibitively expensive to feed directly into a language model.

To address this, Flamingo introduces a \emph{Perceiver Resampler}, a
compact transformer with latent query vectors and cross-attention. For
each input image, a fixed set of latent queries attends to the visual
feature map and iteratively aggregates information, yielding a
fixed-length set of image tokens regardless of the original resolution
(Figure~\ref{fig:flamingo-arch}, purple blocks above the vision
encoder). This resampling step decouples the LM’s context length from
the visual input resolution and provides a uniform interface for single
images, image sequences, or video frames.

\paragraph{Frozen LM with gated cross-attention.}

\begin{figure}[ht]
    \centering
    \includegraphics[width=\linewidth]{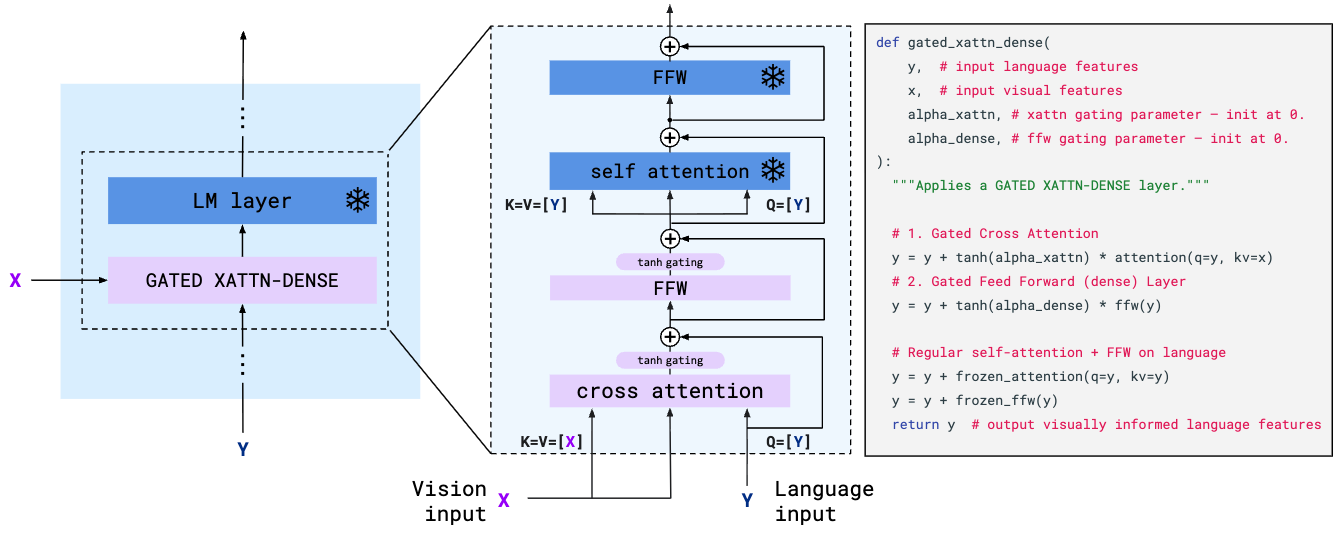}
    \caption{Internal structure of Flamingo’s gated cross-attention
    (GATED XATTN-DENSE) block~\cite{alayrac2022flamingo}. Given language
    features $Y$ and visual features $X$, the block first applies
    gated cross-attention, where the influence of visual features is
    scaled by a learnable parameter $\alpha_{\mathrm{xattn}}$ (via a
    $\tanh$ gate). This is followed by a gated feed-forward (dense)
    layer controlled by $\alpha_{\mathrm{dense}}$. The resulting
    visually informed language features are then passed to the frozen LM
    layer, which applies its usual self-attention and feed-forward
    network. Only the purple components (cross-attention, dense layers,
    and gates) are trained, while the blue LM sub-layers remain frozen,
    allowing Flamingo to inject visual information into selected layers
    without disturbing the pretrained language model.}
    \label{fig:flamingo-gated}
\end{figure}

The language component is a large, pre-trained autoregressive LM (the
blue “LM block” stack in Figure~\ref{fig:flamingo-arch}). Flamingo keeps
the LM weights frozen and augments a subset of its layers with
\emph{gated cross-attention} blocks, whose internal structure is shown
in detail in Figure~\ref{fig:flamingo-gated}. At each such layer, the
current textual hidden states $Y$ serve as queries and attend to the
visual tokens $X$ produced by the Perceiver Resampler. A learned scalar
gate $\alpha_{\mathrm{xattn}}$ modulates the contribution of the visual
features in the cross-attention output, and a second gate
$\alpha_{\mathrm{dense}}$ controls an additional feed-forward (dense)
layer. Only these cross-attention and dense components, together with
their gates, are trained; the original self-attention and feed-forward
sub-layers of the LM remain frozen.

\paragraph{Interleaved sequences and few-shot behavior.}

The bottom of Figure~\ref{fig:flamingo-arch} depicts the interleaved
visual/text input format used during training and inference. Flamingo is
trained on large corpora of sequences in which images and text snippets
co-occur (for example, dialogues with images or videos with commentary).
In the textual stream, images are represented by special tokens such as
\texttt{<image>}; at the same positions, the corresponding resampled
visual tokens are supplied to the gated cross-attention blocks. Because
the core LM remains an autoregressive transformer over this combined
sequence, Flamingo can perform few-shot learning across multimodal
examples in a single context: earlier image-text pairs act as in-context
demonstrations that condition the model’s responses to later images and
queries.

Architecturally, Flamingo exemplifies the template of a \emph{frozen
large language model augmented with learned visual adapters}. A powerful
vision encoder and Perceiver Resampler provide compact visual tokens;
gated cross-attention modules (Figures~\ref{fig:flamingo-arch}
and~\ref{fig:flamingo-gated}) inject these tokens into selected LM
layers; and the bulk of the LM parameters remain unchanged, preserving
general linguistic and reasoning skills while adding strong multimodal
capabilities.

\subsection{Open-Source Visual Assistants: LLaVA}
\label{subsec:llava-arch}

\begin{figure}[ht]
    \centering
    \includegraphics[width=0.8\linewidth]{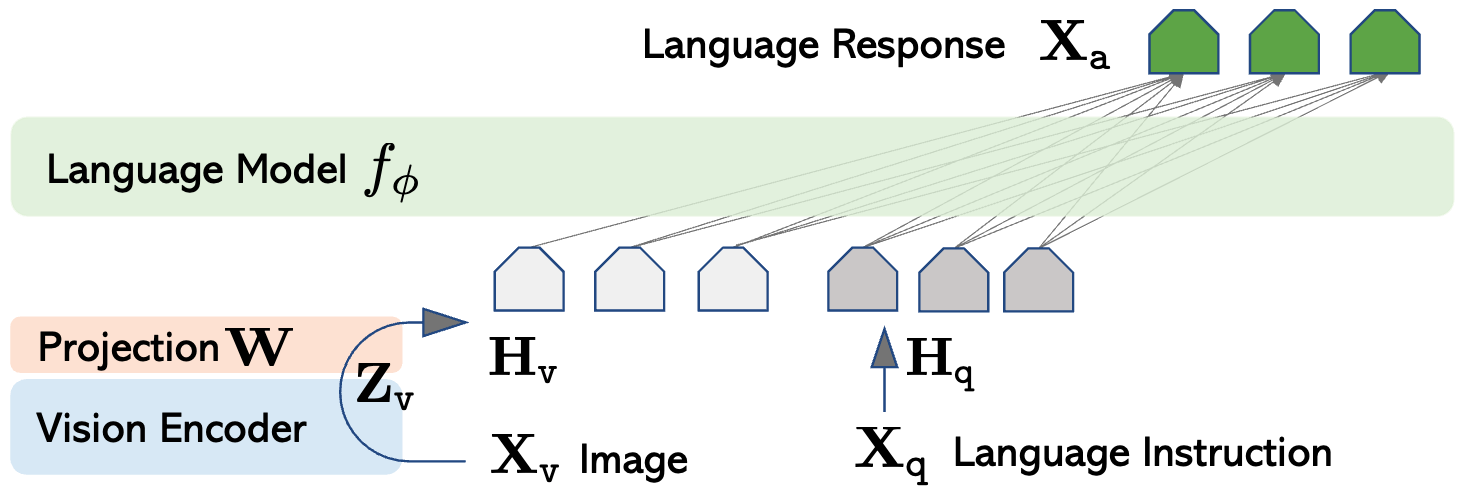}
    \caption{Schematic of the LLaVA architecture~\cite{liu2023visual}.
    An input image $X_v$ is encoded by a frozen CLIP Vision Transformer,
    producing visual features $Z_v$. A learned projection matrix
    $W$ (implemented as a small MLP) maps $Z_v$ into the language
    model’s hidden space, yielding a sequence of visual embeddings
    $H_v$. These visual tokens are concatenated with the hidden
    representations $H_q$ of a text instruction $X_q$ and passed into a
    frozen (or lightly fine-tuned) large language model $f_{\phi}$,
    which autoregressively generates an answer sequence $X_a$. Only the
    projection module and, in later stages, a subset of LM parameters
    are updated; the CLIP vision encoder and most of the language model
    remain frozen.}
    \label{fig:llava-arch}
\end{figure}

LLaVA (Large Language and Vision Assistant)~\cite{liu2023visual} shows
how a relatively simple architecture, combined with visual instruction
tuning, can yield a strong open-source multimodal assistant. At a high
level, LLaVA maps image features into the hidden space of a conversational
language model and treats them as additional prefix tokens, as
illustrated in Figure~\ref{fig:llava-arch}.

\paragraph{Components.}

As shown on the left side of Figure~\ref{fig:llava-arch}, LLaVA couples
three main components:
\begin{itemize}
  \item a frozen CLIP ViT-L/14 vision encoder that produces global and
        patch-level image embeddings $Z_v$~\cite{radford2021learning};
  \item a Vicuna language model (a fine-tuned LLaMA derivative) serving
        as the conversational backbone; and
  \item a small multi-layer perceptron (MLP) acting as a projection
        head $W$ that maps visual features into the Vicuna embedding
        space to form visual tokens $H_v$.
\end{itemize}
The image is thus represented as a short sequence of “visual words” in
the same vector space as text tokens.

\paragraph{Vision-language bridge and training.}

The vision–language bridge is deliberately simple. The projected image
embeddings $H_v$ are inserted as prefix tokens before the textual
instruction $X_q$ (Figure~\ref{fig:llava-arch}, middle). The resulting
sequence of visual and textual embeddings is fed into the language model
$f_{\phi}$, which generates a response $X_a$ autoregressively. From the
LM’s perspective, the visual tokens behave like additional context
tokens, so no architectural changes to the transformer are required.

Training is typically carried out in two stages~\cite{liu2023visual}:
\begin{enumerate}
  \item \textbf{Feature alignment.} With both the CLIP encoder and the
        language model frozen, the projection MLP $W$ is trained on
        image-caption pairs so that the LM can reconstruct captions
        conditioned on the projected visual tokens. This stage teaches
        the projector to produce LM-compatible visual embeddings.
  \item \textbf{Visual instruction tuning.} The full model (including
        $W$ and, in practice, a subset of LM parameters) is then
        fine-tuned on a large collection of multimodal
        instruction–response pairs, many generated by GPT-4. Prompts
        combine an image placeholder (whose position corresponds to
        $H_v$) and a natural-language instruction; the target is a
        detailed assistant-style answer. This stage aligns the model’s
        behavior with conversational user expectations while preserving
        much of the original LM’s linguistic competence.
\end{enumerate}

LLaVA therefore instantiates a minimal LM-centric template: a strong
frozen vision encoder, a powerful conversational language model, and a
shallow projector that converts image features into prefix tokens. Most
of the multimodal behavior arises not from architectural complexity but
from the visual instruction tuning regime built on top of this simple
interface.

\subsection{Industrial-Scale LVLMs: Qwen-VL}
\label{subsec:qwen-arch}

\begin{figure}[ht]
    \centering
    \includegraphics[width=0.9\linewidth]{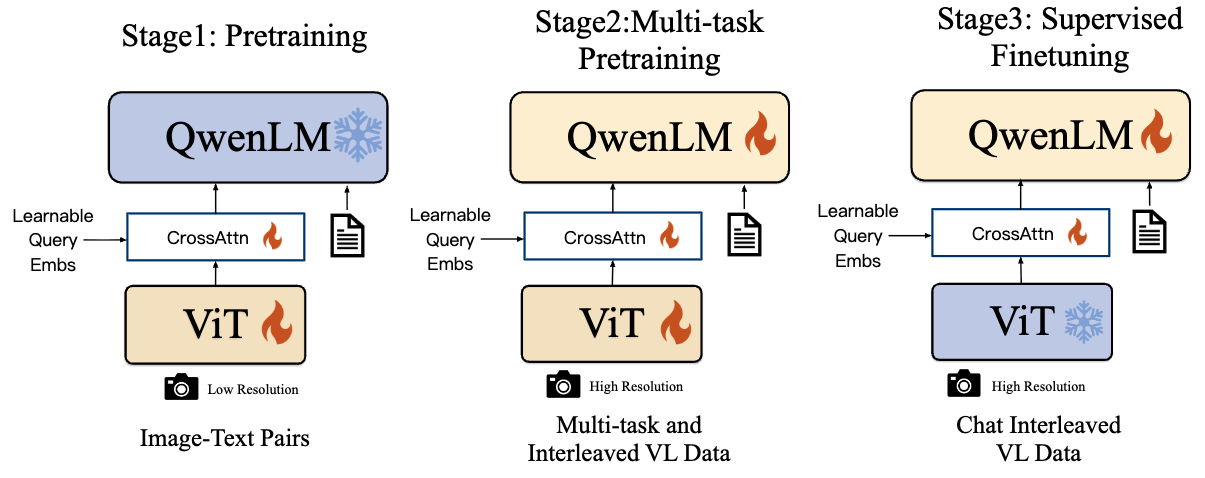}
    \caption{Three-stage training pipeline for Qwen-VL
    \cite{bai2023qwen}. \textbf{Stage 1: Pretraining.} A ViT-based
    vision encoder and a cross-attention module with learnable query
    embeddings are trained on large-scale image-text pairs, while the
    QwenLM remains frozen. \textbf{Stage 2: Multi-task pretraining.} On
    high-resolution and interleaved vision-language (VL) data, both the
    ViT and QwenLM are jointly optimized together with the cross-
    attention bridge, enabling stronger multimodal coupling. \textbf{Stage
    3: Supervised fine-tuning.} Using chat-style, interleaved VL data,
    the high-resolution ViT is frozen and the QwenLM plus cross-
    attention module are fine-tuned to follow multimodal instructions.}
    \label{fig:qwen-stages}
\end{figure}

The Qwen-VL series~\cite{bai2023qwen} scales the LM-centric VLM template
to industrial settings such as document understanding, UI grounding, and
tool-oriented assistants. Architecturally, it combines a strong
Vision Transformer backbone with a Qwen text LLM via cross-attention and
learnable query embeddings, and it is optimized through a multi-stage
training pipeline illustrated in Figure~\ref{fig:qwen-stages}.

\paragraph{Multilevel vision encoder.}

Qwen-VL employs ViT-style backbones that produce multi-scale feature
maps suitable for high-resolution inputs such as documents, charts, and
mobile UI screenshots. Later variants (e.g., Qwen2.5-VL) support dynamic
resolution and multi-frame processing, allowing the model to adaptively
trade off coverage and compute when handling long documents or short
video clips. Features from multiple layers are aggregated so that both
fine-grained details (e.g., small text regions) and global layout cues
are available to the language model.

\paragraph{Language backbone and structured outputs.}

The language core is a Qwen text LLM available at several parameter
scales. Visual features are transformed into query-like embeddings and
fed to the LLM through cross-attention, following the LM-centric pattern
seen in Flamingo and BLIP-2. On top of the base LM, Qwen-VL adds
task-specific decoding heads that can emit:
\begin{itemize}
  \item free-form natural language for captioning, open-ended VQA, or
        explanation;
  \item structured fields (often JSON-like) for information extraction
        and document understanding; and
  \item spatial outputs such as bounding boxes or points for grounding
        objects, UI elements, or regions in an image.
\end{itemize}
This design allows a single backbone to support diverse industrial
workflows, from OCR-style extraction to grounding-based tool calling.

\paragraph{Multi-stage training recipe.}

As summarized in Figure~\ref{fig:qwen-stages}, Qwen-VL is trained using
a three-stage pipeline:

\begin{enumerate}
  \item \textbf{Stage 1: Vision-to-language pretraining.} A ViT encoder
        and cross-attention bridge with learnable visual queries are
        trained on large-scale image-text pairs. The QwenLM is kept
        frozen, so this stage focuses on aligning visual features with
        the existing language representation space while controlling
        compute.

  \item \textbf{Stage 2: Multi-task multimodal pretraining.} On
        higher-resolution and interleaved VL data (including captioning,
        VQA, OCR, and grounding-style tasks), both the ViT and QwenLM
        are jointly optimized together with the cross-attention module.
        This stage strengthens multimodal coupling and improves
        performance on a wide range of benchmarks.

  \item \textbf{Stage 3: Supervised chat fine-tuning.} Finally, using
        curated chat-style, interleaved VL data, the high-resolution ViT
        is frozen and the QwenLM plus bridge module are fine-tuned as a
        multimodal assistant. This step focuses on instruction following
        and safety, aligning the model’s responses with user expectations
        while preserving the visual understanding acquired in previous
        stages.
\end{enumerate}

Overall, Qwen-VL extends the LLaVA-style LM-centric template with
richer, multi-resolution visual encoders, a cross-attention–based bridge
with learnable queries, structured output heads, and a carefully staged
training pipeline that separates representation learning from
assistant-style alignment.

\begin{didyouknow}
LVLMs like Qwen-VL are often engineered as much around
\emph{operations} as around model capacity. The three-stage pipeline is
not just a training trick, but a way to de-risk deployment: Stage~1
aligns vision to an already trusted language backbone, Stage~2 improves
raw multimodal capability under close monitoring, and Stage~3 layers on
chat-style behaviour and safety controls. This separation lets teams
iterate on assistant behaviour without constantly retraining the
high-resolution vision stack from scratch.
\end{didyouknow}

\subsection{Evolving Generalist Multimodal Foundation Models}
\label{subsec:generalist-arch}

Beyond task-specific or LM-centric designs, a growing family of systems
treat multimodality as a first-class objective at pretraining time and
aim to serve as \emph{generalist} foundation models. These models are
typically trained on large, heterogeneous corpora with interleaved text
and images (and sometimes video), and are evaluated across a broad range
of downstream tasks.

\paragraph{Kosmos-1.}
Kosmos-1~\cite{huang2023language} is trained as a multimodal large
language model on web-scale corpora containing interleaved text and
images. A vision transformer encodes images into visual tokens that are
fed directly into a unified transformer decoder alongside text tokens.
Unlike LM-centric approaches that retrofit a visual front-end onto a
frozen text-only LM, Kosmos-1 trains a single transformer jointly on
multimodal inputs from the outset, and is evaluated on captioning, VQA,
visual reasoning, and even zero-shot multimodal tasks such as exam-style
questions with diagrams.

\paragraph{PaLI and PaLI-X.}
PaLI and PaLI-X~\cite{chen2022pali,chen2024palix} adopt a T5-style
encoder-decoder architecture with a powerful ViT image encoder and a
multilingual text backbone. Visual tokens are included in the encoder
input, while the decoder generates text in more than one hundred
languages. Training mixes captioning, VQA, OCR, translation, and
cross-lingual objectives, yielding a multilingual, multimodal foundation
model that can, for example, read a chart in one language and answer
questions in another.

\paragraph{LLaVA-NeXT.}
Building on the original LLaVA design, LLaVA-NeXT extends the
``CLIP + projector + LM'' template with higher-resolution visual inputs,
richer visual instruction data, and support for multi-image and
short-video contexts~\cite{llava2024next,li2024llavanextinterleave}. The
architecture remains relatively simple - a strong vision encoder plus a
projection layer into an open-source LM - but the data scale and task
diversity are substantially expanded, leading to improved OCR,
commonsense reasoning, and open-ended dialogue performance.

\paragraph{Qwen2.5-VL and Qwen3-VL.}
The Qwen2.5-VL and Qwen3-VL generations extend the Qwen-VL
series~\cite{bai2023qwen,bai2025qwen25vl,bai2025qwen3vl} toward more
general-purpose LVLMs. Qwen2.5-VL introduces a dynamic-resolution ViT
with multi-scale features and efficient windowed attention, enabling
robust handling of long documents, charts, and videos. Qwen3-VL further
scales the language backbone, lengthens the multimodal context window,
and strengthens reasoning and tool-usage capabilities. Both retain the
core LM-centric pattern in which a Qwen text LM is coupled to a powerful
vision encoder via cross-attention and learnable query embeddings.

\paragraph{Ovis and Ovis 2.5.}
The Ovis series (Open VISion) focuses on compact yet capable multimodal
models, with particular emphasis on high-resolution understanding of
documents, charts, and UI screenshots~\cite{aidcai2025ovis25}.
Ovis 2.5 incorporates a native-resolution ViT that can process images at
their original (and possibly heterogeneous) resolutions and aligns its
visual representations with an LLM via lightweight adapters. Later
versions place additional focus on chain-of-thought and reflective
reasoning, yielding relatively small models that perform competitively
on OCR- and reasoning-intensive benchmarks.

\paragraph{InternVL.}
InternVL and its successors, culminating in InternVL3, scale the vision
backbone and multimodal pretraining to industrial regimes, combining
very large ViT-like encoders with a powerful LLM and multi-level feature
fusion~\cite{zhu2025internvl3}. The models are trained on diverse
image-text corpora and evaluated on scenarios ranging from everyday
photographs to industrial inspection images, GUI interaction, and
tool-augmented agents. Architecturally, InternVL sits between
PaLI-style joint pretraining and Qwen-style LM-centric designs: it
emphasizes a strong vision foundation while still treating the LLM as
the central reasoning component.

Taken together, these systems underscore that there is no single
``correct'' way to marry a visual encoder and a language model. Some
architectures emphasize encoder-decoder symmetry and joint multimodal
pretraining; others treat the language model as the core and wrap it
with increasingly sophisticated visual encoders, bridging modules, and
instruction-tuning regimes. This diversity of design patterns provides a
rich toolbox for future VLM research and application development.

\section{Multimodal Alignment Objectives}
\label{sec:alignment-objectives}

Architecture alone does not guarantee effective multimodal behavior. The
visual encoder and language model must be trained (or adapted) with
objectives that encourage meaningful alignment between their
representations. Several families of objectives are commonly used.

\subsection{Contrastive Alignment}
\label{subsec:contrastive-alignment}

Contrastive alignment objectives train the model to assign high
similarity to corresponding image-text pairs and low similarity to
mismatched pairs. Given a batch of $N$ images
$\{I_i\}_{i=1}^N$ and associated captions
$\{\mathbf{t}_i\}_{i=1}^N$, a visual encoder $f(\cdot)$ and a text
encoder $g(\cdot)$ produce $d$-dimensional embeddings
\[
  v_i = f(I_i) \in \mathbb{R}^d,
  \qquad
  u_i = g(\mathbf{t}_i) \in \mathbb{R}^d.
\]
In CLIP-style training~\cite{radford2021learning,jia2021scaling}, these
embeddings are $\ell_2$-normalized,
\[
  \hat{v}_i = \frac{v_i}{\|v_i\|}, \qquad
  \hat{u}_i = \frac{u_i}{\|u_i\|},
\]
and a similarity matrix $S \in \mathbb{R}^{N \times N}$ is formed using
cosine similarity scaled by a learned temperature parameter
$\tau > 0$:
\[
  S_{ij} = \frac{\hat{v}_i^\top \hat{u}_j}{\tau}.
\]

The image-to-text contrastive loss treats each row of $S$ as a
logit vector over captions and encourages the matching caption
$\mathbf{t}_i$ to be the most likely for image $I_i$:
\begin{equation}
  \mathcal{L}_{\text{I}\rightarrow\text{T}}
  = \frac{1}{N} \sum_{i=1}^N
    -\log
    \frac{\exp(S_{ii})}
         {\sum_{j=1}^N \exp(S_{ij})}.
  \label{eq:clip-i2t}
\end{equation}
Symmetrically, the text-to-image loss encourages $I_i$ to be the most
likely image for caption $\mathbf{t}_i$:
\begin{equation}
  \mathcal{L}_{\text{T}\rightarrow\text{I}}
  = \frac{1}{N} \sum_{i=1}^N
    -\log
    \frac{\exp(S_{ii})}
         {\sum_{j=1}^N \exp(S_{ji})}.
  \label{eq:clip-t2i}
\end{equation}
The final contrastive objective is usually the average of the two:
\begin{equation}
  \mathcal{L}_{\text{contrast}}
  = \frac{1}{2}
    \big(
      \mathcal{L}_{\text{I}\rightarrow\text{T}}
      + \mathcal{L}_{\text{T}\rightarrow\text{I}}
    \big).
  \label{eq:clip-loss}
\end{equation}

In this formulation, all other items in the batch implicitly act as
\emph{negative} examples for a given pair $(I_i, \mathbf{t}_i)$, making
the loss highly efficient: a single forward pass produces $N^2$
image–text comparisons. Extensions to this basic setup introduce
multiple positives per image (e.g., multiple captions), hard-negative
mining, or additional regularizers, but the core idea remains the same:
learn a shared embedding space in which aligned image-text pairs are
close and misaligned pairs are far apart.

Contrastive alignment can be applied purely at the embedding level,
without explicit language generation, which makes it computationally
attractive at scale. Many VLMs therefore use contrastive pretraining to
bootstrap a strong multimodal backbone and then adapt that backbone for
generative tasks via additional training or by coupling it to an
autoregressive LM, as in CLIP-based captioners, BLIP, and BLIP-2.

\begin{didyouknow}
A practical side effect of CLIP-style contrastive training is that
\emph{every} other sample in the batch implicitly becomes a negative
example. This means that larger batch sizes not only stabilize
optimization but also improve the quality of the learned embedding
space, because each step sees a richer set of image-text mismatches.
Many large-scale systems therefore invest heavily in distributed
training infrastructure primarily to support extremely large effective
batch sizes for contrastive objectives.
\end{didyouknow}

\subsection{Matching and Classification Losses}
\label{subsec:matching-losses}

For tasks such as Visual Question Answering (VQA), image-text matching,
or phrase grounding, it is common to use classification-style losses
that directly predict whether a given image-text pair matches, or which
candidate answer or region is correct.

\paragraph{Image-text matching.}
In image-text matching (ITM), the model receives an image $I$ and a
sentence $\mathbf{t}$ and must decide whether they describe one another.
Let $h(I,\mathbf{t})$ denote a pooled multimodal representation
(e.g., the \texttt{[CLS]} token from a cross-modal transformer as in
UNITER~\cite{chen2020uniter} or VinVL~\cite{zhang2021vinvl}). A binary
classifier with parameters $w \in \mathbb{R}^d$ and bias $b$ predicts
the probability that the pair matches:
\[
  p(y=1 \mid I,\mathbf{t})
  = \sigma\!\big(w^\top h(I,\mathbf{t}) + b\big),
\]
where $\sigma(\cdot)$ is the logistic sigmoid and $y \in \{0,1\}$
indicates whether $(I,\mathbf{t})$ is a positive or negative pair. The
ITM loss over a minibatch is then the standard binary cross-entropy:
\begin{equation}
  \mathcal{L}_{\text{ITM}}
  = - \frac{1}{N} \sum_{i=1}^N
    \Big(
      y_i \log p_i
      + (1 - y_i) \log (1 - p_i)
    \Big),
  \label{eq:itm-loss}
\end{equation}
with $p_i = p(y=1 \mid I_i,\mathbf{t}_i)$. In practice, positive pairs
are sampled from aligned image-caption data, and negatives are obtained
by pairing images with unrelated captions from the same minibatch.

\paragraph{Answer and region classification.}
For multiple-choice VQA, the model may produce a multimodal
representation $h(I,\mathbf{q},a_k)$ for each candidate answer $a_k$ to
question $\mathbf{q}$, and predict the correct index via a softmax:
\[
  s_k = w^\top h(I,\mathbf{q},a_k) + b,
  \quad
  p(a_k \mid I,\mathbf{q})
  = \frac{\exp(s_k)}{\sum_{k'} \exp(s_{k'})}.
\]
The loss is the categorical cross-entropy
\begin{equation}
  \mathcal{L}_{\text{VQA}}
  = - \frac{1}{N} \sum_{i=1}^N
      \log p(a_{k_i} \mid I_i,\mathbf{q}_i),
  \label{eq:vqa-loss}
\end{equation}
where $a_{k_i}$ is the ground-truth answer for example $i$.

For phrase grounding, given a phrase $\mathbf{p}$ and a set of region
proposals $\{r_j\}$ extracted by a detector, the model computes scores
$s_j = \phi(h(I,\mathbf{p},r_j))$ and applies a softmax over regions:
\[
  p(r_j \mid I,\mathbf{p})
  = \frac{\exp(s_j)}{\sum_{j'} \exp(s_{j'})},
\]
with a corresponding cross-entropy loss over the ground-truth region
index. VinVL~\cite{zhang2021vinvl}, for example, combines ITM with
region-level classification and regression objectives to refine both
detection and multimodal alignment.

These matching and classification objectives encourage fine-grained
alignment between local visual features (regions, patches) and specific
linguistic expressions (phrases, answers), complementing the more global
embedding-level alignment obtained from contrastive pretraining.

\begin{didyouknow}
In many VQA and grounding models, image-text matching and answer/region
classification are not just training losses but also act as implicit
\emph{curriculum}. Early in training, the model learns coarse ``does
this go together?'' signals from ITM; only later does it reliably
distinguish fine-grained answer options or overlapping region proposals.
Ablation studies in systems like UNITER and VinVL show that removing the
ITM objective often degrades not only retrieval performance but also
downstream VQA accuracy, underscoring how closely these objectives are
entangled in practice.
\end{didyouknow}

\subsection{Generative and Grounded Language Modeling}
\label{subsec:generative-grounded}

Generative objectives extend the basic language modeling framework to
condition on visual inputs, as in (\ref{eq:vlm-conditional}) from
Chapter~\ref{chap:lms-for-vlms}. Let
$\mathbf{t} = (t_1,\dots,t_n)$ denote a target text (caption, answer, or
explanation) associated with an image $I$ and visual tokens
$\mathbf{v} = (v_1,\dots,v_m)$. A conditional language model defines
\[
  p(\mathbf{t} \mid I)
  = p(\mathbf{t} \mid \mathbf{v})
  = \prod_{i=1}^n p(t_i \mid t_{<i}, \mathbf{v}),
\]
and is trained by minimizing the negative log-likelihood
\begin{equation}
  \mathcal{L}_{\text{LM}}
  = - \sum_{i=1}^n \log p(t_i \mid t_{<i}, \mathbf{v}).
  \label{eq:cond-lm-loss}
\end{equation}
In practice, $\mathbf{v}$ may enter the model as prefix tokens, via
cross-attention, or through an encoder-decoder interface (see
Chapter~\ref{chap:lms-for-vlms}). BLIP~\cite{li2022blip}, for instance,
uses a multimodal encoder-decoder transformer and optimizes both ITM
and caption-generation objectives, enabling the same backbone to support
retrieval-style understanding and fluent image-conditioned generation.

For \emph{grounded} language modeling, additional supervision enforces
consistency between specific words or spans and visual regions. Given a
set of aligned word–region pairs $(i, j)$, where word $t_i$ refers to
region $r_j$, the model may introduce auxiliary losses that encourage
the attention from $t_i$ to focus on the features of $r_j$, or directly
predict region indices conditioned on textual spans. Such losses can be
implemented as auxiliary cross-entropies over regions, or via
regularizers on attention maps.

Generative objectives thus serve two purposes. First, they teach the
model to integrate visual information into its internal state in a way
that supports next-token prediction. Second, when combined with
grounding signals and appropriate evaluation protocols, they encourage
the model to \emph{express} visually grounded facts faithfully in
natural language, helping to mitigate hallucinations that are not
supported by the input image.

\subsection{Instruction Tuning and Task Formatting}
\label{subsec:instruction-tuning}

Recent Vision-Language Models increasingly adopt \emph{instruction
tuning} regimes, in which the model is exposed to multimodal prompts
phrased as natural language instructions and trained to produce
appropriate responses. Concretely, training data are organized as
dialogue-style turns of the form
\begin{center}
  \fbox{%
    \parbox{0.9\linewidth}{%
      \textbf{User:} ``Describe this image.'' \\
      \textbf{Assistant:} ``A small dog is jumping over a log \dots''%
    }%
  }
\end{center}
optionally interleaved with one or more images. A single model may be
trained to handle prompts such as “Describe this image,” “What is
written on the sign?”, “Where is the red car located? (return a
bounding box),” or “Extract all key-value pairs from this receipt,”
each associated with a different output format (natural language,
coordinates, or structured JSON).

In the purely textual setting, instruction tuning and alignment with
human feedback have proved crucial for making large language models
behave in a user-aligned, helpful, and safe manner~\cite{ouyang2022training}.
Analogous approaches are now deployed at scale for VLMs such as
LLaVA-style assistants and Qwen-VL-family models, where the instructions
reference both images and text (e.g., “Given the following screenshot,
explain why the button is disabled”).

\paragraph{Unifying tasks via prompts.}

Instruction tuning serves first as a unifying interface across tasks.
Rather than training separate models for captioning, VQA, OCR, or
grounding, a single VLM is trained on a mixture of instruction–response
pairs where the \emph{prompt} implicitly specifies the task:
\begin{itemize}
  \item captioning-style prompts (``Describe this image in one
        sentence.'');
  \item question answering (``What is written on the sign?'', ``How
        many people are in the picture?'');
  \item grounding and detection (``Draw a box around the red car'',
        ``Return the coordinates of the submit button.'');
  \item structured extraction (``Extract all dates and total amount from
        this invoice as JSON.'').
\end{itemize}
The model learns to infer the intended task from the instruction
wording, while using the same underlying architecture and multimodal
representations. This greatly simplifies deployment: changing behavior
often requires only changing the prompt, not the model.

\paragraph{Formatting outputs and grounding behavior.}

A second, equally important role of instruction tuning is to align model
behavior with user expectations and downstream systems. During training,
each instruction is paired not only with images but also with outputs in
the \emph{desired format}: free-form explanations, short answers, boxed
coordinates, segmentation masks encoded as text, or machine-readable
structures such as JSON. By repeatedly seeing such examples, the VLM
learns conventions like:
\begin{itemize}
  \item responding concisely versus verbosely, depending on the prompt;
  \item including or omitting rationales (e.g., “Explain your reasoning”);
  \item emitting well-formed structured outputs (e.g., key–value lists,
        lists of bounding boxes) suitable for downstream tools.
\end{itemize}
In many systems, this stage also incorporates safety filters and
preference data, encouraging the model to avoid unsafe content and to
refuse or redirect when prompts are ambiguous or harmful.

\begin{didyouknow}
Multimodal instruction tuning quietly changes what ``grounding'' means in
practice. Instead of only learning to map images and text into a shared
representation space, the model also learns social and interface
conventions: how long an answer should be, when to say ``I don’t know'',
how to format bounding boxes or JSON so downstream tools can consume
them, and how to respond when an image contains sensitive content. In
many deployed LVLMs, this instruction layer has as much influence on
user experience as the underlying architecture or pretraining corpus.
\end{didyouknow}

\chapter{Datasets \& Evaluation Benchmarks}
\label{chap:datasets-benchmarks}

\textit{The behavior of a Vision-Language Model is shaped as much by its data as by its architecture and training objectives. The previous
chapters focused on how visual encoders, language models, and bridge
modules are designed and coupled, and on the losses used to align their
representations. In practice, however, these design choices interact
strongly with the datasets on which models are pre-trained, aligned, and
evaluated. Scale, domain coverage, annotation quality, and benchmark
design all play a central role in determining what a VLM can do - and
where it will fail.}

\section{Multimodal Data for Pretraining}
\label{sec:pretraining-data}

\subsection{Web-Scale Image-Text Corpora}

A defining trend in recent VLMs is the use of web-scale image-text
collections built from HTML alt-text, surrounding text, or other weak
metadata. ALIGN~\cite{jia2021scaling} demonstrated that a simple
dual-encoder trained on hundreds of millions of noisy image-alt-text
pairs could rival or surpass supervised ImageNet pretraining on many
downstream tasks, provided the corpus is large enough to compensate for
its noise.

\begin{figure}[ht]
    \centering
    \includegraphics[width=0.75\linewidth]{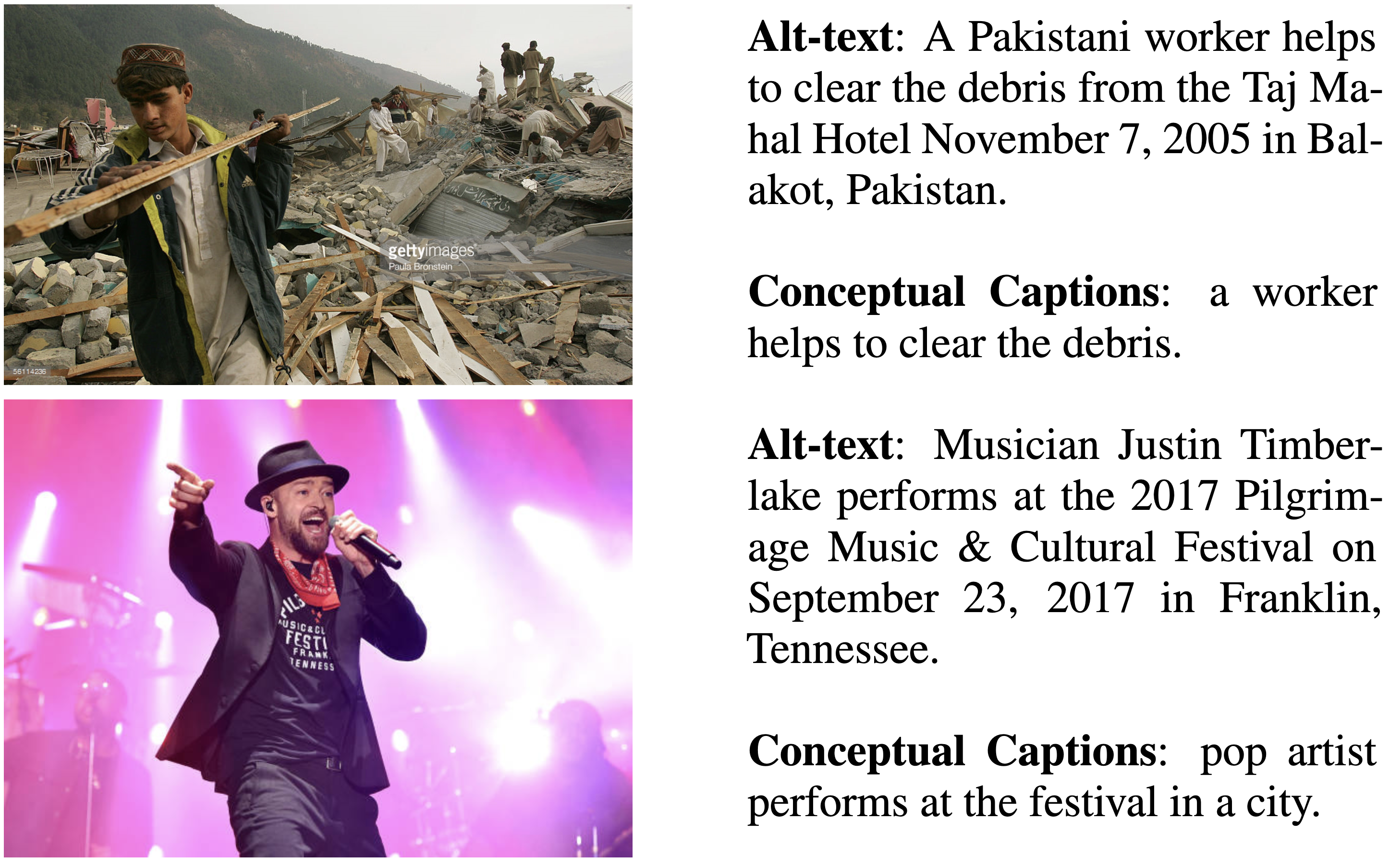}
    \caption{Illustration of the Conceptual Captions (CC3M) text
    normalization pipeline~\cite{sharma2018conceptual}. The original
    web alt-text (right, in bold) often contains detailed, sometimes
    personally identifiable metadata such as locations, dates, and
    celebrity names. CC3M transforms these into shorter, more generic
    captions (``Conceptual Captions'') that retain the visual semantics
    of the scene (e.g., ``a worker helps to clear the debris'', ``pop
    artist performs at the festival in a city'') while removing
    sensitive or overly specific information.}
    \label{fig:cc3m-examples}
\end{figure}

Conceptual Captions (CC3M)~\cite{sharma2018conceptual} follows a related
philosophy at smaller scale: captions are extracted from web alt-text
and titles, then aggressively cleaned, normalized, and hypernymed to
remove personally identifiable information and low-quality strings,
yielding $\sim$3.3M image-caption pairs suitable for training
captioning models (see Figure~\ref{fig:cc3m-examples}). A relaxed
version of this pipeline leads to Conceptual 12M
(CC12M)~\cite{changpinyo2021conceptual}, which trades some annotation
cleanliness for greater diversity and long-tail coverage, making it more
appropriate for large-scale representation learning.

\begin{figure}[ht]
    \centering
    \includegraphics[width=\linewidth]{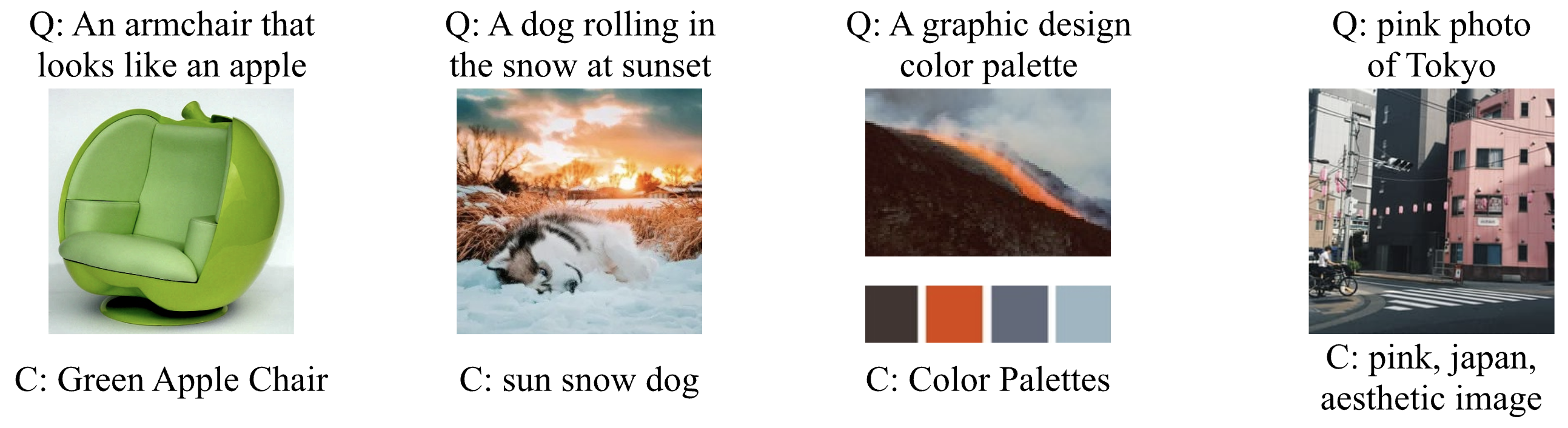}
    \caption{Examples from LAION-5B~\cite{schuhmann2022laion}. Each
    column shows a text query (\textbf{Q}) and the image with its
    caption (\textbf{C}) returned as the nearest neighbor in CLIP
    embedding space. The queries behave like natural user prompts (e.g.,
    ``An armchair that looks like an apple'', ``pink photo of Tokyo''),
    while the associated captions summarize the retrieved images.}
    \label{fig:laion-examples}
\end{figure}

Open datasets such as LAION-400M and LAION-5B push this paradigm to the
billion-example regime. LAION-5B~\cite{schuhmann2022laion} contains
$5.85$B CLIP-filtered image-text pairs collected from the web, with
associated CLIP embeddings and metadata. Figure~\ref{fig:laion-examples}
shows sample images retrieved by nearest-neighbor search in the CLIP
embedding space: for each user-style query (\textbf{Q}), the
corresponding image and caption (\textbf{C}) depict the top-ranked match
in the dataset. This illustrates both the diversity of the corpus and
the way CLIP-style features support large-scale retrieval.

Filtering with a pre-trained CLIP model helps remove mismatched captions
and non-photographic content, and additional detectors are applied for
watermarks, NSFW material, and toxic content. These large-scale, weakly
supervised corpora underlie many contemporary contrastive and
generative vision-language models, including CLIP-like encoders and
text-guided diffusion models.

From a VLM perspective, web-scale corpora provide broad coverage of
everyday and long-tail concepts, styles, and domains. However, they also
inherit the biases, stereotypes, and geographic skew of public web
content, and their weak supervision can encourage models to rely on
spurious correlations unless complemented by more carefully curated
data.

\begin{didyouknow}
Web-scale image-text corpora such as CC12M and LAION-5B are often
dominated by a surprisingly small set of visual and linguistic patterns:
stock photos, product shots, Western-centric scenes, and meme-style
captions. As a result, models trained purely on these datasets can
appear very capable in everyday internet settings yet behave
unpredictably on documents, scientific figures, non-Western locales, or
specialized professional imagery. Many strong VLMs therefore rely on a second stage of training on smaller, carefully curated
datasets to ``rebalance'' what they learned from the raw web.
\end{didyouknow}

\subsection{Region-Level and Structured Annotations}

\begin{figure}[ht]
    \centering
    \includegraphics[width=0.9\linewidth]{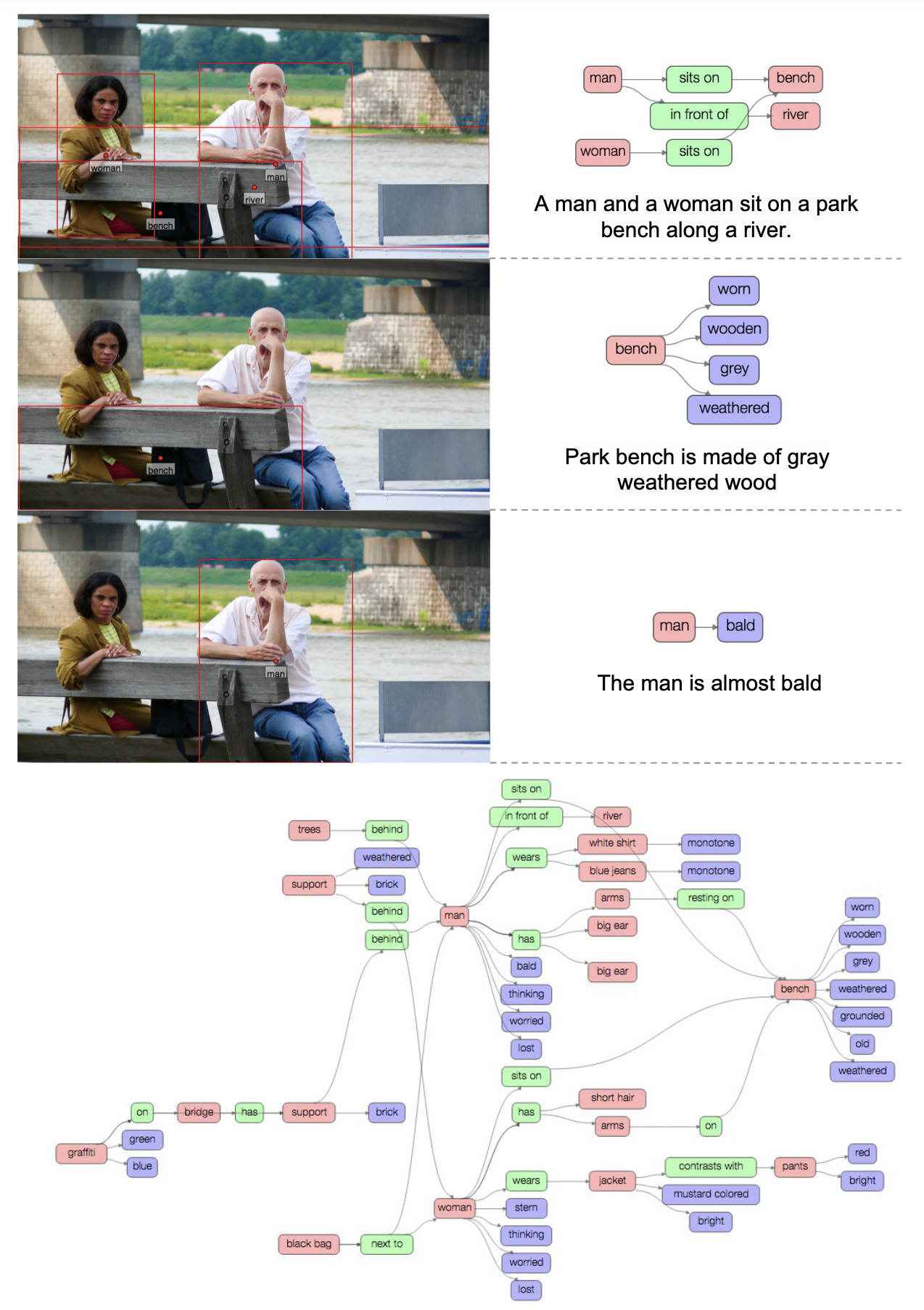}
    \caption{Example from Visual Genome~\cite{krishna2017visual}. The
    image is annotated with object bounding boxes (left), region-level
    captions describing localized parts of the scene (middle), and a
    scene graph (bottom) whose nodes denote objects and attributes and
    whose edges encode labeled relations (e.g., \emph{man sits on
    bench}, \emph{bench in front of river}).}
    \label{fig:vg-example}
\end{figure}

Several pretraining regimes benefit from datasets that provide
fine-grained alignment between localized visual regions and structured
linguistic annotations. Visual Genome~\cite{krishna2017visual} offers
dense scene annotations over $\sim$$108$k images, including object
bounding boxes, attributes, region-level descriptions, relationships,
and question-answer pairs. As illustrated in
Figure~\ref{fig:vg-example}, each image is accompanied by multiple
region crops, localized captions, and a scene-graph-style representation
linking objects (e.g., \emph{man}, \emph{bench}, \emph{river}) via
labeled relations (e.g., \emph{sits on}, \emph{in front of}). This
supports tasks such as region captioning, scene graph prediction, and
phrase grounding.

\begin{figure}[ht]
    \centering
    \includegraphics[width=\linewidth]{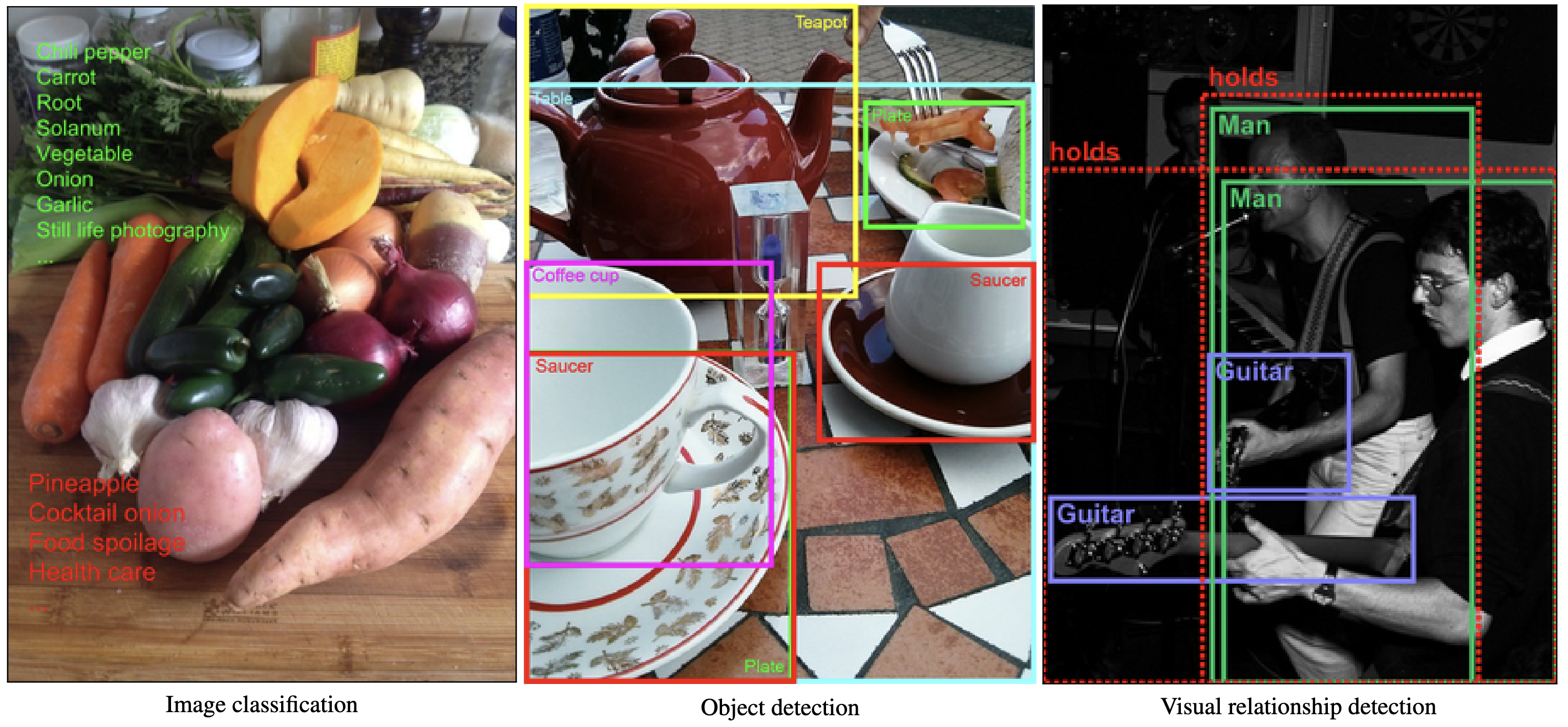}
    \caption{Examples from the Open Images dataset~\cite{kuznetsova2020openimages}.
    Left: image-level labels for multi-label classification. Middle:
    object detection annotations with category-specific bounding boxes.
    Right: visual relationship detection annotations, where pairs of
    boxes (e.g., \emph{man}, \emph{guitar}) are linked by relation
    labels (e.g., \emph{holds}).}
    \label{fig:openimages-example}
\end{figure}

Open Images~\cite{kuznetsova2020openimages} supplies millions of images
with image-level labels, bounding boxes, instance masks, and visual
relations. Figure~\ref{fig:openimages-example} shows representative
examples for image classification, object detection, and visual
relationship detection, highlighting the breadth of supervision
available in a single dataset. Such rich annotations enable the training
of strong detection and segmentation backbones that can later be
integrated into region-based VLM pipelines.

\begin{didyouknow}
Region-level datasets such as Visual Genome and Open Images are often
used not just to train better detectors, but also as a kind of
\emph{``alignment scaffold”} for multimodal models. By forcing the model
to tie individual words or phrases to specific boxes and relations, they
help disentangle visual grounding from pure language priors. Ablations
in many VLMs show that removing this kind of structured supervision
often leaves models fluent but noticeably less precise about \emph{where}
things are and \emph{which} object a phrase refers to.
\end{didyouknow}

\subsection{Video and Temporal Multimodal Data}

\begin{figure}[ht]
    \centering
    \includegraphics[width=\linewidth]{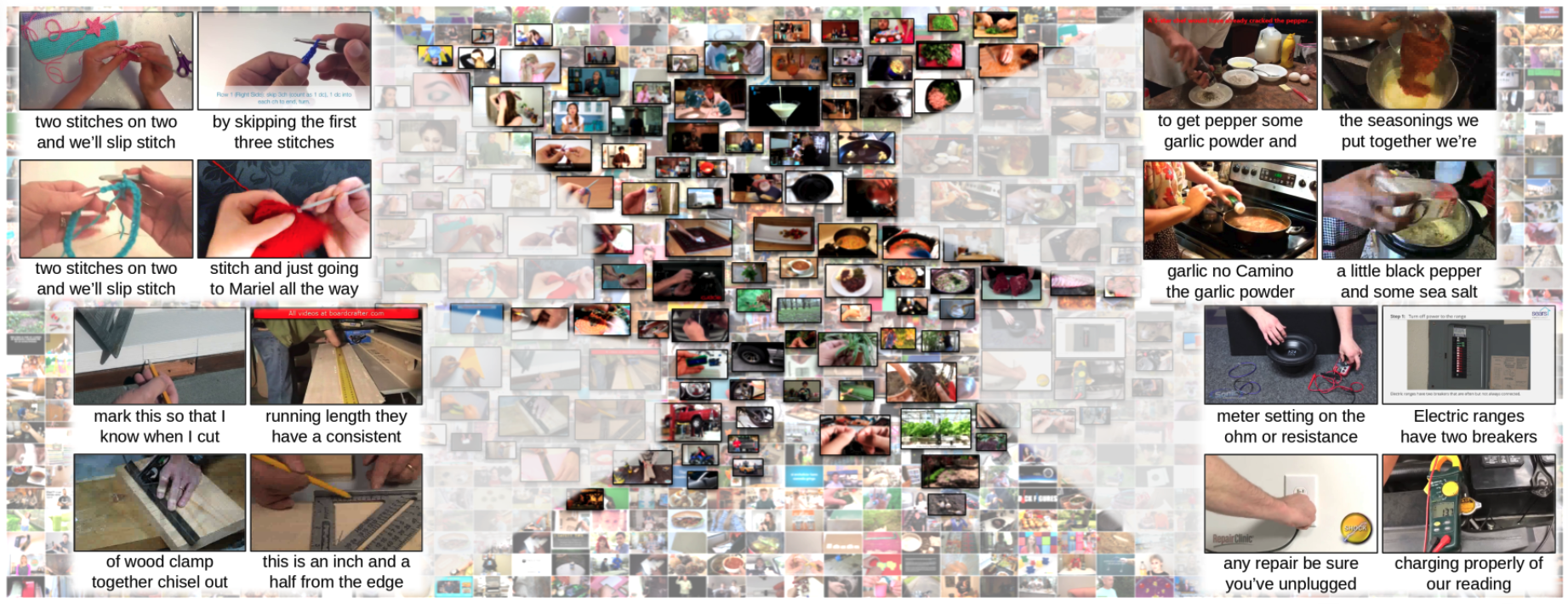}
    \caption{Examples of clip-caption pairs from
    HowTo100M~\cite{miech2019howto100m}. Short clips from narrated
    instructional videos are aligned with automatically transcribed
    speech segments. Pairs are retrieved using a joint video-text
    embedding and arranged into clusters corresponding to coherent
    activities (e.g., knitting, woodworking, cooking, electrical
    maintenance), illustrating the diversity of procedural content.}
    \label{fig:howto100m-examples}
\end{figure}

Static image-text pairs capture only a snapshot of the visual world.
To model temporal dynamics, narrated video corpora have become an
important ingredient in multimodal pretraining.
HowTo100M~\cite{miech2019howto100m} comprises roughly $136$ million video
clips extracted from $1.2$M narrated instructional videos on YouTube,
paired with automatically transcribed speech. The dataset emphasizes
step-by-step human activities and procedural tasks, making it a natural
source for learning grounded action representations and text-video
embeddings. Figure~\ref{fig:howto100m-examples} shows example
clip - caption pairs retrieved using the joint video - text embedding,
revealing clusters of semantically related activities such as knitting,
woodworking, cooking, and electrical maintenance.

\begin{figure}[ht]
    \centering
    \includegraphics[width=\linewidth]{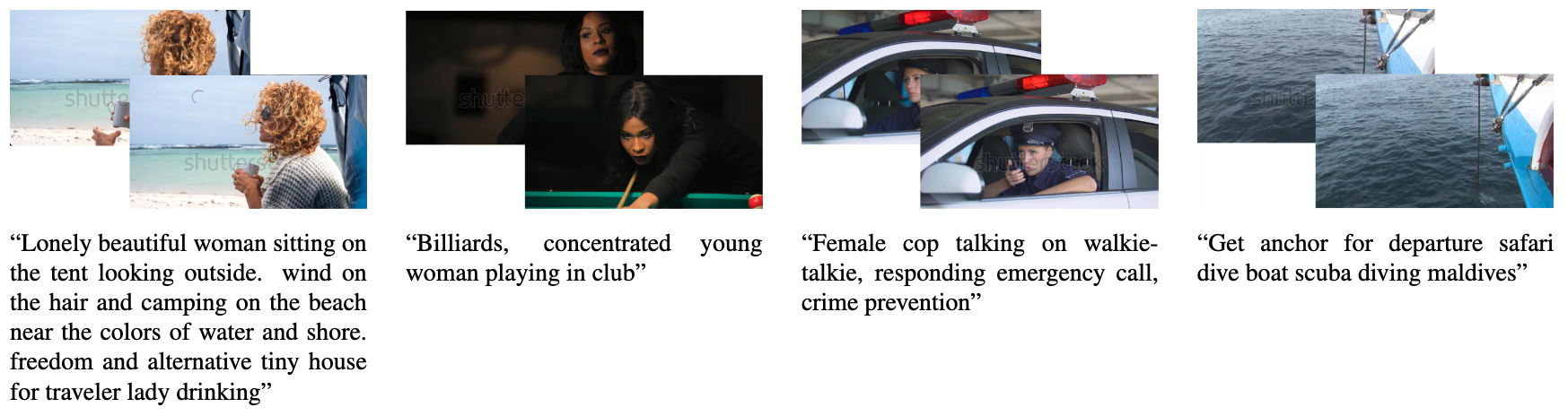}
    \caption{Example video-caption pairs from the WebVid dataset
    (WebVid2M)~\cite{bain2021frozen}. For each clip, two randomly
    sampled frames are shown together with the associated caption.
    Captions vary in style: some are long and descriptive, others are
    succinct, and some resemble keyword strings or include specific
    place names (e.g., ``maldives''), reflecting the heterogeneity of
    web video metadata.}
    \label{fig:webvid-examples}
\end{figure}

More recent datasets such as WebVid~\cite{bain2021frozen} scale curated
text-video pairs to tens of millions of short clips with concise
descriptions, and smaller benchmarks like MSR-VTT~\cite{xu2016msrvtt}
serve as evaluation targets for retrieval and captioning. As illustrated
in Figure~\ref{fig:webvid-examples}, WebVid captions exhibit a range of
styles - from long, loosely structured descriptions to short keyword-like
phrases, sometimes mentioning specific locations - providing a realistic,
noisy supervision signal for text-video models.

For VLMs, video pretraining offers two main benefits. First, it exposes
models to motion patterns and temporal context that cannot be inferred
from still images alone. Second, it provides a richer form of weak
supervision: narrations and subtitles often describe actions, goals, and
object interactions that are underrepresented in static alt-text. Many
modern architectures reuse the same language backbone for both
image-based and video-based tasks, relying on specialized visual
encoders (e.g., 3D CNNs or time-augmented ViTs) to provide temporal
tokens.

\begin{didyouknow}
Narrated video corpora like HowTo100M are one reason modern VLMs can
handle ``how-to'' style questions surprisingly well. Even though the
transcripts are noisy and loosely aligned in time, the constant pairing
of actions with natural language phrases (``stir the mixture’’, ``tighten
the screw’’) gives models a kind of weak procedural grounding that is
almost impossible to obtain from static images alone. As a result, some
LVLMs trained on video data can describe \emph{what will likely happen
next} in a clip, not just what is visible in a single frame.
\end{didyouknow}

\subsection{Multilingual and Domain-Specific Corpora}

\begin{figure}[ht]
    \centering
    \includegraphics[width=\linewidth]{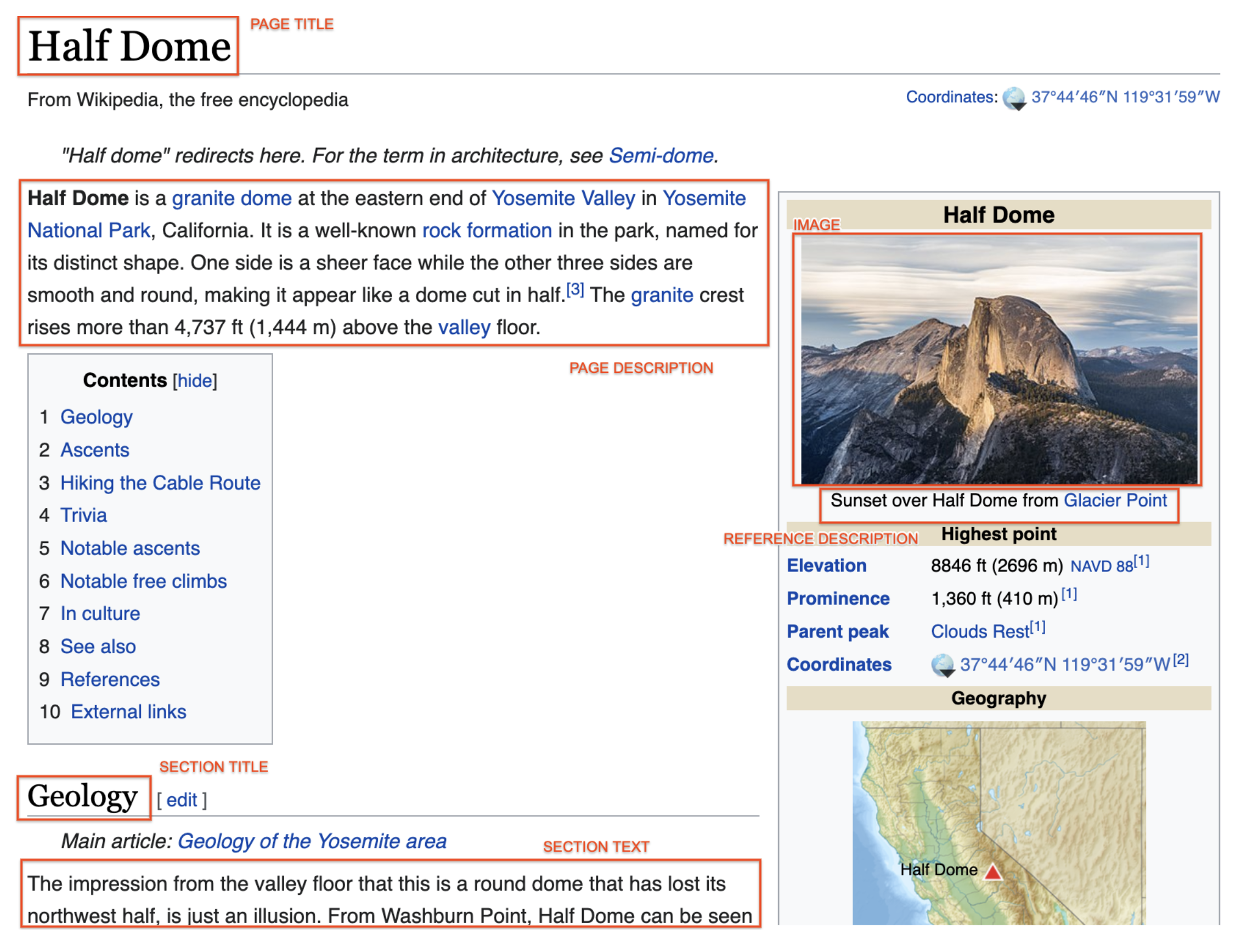}
    \caption{Example entry from the WIT dataset~\cite{srinivasan2021wit},
    based on the Wikipedia page for \emph{Half Dome}. The figure
    highlights the different textual fields extracted for each image:
    page title, lead paragraph, section titles, image caption, and
    reference description. WIT stores these fields (often in multiple
    languages) alongside the associated image, providing rich,
    structured supervision for multilingual multimodal pretraining.}
    \label{fig:wit-example}
\end{figure}

Finally, there is growing interest in pretraining VLMs on data that
extends beyond English web photographs. The WIT (Wikipedia-based Image
Text) dataset~\cite{srinivasan2021wit} constructs $\sim$37.6M
image-text examples from Wikipedia across 108 languages, pairing images
with multiple textual fields (captions, alt-text, surrounding
paragraphs, and other structured metadata). As illustrated in
Figure~\ref{fig:wit-example}, a single Wikipedia page (here, for \emph{Half
Dome}) yields several aligned text snippets: the page title, lead
paragraph, section titles, image captions, and reference descriptions.
This design enables multimodal pretraining that is both multilingual and
grounded in encyclopedic knowledge. Smaller datasets such as Multi30K
and its extensions provide carefully curated multilingual captions for
Flickr-style images, supporting research on cross-lingual grounding.

\begin{figure}[ht]
    \centering
    \includegraphics[width=\linewidth]{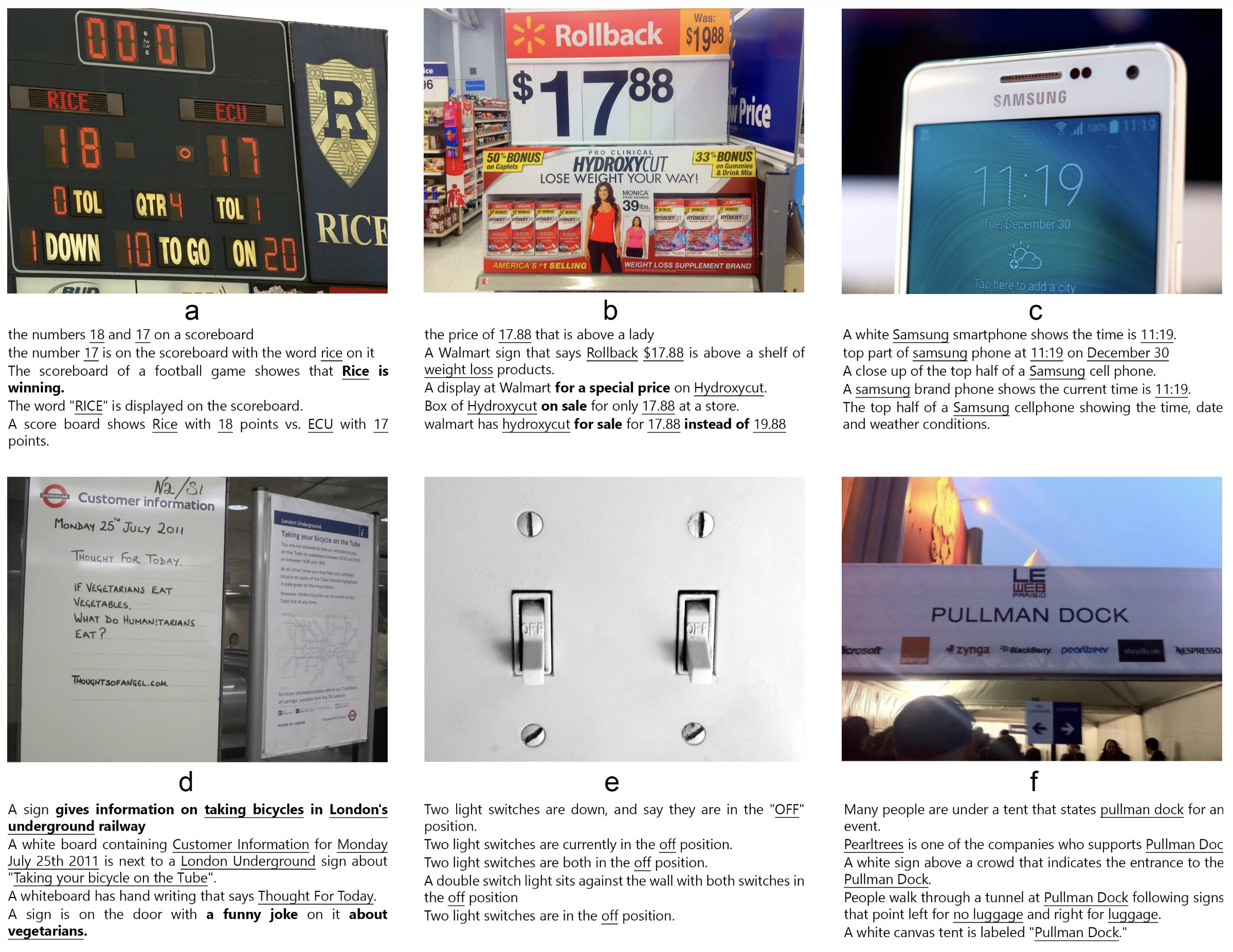}
    \caption{Illustrative examples from TextCaps~\cite{sidorov2020textcaps}.
    Each image is paired with multiple captions that explicitly refer to
    the text visible in the scene (e.g., digits on displays, brand
    names, slogans). Some caption tokens directly copy text from the
    image, while others paraphrase it or add inferred information,
    making the dataset a strong benchmark for text-aware visual
    captioning.}
    \label{fig:textcaps-example}
\end{figure}

Domain-specific corpora target particular application areas. Document
image datasets like RVL-CDIP~\cite{harley2015rvlcdip} and
PubLayNet~\cite{zhong2019publaynet} provide large collections of scanned
or born-digital pages with layout and category annotations, which are
widely used to pretrain encoders for document understanding and
OCR-centric VLMs. In the text-centric vision domain, TextCaps~\cite{sidorov2020textcaps}
focuses on captioning that requires reading and interpreting text in
images. As shown in Figure~\ref{fig:textcaps-example}, images contain
prominent textual content (e.g., product labels, signage), and each
image is accompanied by several human-written captions. Some words are
copied directly from the image, while others paraphrase or infer meaning
beyond the raw text (e.g., describing what the sign implies rather than
only what it says), encouraging models to integrate reading with
semantic understanding.

Multilingual and domain-specific data are particularly important for
deploying VLMs in non-English regions and specialized settings such as
finance, healthcare, or scientific publishing. In practice, large-scale
pretraining pipelines often combine web-scale generic corpora (e.g.,
LAION-5B, CC12M, HowTo100M) with smaller but more curated datasets that
inject the desired linguistic diversity or domain coverage.

\section{Supervised Datasets for Vision-Language Tasks}
\label{sec:supervised-vl-datasets}

Supervised datasets provide the backbone for training and evaluating
Vision-Language Models on concrete downstream tasks. Unlike the
largely weakly supervised corpora used for pretraining, these datasets
typically offer higher-quality annotations, well-defined task
specifications, and standardized evaluation protocols. This section
surveys key resources for several canonical tasks.

\subsection{Image Captioning}
\label{subsec:caption-datasets}

Image captioning datasets pair photographs with natural-language
descriptions and are widely used both for training generative VLMs and
for benchmarking grounded language generation.

\begin{figure}[ht]
    \centering
    \includegraphics[width=0.8\linewidth]{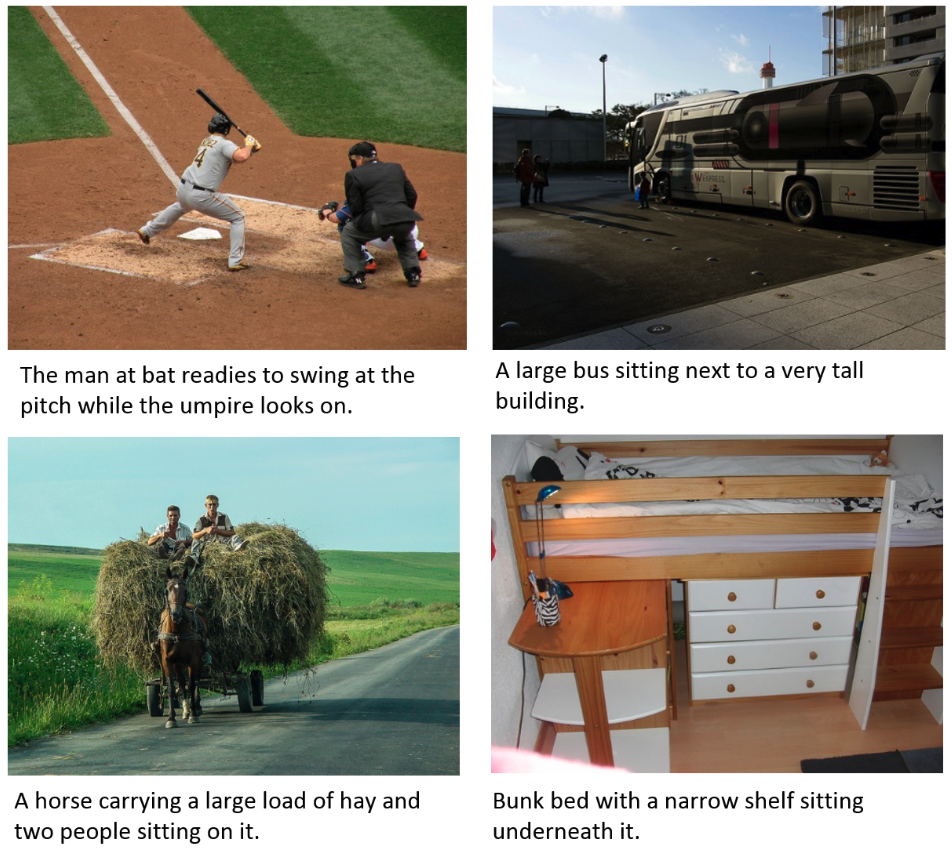}
    \caption{Example image-caption pairs from the MS COCO Captions
    benchmark~\cite{lin2014coco}. Each photograph is associated with
    multiple human-written captions that describe the main entities and
    activities in the scene (e.g., a batter preparing to swing, a bus
    parked beside a tall building, a horse-drawn cart loaded with hay,
    a bunk bed and furniture in a bedroom). These diverse, free-form
    descriptions provide rich supervision for training and evaluating
    image captioning models.}
    \label{fig:coco-captions}
\end{figure}

\paragraph{MS COCO Captions.}
The MS COCO dataset~\cite{lin2014coco} underlies many image captioning
benchmarks. The standard caption split (often referred to as COCO
Captions) contains roughly $123$k images, each annotated with five
independent captions written by crowdworkers. As illustrated in
Figure~\ref{fig:coco-captions}, images typically depict complex
everyday scenes with multiple objects and interactions, and the
captions describe salient entities, activities, and spatial relations.
Typical evaluation follows the Karpathy split protocol, which defines
train/validation/test partitions and uses automatic metrics such as
BLEU, METEOR, ROUGE-L, CIDEr, and SPICE to compare generated captions
against reference descriptions.

\begin{figure}[ht]
    \centering
    \includegraphics[width=\linewidth]{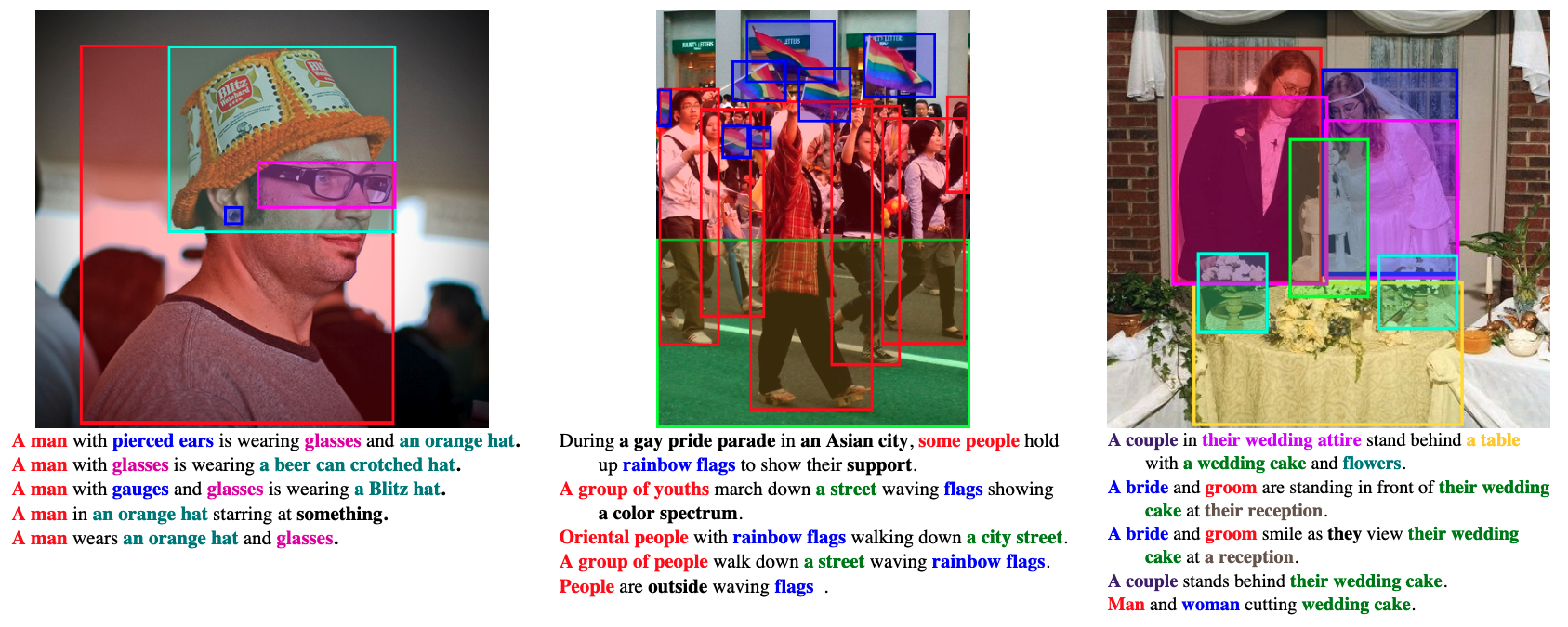}
    \caption{Examples from Flickr30k Entities~\cite{plummer2015flickrentities}.
    Each image is annotated with bounding boxes around salient entities
    (e.g., \textcolor{red}{\textbf{man}}, \textcolor{blue}{\textbf{glasses}},
    \textcolor{green}{\textbf{wedding cake}}), and corresponding noun
    phrases in the captions are color-coded to match these regions. This
    phrase-to-region supervision supports training and evaluation of
    models that must ground textual mentions to specific objects in the
    image.}
    \label{fig:flickr30k-entities}
\end{figure}

\paragraph{Flickr30k and Flickr30k Entities.}
Flickr30k~\cite{young2014flickr30k} consists of 31k Flickr images with
five crowd-sourced captions each, focusing on people-centric everyday
activities. Compared to COCO, the images are fewer but the captions are
often longer and more descriptive, making the dataset popular for both
captioning and image-text retrieval (see
Section~\ref{subsec:retrieval-datasets}). Flickr30k Entities further
augments this corpus with phrase-to-region correspondences, linking
noun phrases in the captions to bounding boxes in the image. As
illustrated in Figure~\ref{fig:flickr30k-entities}, different phrases
within a caption are color-coded and associated with specific regions,
which enables fine-grained evaluation of phrase grounding and
region-level alignment.

\paragraph{Novel object and text-aware captioning datasets.}
To test a model’s ability to describe objects unseen in caption
training data, nocaps~\cite{agrawal2019nocaps} draws images from the
Open Images validation and test sets, focusing on $\sim$400 object
categories that are rare or absent in COCO captions. Models are
trained on COCO plus Open Images labels and evaluated on their
ability to describe these novel objects in context.

\subsection{Visual Question Answering}
\label{subsec:vqa-datasets}

Visual Question Answering (VQA) datasets pair images with natural
language questions and short answers, providing a testbed for
multimodal reasoning.

\begin{figure}[ht]
    \centering
    \includegraphics[width=\linewidth]{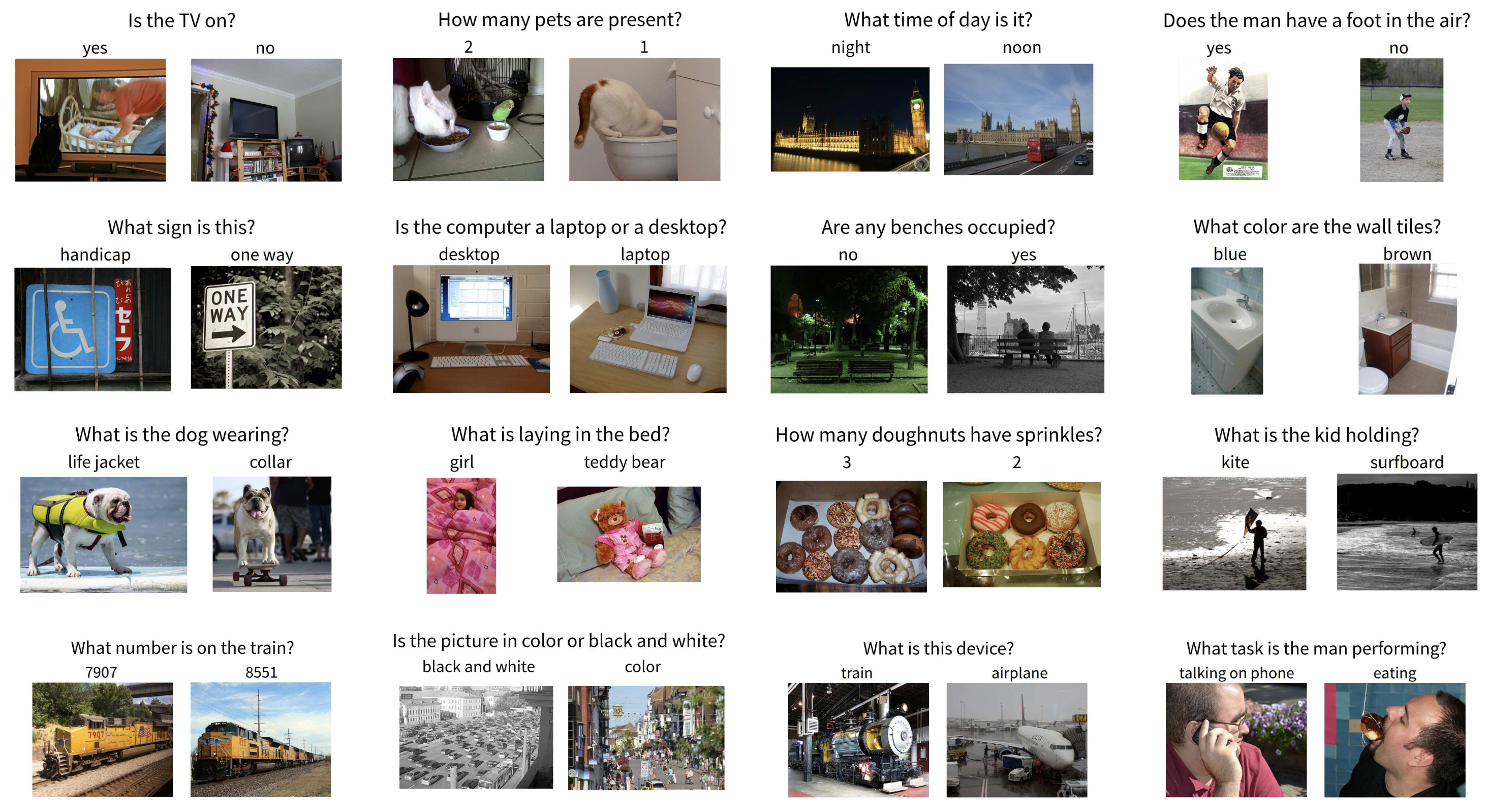}
    \caption{Example question-image-answer triplets from the balanced
    VQA v2 dataset~\cite{goyal2017vqa2}. For each natural-language
    question (e.g., ``Is the TV on?'', ``How many pets are present?''),
    the dataset includes corresponding images and short answers. Many
    questions are paired with visually similar images that require
    different answers, encouraging models to rely on the visual signal
    rather than dataset-specific language shortcuts.}
    \label{fig:vqa2-examples}
\end{figure}

\paragraph{VQA v2.}
The VQA v2 dataset~\cite{goyal2017vqa2} builds on the original VQA
benchmark by explicitly balancing question-answer pairs to reduce
language priors. For many questions, the dataset provides two similar
images that yield different answers, forcing models to attend to visual
content rather than exploiting superficial textual cues. As illustrated
in Figure~\ref{fig:vqa2-examples}, questions range from yes/no and
attribute queries (e.g., ``Is the TV on?'', ``What color are the wall
tiles?'') to counting (``How many doughnuts have sprinkles?'') and
object recognition (``What is this device?''). Answers are typically
single words or short phrases drawn from a restricted vocabulary,
making VQA v2 a standard benchmark for evaluating fine-grained
vision–language reasoning.

\begin{figure}[ht]
    \centering
    \includegraphics[width=0.6\linewidth]{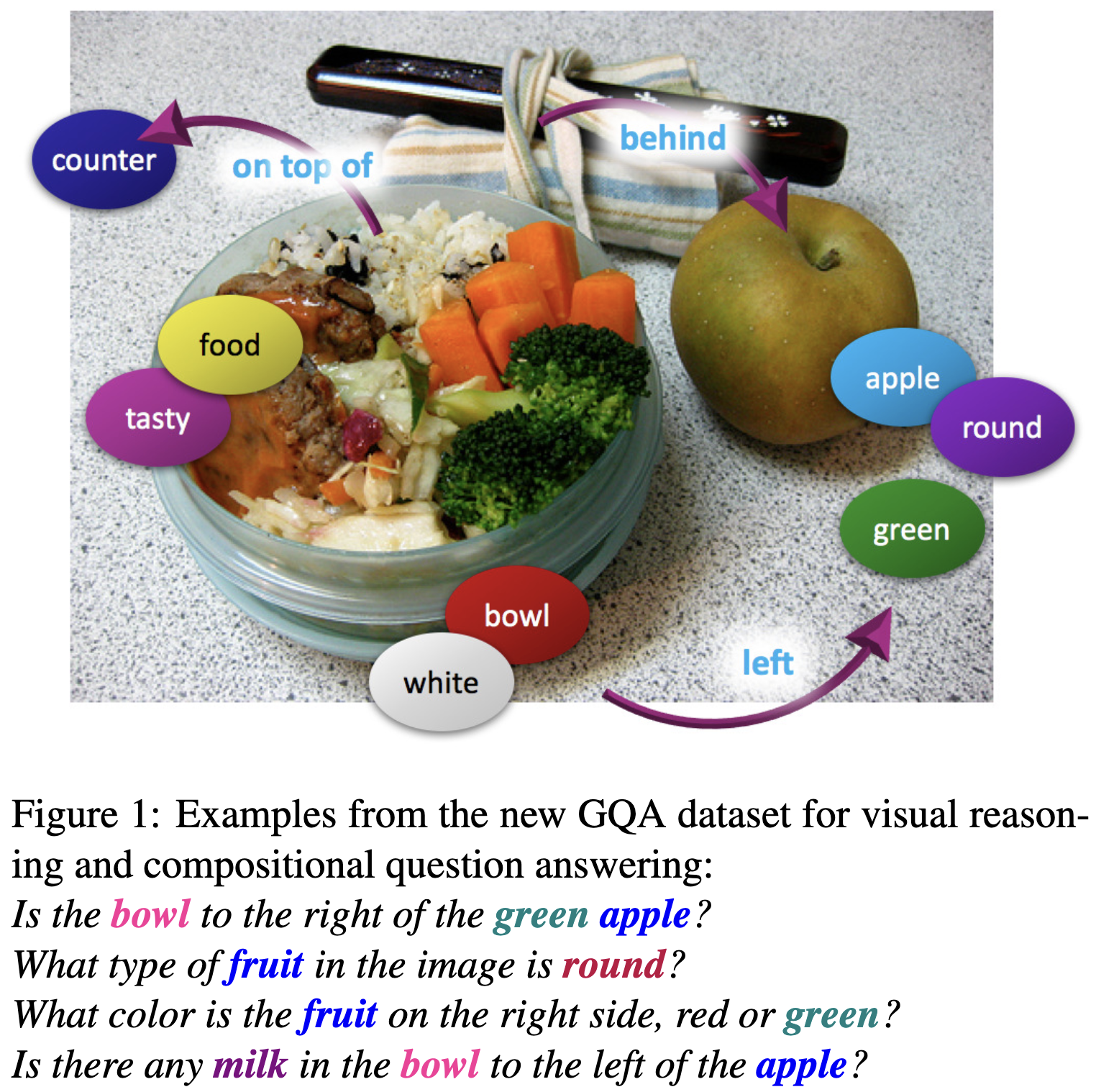}
    \caption{Example from the GQA dataset~\cite{hudson2019gqa}. Objects,
    attributes, and relations (e.g., \emph{bowl}, \emph{apple},
    \emph{green}, \emph{behind}, \emph{on top of}) are annotated in the
    image, and natural-language questions are derived from the
    underlying scene graph, such as ``Is the bowl to the right of the
    green apple?'' or ``What type of fruit in the image is round?''.
    This design explicitly links questions to structured semantics and
    supports detailed evaluation of compositional visual reasoning.}
    \label{fig:gqa-examples}
\end{figure}

\paragraph{GQA.}
GQA~\cite{hudson2019gqa} targets compositional visual reasoning and
fine-grained visual grounding. It derives around 22M questions from
structured scene graphs built on top of Visual Genome images, where
nodes represent objects and attributes and edges represent relations
(e.g., \textit{on top of}, \textit{behind}, \textit{left of}). Each
question is associated with a functional program that encodes the
underlying reasoning steps, enabling detailed analysis of model errors.
In Figure~\ref{fig:gqa-examples}, questions often require
multi-step relational reasoning such as querying attributes of objects
(e.g., the color of the fruit in the bowl) or comparing spatial
relations (e.g., whether one object is to the right of another). The
benchmark reports not only overall accuracy but also metrics for
consistency, grounding, and plausibility, making it a stringent testbed
for structured visual reasoning beyond the shorter, single-hop questions
typical of VQA v2.

\begin{figure}[ht]
    \centering
    \includegraphics[width=\linewidth]{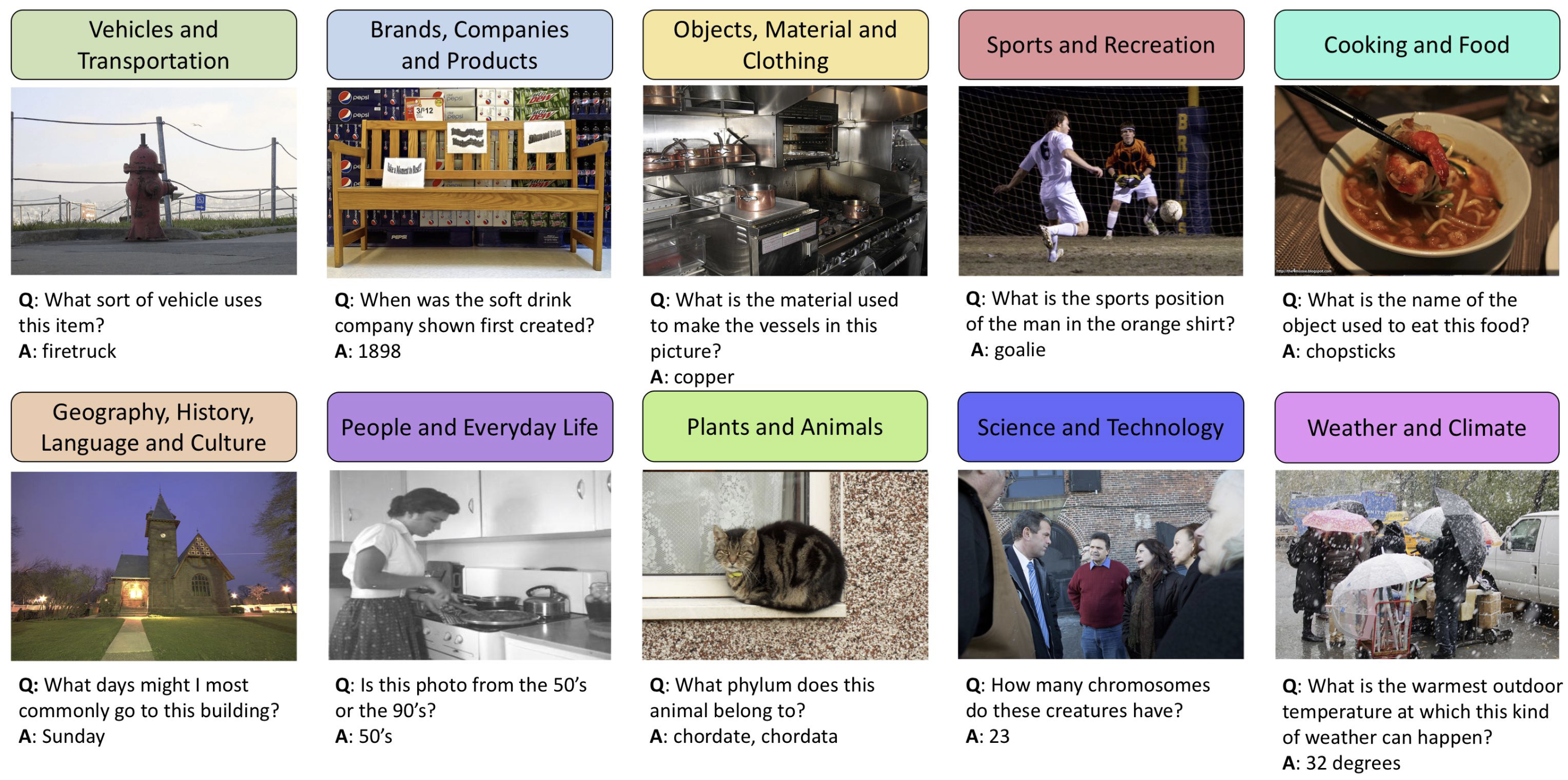}
    \caption{Example question-image-answer triplets from
    OK-VQA~\cite{marino2019okvqa}. Each panel corresponds to a semantic
    category (e.g., \emph{Vehicles and Transportation},
    \emph{Objects, Material and Clothing}, \emph{Science and
    Technology}) and shows a natural image with a question whose answer
    (A) depends on external world knowledge rather than direct visual
    recognition alone, such as identifying which vehicle uses a fire
    hydrant or the material of cookware.}
    \label{fig:okvqa-examples}
\end{figure}

\begin{figure}[ht]
    \centering
    \includegraphics[width=\linewidth]{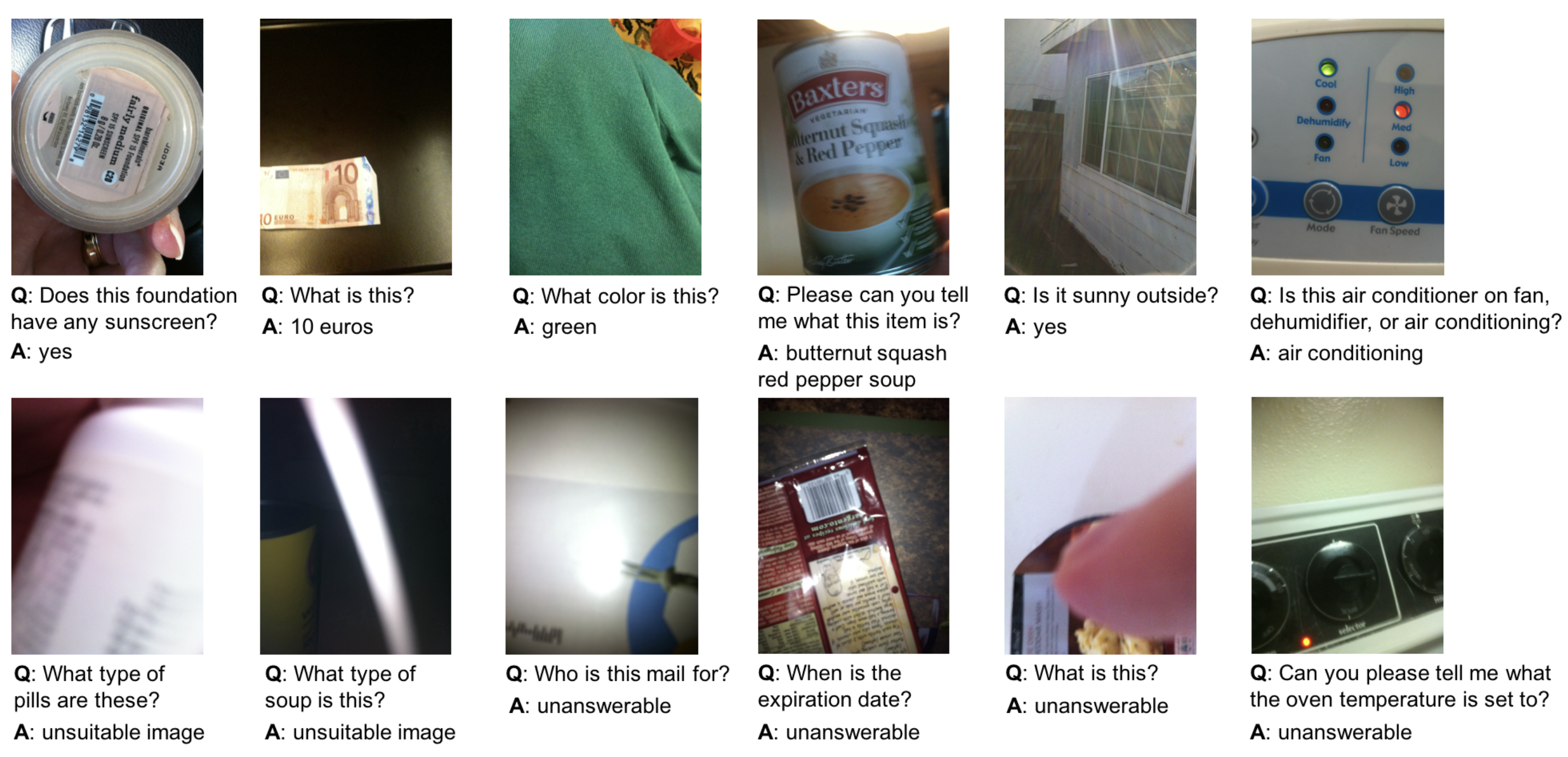}
    \caption{Example images and questions from the VizWiz
    dataset~\cite{gurari2018vizwiz}. Photographs are taken by blind or
    low-vision users in everyday settings, leading to off-center,
    blurry, or partially occluded views. Questions (Q) are spoken by
    users and later transcribed; many answers (A) are short phrases
    such as ``10 euros'' or ``air conditioning'', while others are
    labeled as \emph{unanswerable} when the visual evidence is
    insufficient. VizWiz thus provides a realistic benchmark for
    accessibility-focused VQA.}
    \label{fig:vizwiz-examples}
\end{figure}

\paragraph{Knowledge- and accessibility-oriented VQA.}
Several datasets emphasize capabilities that go beyond generic object
recognition. OK-VQA~\cite{marino2019okvqa} focuses on questions whose
answers require external world knowledge (e.g., tools, history, or
science) (Figure~\ref{fig:okvqa-examples}). Images are
grouped into semantic categories such as \emph{Vehicles and
Transportation}, \emph{Science and Technology}, or \emph{Cooking and
Food}, and questions explicitly probe background knowledge that is not
visually present (e.g., the phylum of an animal or the typical use of an
object). VizWiz~\cite{gurari2018vizwiz} instead targets accessibility
and assistive technology: images are captured by blind or low-vision
users with mobile phones, paired with spoken questions that are later
transcribed. As shown in Figure~\ref{fig:vizwiz-examples}, the photos
are often poorly framed, blurred, or over/under-exposed, and many
questions are inherently unanswerable, introducing realism and
ambiguity that are largely absent from curated lab datasets.

\subsection{Image-Text Retrieval}
\label{subsec:retrieval-datasets}

Image-text retrieval datasets evaluate a model’s ability to rank
matching images for a given caption (image retrieval) or matching
captions for an image (text retrieval). They are typically used with
embedding-based models, including CLIP-style dual encoders.

\paragraph{MS COCO and Flickr30k retrieval.}
MS COCO and Flickr30k provide standard retrieval splits, popularized
by Karpathy and Fei-Fei~\cite{karpathy2015deep}. For each test image,
the goal is to retrieve its ground-truth captions from a large pool,
and conversely to retrieve the correct image given a caption. Models
are evaluated with Recall@K (R@1, R@5, R@10) and median rank, measuring
how often the true match appears among the top-$K$ results.

\paragraph{Web-scale retrieval and domain-specific setups.}
On the web scale, CLIP-style pretraining on datasets such as
LAION-5B is often evaluated using held-out retrieval subsets or
downstream benchmarks where image and text come from different
domains (e.g., artworks, memes, or medical images). In specialized
domains, retrieval datasets may pair images with structured metadata
(e.g., product catalogs or document pages), and evaluation focuses on
cross-modal search quality under domain constraints.

\subsection{Referring Expressions and Grounding}
\label{subsec:grounding-datasets}

Grounding benchmarks require models to link language expressions to
specific regions, objects, or coordinates in an image, rather than
predicting only global labels.

\begin{figure}[ht]
    \centering
    \includegraphics[width=0.7\linewidth]{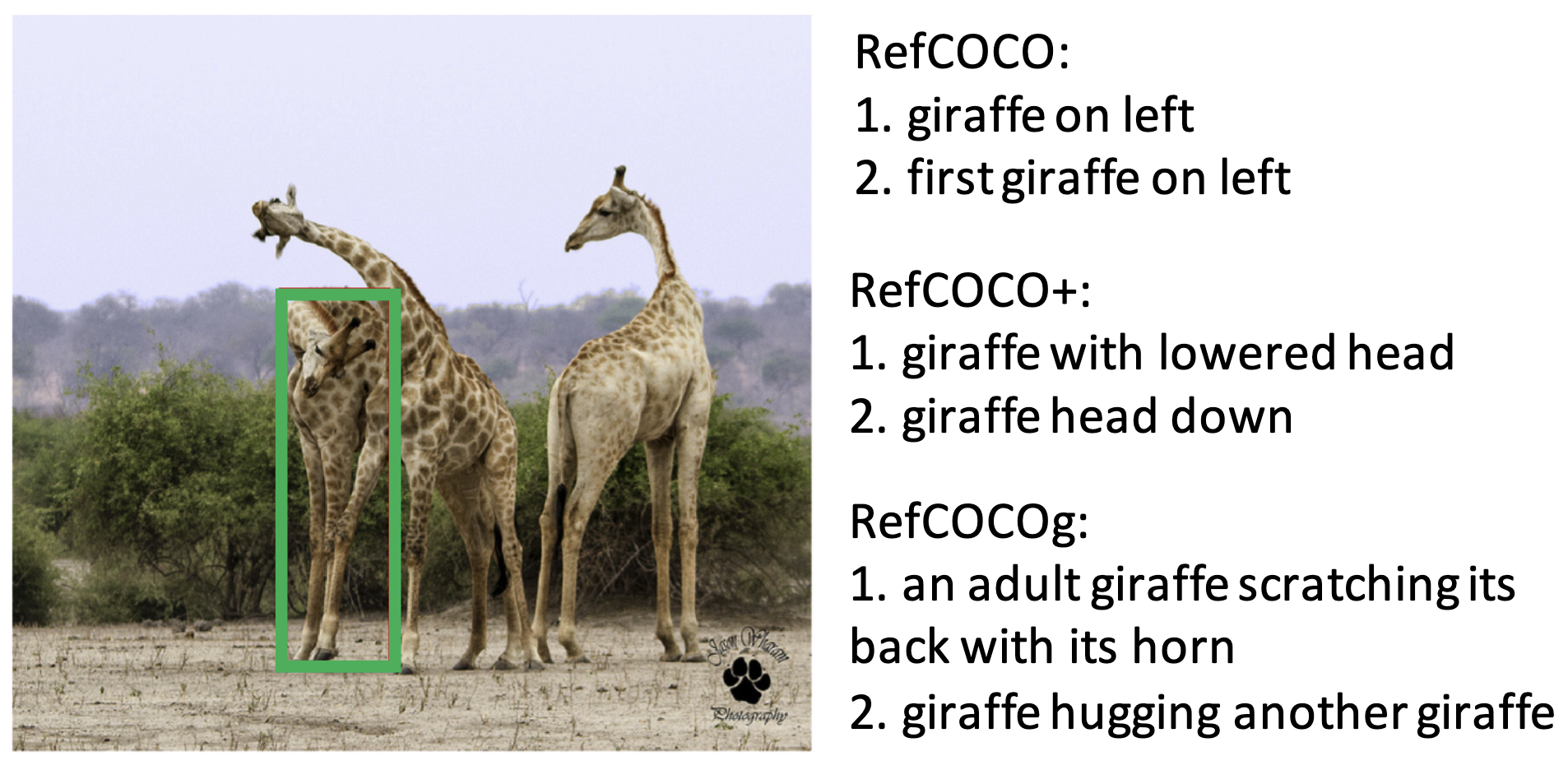}
    \caption{Example from the RefCOCO family of referring expression
    datasets~\cite{yu2016refcoco,kazemzadeh2014refexp}. The same target
    giraffe (highlighted in green) is described with different
    expressions across datasets: short, spatial descriptions in
    RefCOCO, appearance-based phrases in RefCOCO+, and longer, more
    descriptive sentences in RefCOCOg. These variations probe how well
    models can ground diverse linguistic referring forms in visual
    regions.}
    \label{fig:refcoco-ex}
\end{figure}

\paragraph{RefCOCO-style datasets.}
Referring expression datasets such as RefCOCO, RefCOCO+, and
RefCOCOg~\cite{yu2016refcoco,kazemzadeh2014refexp} are built on
MS~COCO images and provide natural-language descriptions of specific
objects. Each example consists of an image, a referring expression
(e.g., ``giraffe with lowered head'') and a ground-truth bounding box
for the target region. From Figure~\ref{fig:refcoco-ex},
RefCOCO focuses on short, spatially oriented phrases (e.g., ``giraffe
on left''), RefCOCO+ prohibits location words and thus emphasizes
appearance (e.g., ``giraffe with lowered head''), while RefCOCOg
contains longer, more descriptive expressions (e.g., ``an adult
giraffe scratching its back with its horn''). Evaluation typically uses
localization accuracy, measured by whether the predicted box overlaps
the ground truth with IoU above a threshold (commonly $0.5$).

\paragraph{Phrase grounding and region-level datasets.}
Flickr30k Entities~\cite{plummer2015flickrentities} annotates noun
phrases in Flickr30k captions with corresponding image regions, making
it possible to evaluate phrase-level localization. Other datasets
extend this idea to more complex structures such as scene graphs or
dense region descriptions. These resources are especially valuable for
training models that represent images as sets of semantically rich
regions and align them with textual spans.

\subsection{Document, OCR, and Chart Understanding}
\label{subsec:ocr-doc-datasets}

Document and chart understanding datasets focus on text-rich,
high-resolution images where layout and reading order are critical.

\begin{figure}[ht]
    \centering
    \includegraphics[width=0.6\linewidth]{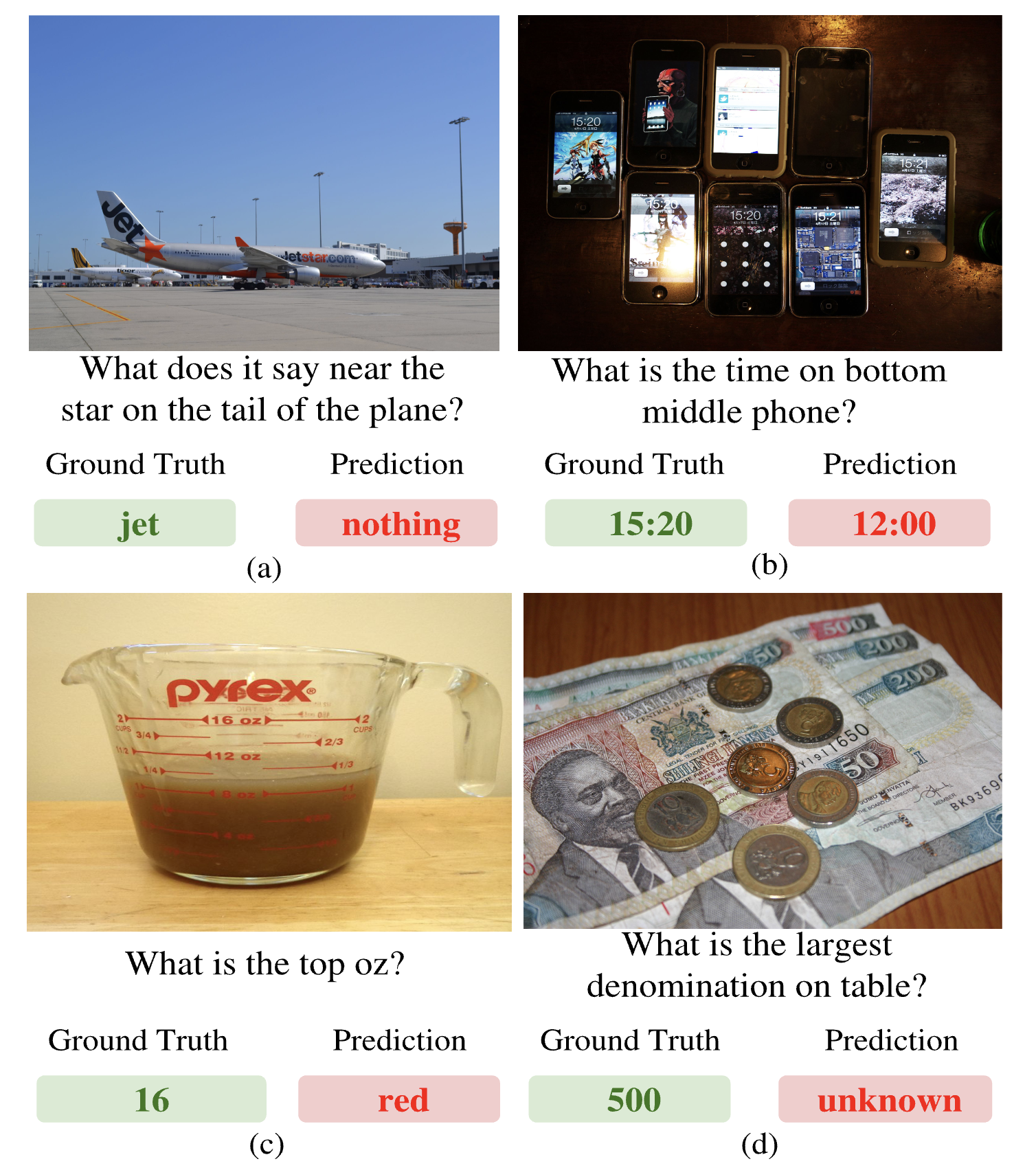}
    \caption{Representative examples from the TextVQA dataset
    \cite{singh2019textvqa}. Each image is paired with a question whose
    answer (shown in green) depends on reading scene text, such as
    airplane tail markings, digital clocks, measurement scales, or
    currency denominations. Baseline VQA models that lack strong OCR
    integration often predict incorrect answers (shown in red), 
    highlighting the challenge of text-centric visual reasoning.}
    \label{fig:textvqa-ex}
\end{figure}

\paragraph{Text-centric VQA.}
TextVQA~\cite{singh2019textvqa} focuses on images where answering the
question requires reading scene text (e.g., signage, product labels,
digital displays). As illustrated in Figure~\ref{fig:textvqa-ex}, many
questions explicitly ask about words or numbers present in the image,
and models frequently fail when they cannot correctly detect or
interpret the text. Answers are open-ended and often out-of-vocabulary
with respect to standard captioning corpora, forcing systems to combine
visual recognition, OCR, and language understanding in a single
pipeline. TextCaps~\cite{sidorov2020textcaps} provides captions for
similar text-rich images and is widely used to evaluate text-aware
captioning models, as discussed earlier.

\begin{figure}[ht]
    \centering
    \includegraphics[width=0.65\linewidth]{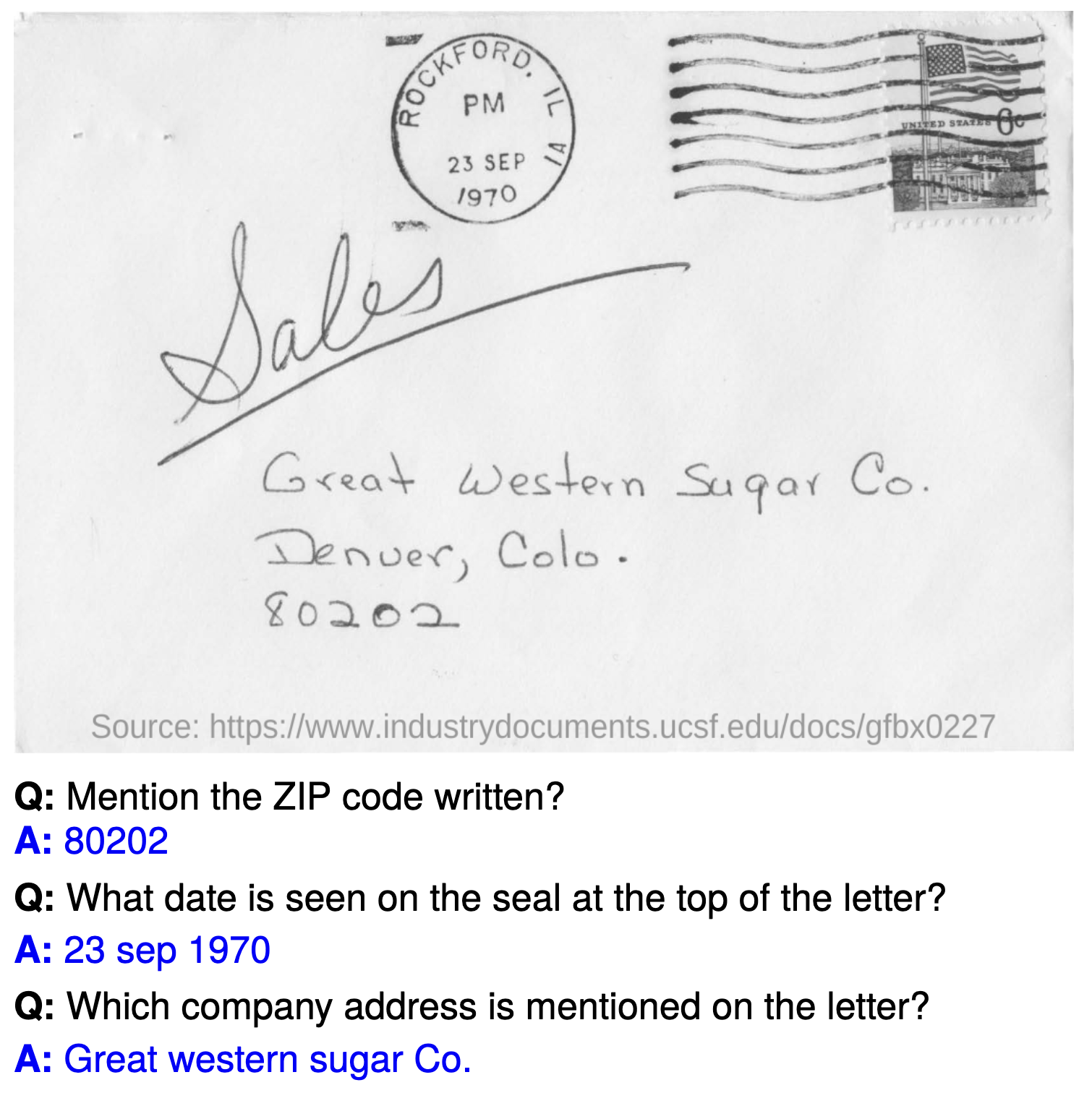}
    \caption{Example from DocVQA~\cite{mathew2021docvqa}. The model must
    answer multiple questions about a scanned envelope, such as the ZIP
    code, the postmark date, and the company name. Successful solutions
    require accurate OCR, spatial layout understanding, and reasoning
    over several textual elements on the page.}
    \label{fig:docvqa-ex}
\end{figure}

\paragraph{Document understanding.}
DocVQA~\cite{mathew2021docvqa} defines multiple tasks over scanned
documents, including text-based question answering and document
structure understanding. Figure~\ref{fig:docvqa-ex} shows a typical
example: questions require reading several fields on the page (e.g.,
dates, ZIP codes, company names) and reasoning over them at the
document level. RVL-CDIP~\cite{harley2015rvlcdip} and
PubLayNet~\cite{zhong2019publaynet} are large-scale document
classification and layout datasets that are frequently used to pretrain
visual backbones for document VLMs before fine-tuning on DocVQA-style QA
tasks.

\begin{figure}[ht]
    \centering
    \includegraphics[width=0.75\linewidth]{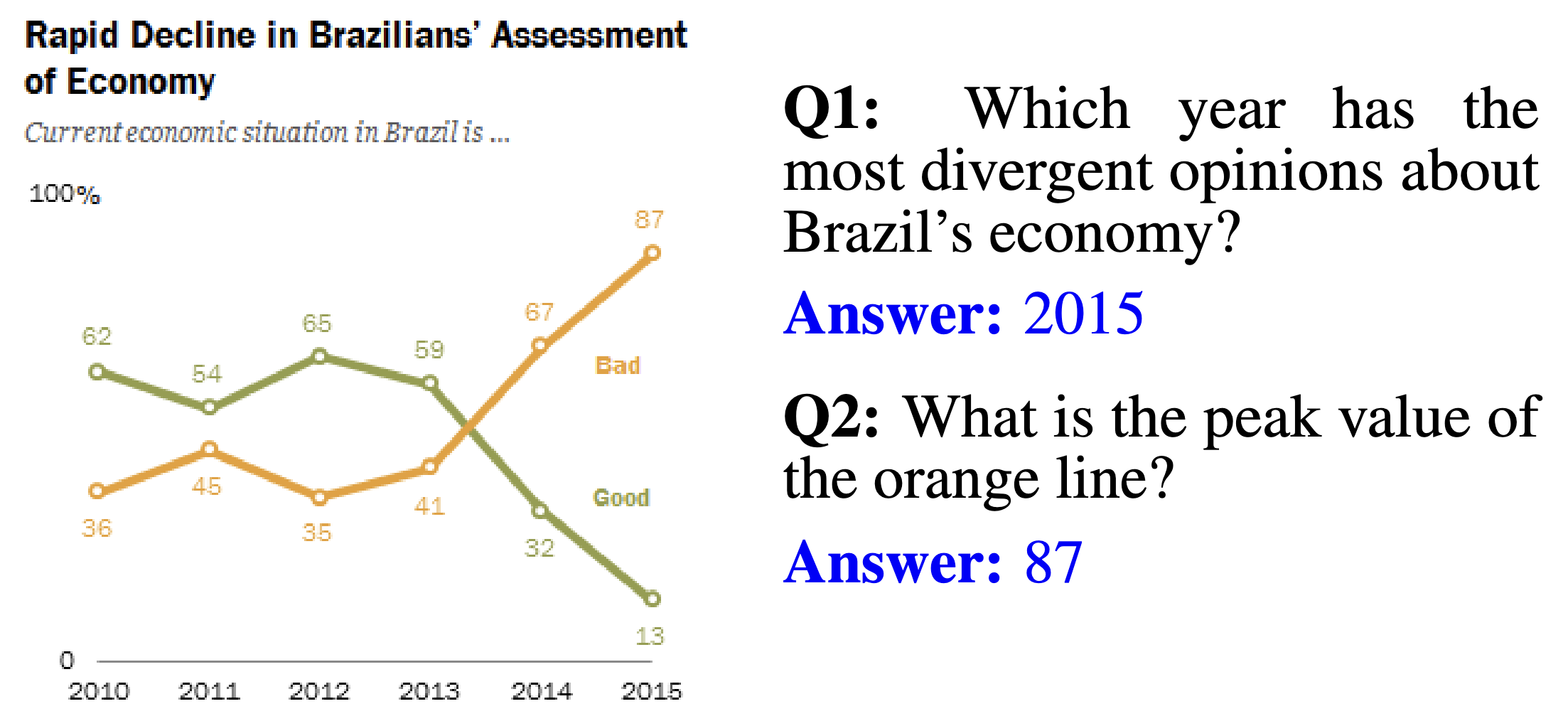}
    \caption{Example from ChartQA~\cite{masry2022chartqa}. The model is
    asked questions about a line chart summarizing survey responses,
    such as \emph{“Which year has the most divergent opinions about
    Brazil’s economy?”} and \emph{“What is the peak value of the orange
    line?”}. Solving such problems requires accurate reading of plotted
    values, comparison between series, and reasoning over trends rather
    than just recognizing objects.}
    \label{fig:chartqa-example}
\end{figure}

\paragraph{Charts and infographics.}
ChartQA~\cite{masry2022chartqa} pairs synthetic and real-world charts
with natural language questions, requiring models to parse axes,
legends, and visual encodings in order to compute answers. An example
item is given in Figure~\ref{fig:chartqa-example}, where the model must
identify the year with the largest gap between two curves and read off a
numerical value from the orange series. InfographicsVQA~\cite{mathew2022infographicsvqa}
focuses on complex infographic posters combining text, icons, and
illustrations; questions often require multi-hop reasoning across
multiple textual and visual elements. 

\subsection{Multimodal Instruction-Tuning Corpora}
\label{subsec:instr-tuning-data}

As VLMs evolve into general-purpose assistants, \emph{instruction
tuning} on multimodal dialogues has become a standard step. Rather
than training solely on captions or short QA pairs, models are exposed
to conversations in which users issue natural-language instructions
about images or documents.

\paragraph{LLaVA-style instruction data.}
LLaVA~\cite{liu2023visual} constructs large collections of
image-instruction-response triples, many of which are generated by a
powerful teacher model (e.g., GPT-4) conditioned on web images. These
prompts cover tasks such as detailed description, reasoning, stepwise
instruction, and open-ended conversation. Instruction tuning on this
data enables relatively small open-source models to behave like
helpful multimodal chatbots.

\paragraph{Instruction-tuned VLM families.}
InstructBLIP~\cite{dai2023instructblip} extends BLIP-2 with a
multi-stage instruction-tuning pipeline that mixes captioning, VQA,
and conversational data. Qwen-VL and Qwen2.5-VL follow a similar
pattern, combining image-text corpora, OCR tasks, chart and document
QA, and synthetic multimodal instructions to support a wide range of
use cases. Many recent systems also incorporate chain-of-thought
rationales, tool calls (e.g., web search or code execution), and
structured output formats into their instruction data, blurring the
boundary between supervised datasets and user-facing interaction logs.

Instruction-tuning corpora thus serve as a bridge between traditional
benchmarks and real-world usage: they align models with human
preferences, normalize input and output formats across tasks, and
provide the conversational scaffolding needed for VLMs to function as
interactive assistants rather than single-task predictors.

\begin{didyouknow}
Multimodal instruction data is often more about \emph{shaping behaviour}
than improving raw perception. Many of the images in LLaVA-style
datasets are visually simple; what really changes the model is the
distribution of prompts and responses: how to hedge when uncertain, how
to explain reasoning in steps, how to say ``I can’t see that clearly'', or
how to format JSON for downstream tools. In practice, a modest amount of
well-designed instruction data can dramatically change how ``assistant-
like'' a VLM feels, even if its underlying visual encoder stays the same.
\end{didyouknow}

\section{Evaluation Benchmarks and Metrics}
\label{sec:eval-benchmarks}

Evaluating VLMs requires benchmarks and metrics that
reflect both linguistic quality and visual faithfulness.

\subsection{Captioning Benchmarks and Metrics}
\label{subsec:caption-metrics}

The dominant captioning benchmarks are MS~COCO Captions and related
datasets such as nocaps and TextCaps
(Section~\ref{subsec:caption-datasets}). Models are typically evaluated
on fixed test splits with automatic text-overlap metrics computed
between system outputs and multiple human reference captions.

\paragraph{n-gram overlap metrics.}
BLEU~\cite{papineni2002bleu} computes a brevity-penalized geometric mean
of modified $n$-gram precisions:
\begin{equation}
  \mathrm{BLEU}
  = \mathrm{BP} \cdot
    \exp\!\Bigg(
      \frac{1}{N} \sum_{n=1}^{N} \log p_n
    \Bigg),
\end{equation}
where $p_n$ is the clipped precision for $n$-grams of length $n$, and
$\mathrm{BP}$ is a brevity penalty to discourage overly short captions.
METEOR~\cite{denkowski2014meteor} instead aligns hypothesis and
reference tokens using exact, stem, and synonym matches, and reports a
recall-oriented F-score with fragmentation penalties for broken chunks.
ROUGE-L~\cite{lin2004rouge} measures the length of the longest common
subsequence (LCS) between hypothesis and references, capturing sentence-
level fluency.

\paragraph{Consensus-based metrics.}
CIDEr~\cite{vedantam2015cider} was designed specifically for image
captioning. Each caption is represented as a TF–IDF weighted
$n$-gram vector $g_n(\cdot)$; the similarity between a candidate $c$ and
a reference set $\{r_m\}$ is
\begin{equation}
  \mathrm{CIDEr}(c, \{r_m\})
  = \frac{1}{M} \sum_{m=1}^{M}
    \sum_{n=1}^{4} w_n
    \frac{ g_n(c)^\top g_n(r_m) }
         { \|g_n(c)\| \, \|g_n(r_m)\| },
\end{equation}
averaged over $n$-gram orders with weights $w_n$. SPICE~\cite{anderson2016spice}
parses captions into scene-graph tuples (objects, attributes,
relations) and computes an F-score over semantic tuples, better
capturing high-level content than raw word overlap.

\paragraph{Learned semantic metrics.}
More recent work employs learned similarity functions such as
BERTScore~\cite{zhang2020bertscore}, which aligns contextual embeddings
from BERT-like encoders, and CLIPScore~\cite{hessel2021clipscore},
which measures cosine similarity between the generated caption and the
image in a CLIP embedding space. These metrics correlate better with
human judgments on some benchmarks, but can inherit biases from their
underlying models (e.g., rewarding captions that mention visually
plausible but absent objects).

In practice, captioning papers report a suite of metrics (BLEU-1/4,
METEOR, ROUGE-L, CIDEr, SPICE, sometimes CLIPScore) to provide a
balanced view of $n$-gram fidelity, semantic adequacy, and visual
relevance.

\subsection{VQA Benchmarks and Metrics}
\label{subsec:vqa-metrics}

Visual Question Answering benchmarks evaluate a model’s ability to
produce short answers conditioned on an image and a natural-language
question. The standard metric is accuracy, but details of the scoring
function matter.

\paragraph{Consensus-based accuracy.}
VQA v2~\cite{goyal2017vqa2} collects 10 human answers per question and
defines a soft accuracy that rewards agreement with the annotator
majority:
\begin{equation}
  \mathrm{Acc}(a)
  = \min\!\Bigg(
      \frac{1}{3}
      \sum_{k=1}^{10} \mathbb{I}[a = a_k],
      1
    \Bigg),
\end{equation}
where $a$ is the model prediction and $a_k$ are human answers. This
mitigates annotation noise and synonymy effects. OK-VQA~\cite{marino2019okvqa}
and VizWiz~\cite{gurari2018vizwiz} adopt similar consensus scoring.

GQA~\cite{hudson2019gqa} uses exact-match accuracy but augments it with
metrics for consistency (agreement across logically related questions),
validity (well-formed answers), and plausibility (answers that are
reasonable given the world), informed by underlying functional programs
derived from scene graphs.

\paragraph{Answer spaces and robustness.}
Benchmarks differ in whether answers are drawn from a closed vocabulary
(VQA v2 multiple-choice settings) or are free-form (TextVQA, OK-VQA).
Free-form evaluation often maps predictions to canonical forms via
lowercasing, number normalization, and simple string rules. Recent
studies also explore robustness metrics, such as performance under
adversarial rephrasings or input perturbations, but these are not yet
standardized.

\begin{didyouknow}
Because most VQA benchmarks report a single accuracy number, it is easy
to overlook how fragile that score can be. Many models show large drops
when questions are slightly rephrased, when answer distributions are
balanced, or when images are perturbed with small crops or blur. As a
result, two systems with nearly identical headline accuracy on VQA v2
can behave very differently in realistic settings, especially on
edge-case questions where annotators themselves disagree.
\end{didyouknow}

\subsection{Retrieval Benchmarks and Metrics}
\label{subsec:retrieval-metrics}

Image-text retrieval benchmarks (e.g., MS~COCO and Flickr30k retrieval
splits) evaluate whether a model can rank the correct caption for a
given image (image-to-text) or the correct image for a caption
(text-to-image).

\paragraph{Recall-based metrics.}
The primary metrics are Recall@K:
\begin{equation}
  \mathrm{R@K}
  = \frac{1}{N}
    \sum_{i=1}^{N}
    \mathbb{I}\big[\mathrm{rank}(y_i \mid x_i) \le K\big],
\end{equation}
where $\mathrm{rank}(y_i \mid x_i)$ is the rank of the true target
$y_i$ for query $x_i$ in the retrieved list. R@1, R@5, and R@10 are
routinely reported, sometimes accompanied by median rank. When multiple
ground-truth captions or images exist per query, the best-scoring
target is used.

\paragraph{Ranking quality.}
For large-scale retrieval (e.g., web-scale search), mean Average
Precision (mAP) and normalized discounted cumulative gain (nDCG) are
used to capture ranking quality across many relevant items. However,
these metrics are less common in image-caption retrieval due to the
small number of labeled matches per query.

\subsection{Grounding and Detection Benchmarks}
\label{subsec:grounding-metrics}

Grounding benchmarks measure the ability of a model to localize visual
regions corresponding to phrases or referring expressions.

\paragraph{IoU-based localization metrics.}
The standard criterion is Intersection over Union (IoU) between a
predicted bounding box $B_p$ and a ground-truth box $B_{\text{gt}}$:
\begin{equation}
  \mathrm{IoU}(B_p, B_{\text{gt}})
  = \frac{|B_p \cap B_{\text{gt}}|}{|B_p \cup B_{\text{gt}}|}.
\end{equation}
A prediction is counted as correct if $\mathrm{IoU} \ge \tau$ (commonly
$\tau = 0.5$). Referring expression benchmarks such as RefCOCO-style
datasets report \emph{localization accuracy}, the fraction of examples
for which at least one predicted box meets the IoU threshold.

\paragraph{Average precision and phrase grounding.}
For phrase grounding with multiple phrases per image, evaluation often
resembles object detection: models output scored boxes for each phrase,
and Average Precision (AP) is computed by integrating precision-recall
curves over different score thresholds and IoU cutoffs. Some works also
use the ``pointing game,'' which checks whether the maximum-response
pixel (e.g., from an attention map) falls inside the ground-truth box.

\subsection{Hallucination and Faithfulness Evaluation}
\label{subsec:hallucination-benchmarks}

Standard captioning and VQA metrics do not directly penalize
\emph{hallucinations}-statements that are fluent and plausible but not
supported by the image. Specialized benchmarks and metrics have been
proposed to address this gap.

\paragraph{CHAIR and caption hallucination.}
\citet{rohrbach2018object} introduce CHAIR
(\emph{Caption Hallucination Assessment with Image Relevance}) to
quantify object hallucination in image captioning. Let $O_{\text{img}}$
denote the set of objects present in the image (from annotations) and
$O_{\text{cap}}$ the set of object nouns mentioned in a caption
(predicted or reference). Hallucinated objects are
$O_{\text{hall}} = O_{\text{cap}} \setminus O_{\text{img}}$.
Two metrics are reported:
\begin{align}
  \mathrm{CHAIR}_i
  &= \frac{|O_{\text{hall}}|}{|O_{\text{cap}}|} \\
  \mathrm{CHAIR}_s
  &= \frac{\#\text{captions with } O_{\text{hall}} \neq \varnothing}
         {\#\text{captions}}
\end{align}
Lower values indicate fewer hallucinated objects. CHAIR is often
computed for both model-generated captions and human references to
control for annotation noise.

\begin{figure}[t]
    \centering
    \includegraphics[width=\linewidth]{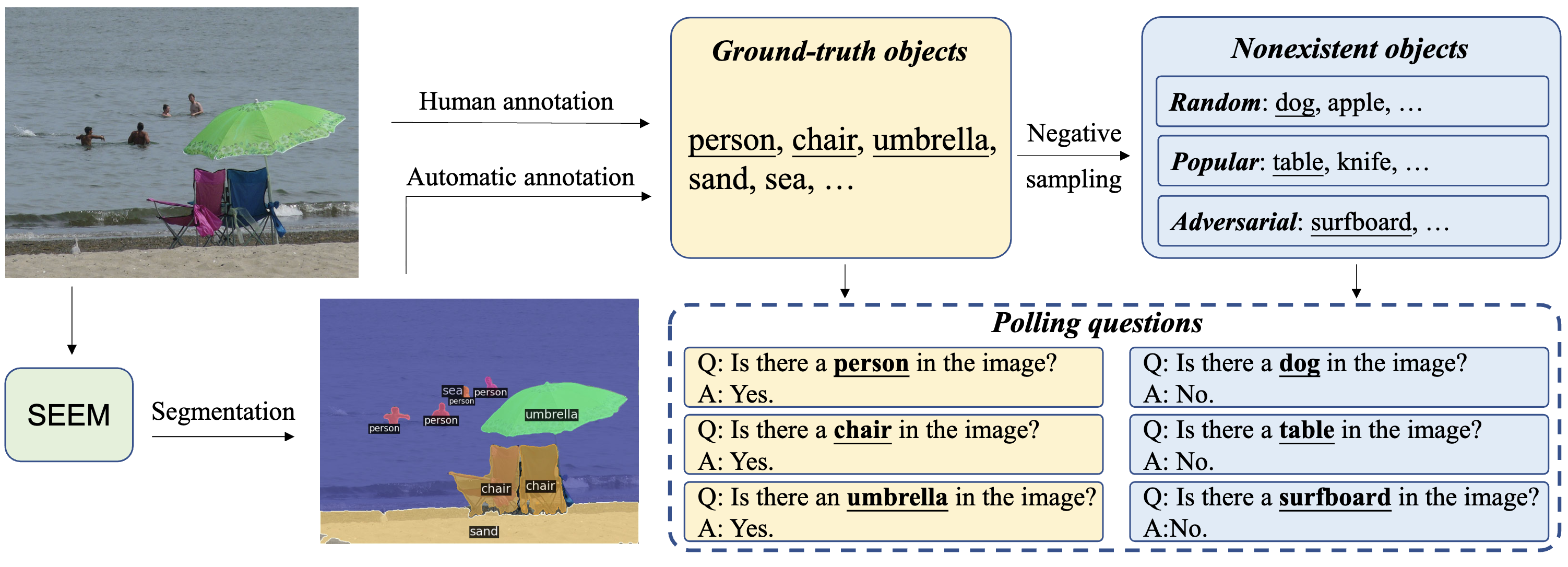}
    \caption{Overview of the POPE (\emph{Polling-based Object Probing Evaluation}) pipeline for measuring object hallucination in LVLMs~\cite{li2023evaluating}. Given an input image, ground-truth object categories (e.g., \textit{person}, \textit{chair}, \textit{umbrella}) are obtained from human or automatic annotations, while nonexistent objects (e.g., \textit{dog}, \textit{table}, \textit{surfboard}) are sampled from random, frequent, or adversarial distributions. For each object, the model answers yes/no polling questions of the form ``Is there a \textit{X} in the image?'', and its responses are summarized using accuracy, precision, recall, F1, and the overall rate of ``yes'' predictions.}
    \label{fig:pope-pipeline}
\end{figure}

\paragraph{POPE and object probing.}
The POPE benchmark (\emph{Polling-based Object Probing Evaluation})~\cite{li2023evaluating} evaluates object hallucination in LVLMs using binary presence queries of the form ``Is there a \textit{X} in the image?''. For each image, a set of ground-truth objects is obtained from human or automatic annotations (e.g., via a segmentation model such as SEEM), while additional nonexistent objects are sampled from random, frequent (``popular''), or adversarial categories. The model is then queried with a sequence of yes/no questions over both present and absent objects (see Figure~\ref{fig:pope-pipeline}). Evaluation metrics include accuracy, precision, recall, F1, and the overall proportion of ``yes'' responses. High precision and balanced ``yes'' rates indicate that the model is neither over-confidently hallucinating objects nor trivially answering “no’’ to avoid mistakes.

\paragraph{Faithfulness in VQA and dialogue.}
Other works probe faithfulness by checking whether model rationales or
attention maps align with annotated evidence, or by constructing counterfactual
images/questions where a faithful model should change its answer.
However, no single faithfulness metric has yet become standard; current
practice is to report CHAIR/CHAIRs, POPE scores, and task-specific
probes alongside traditional accuracy or CIDEr.

\begin{didyouknow}
Among the many evaluation topics in this chapter, hallucination is
arguably one of the most practically important. In real applications,
users rarely notice a small drop in BLEU or VQA accuracy - but they do
notice when a model confidently describes objects that are not in the
image, or invents text on a receipt that does not exist. Metrics like
CHAIR and POPE are therefore not just academic add-ons: they directly
measure whether we can trust a VLM's descriptions enough to use them in
assistive tools, document workflows, or decision-support systems.
\end{didyouknow}

\subsection{Holistic and Multi-Task Benchmarks}
\label{subsec:holistic-benchmarks}

As LVLMs evolve into general-purpose assistants, evaluation has moved
beyond single-task datasets toward broad suites that probe diverse
perceptual and reasoning skills.

\begin{figure}[t]
    \centering
    \includegraphics[width=0.7\linewidth]{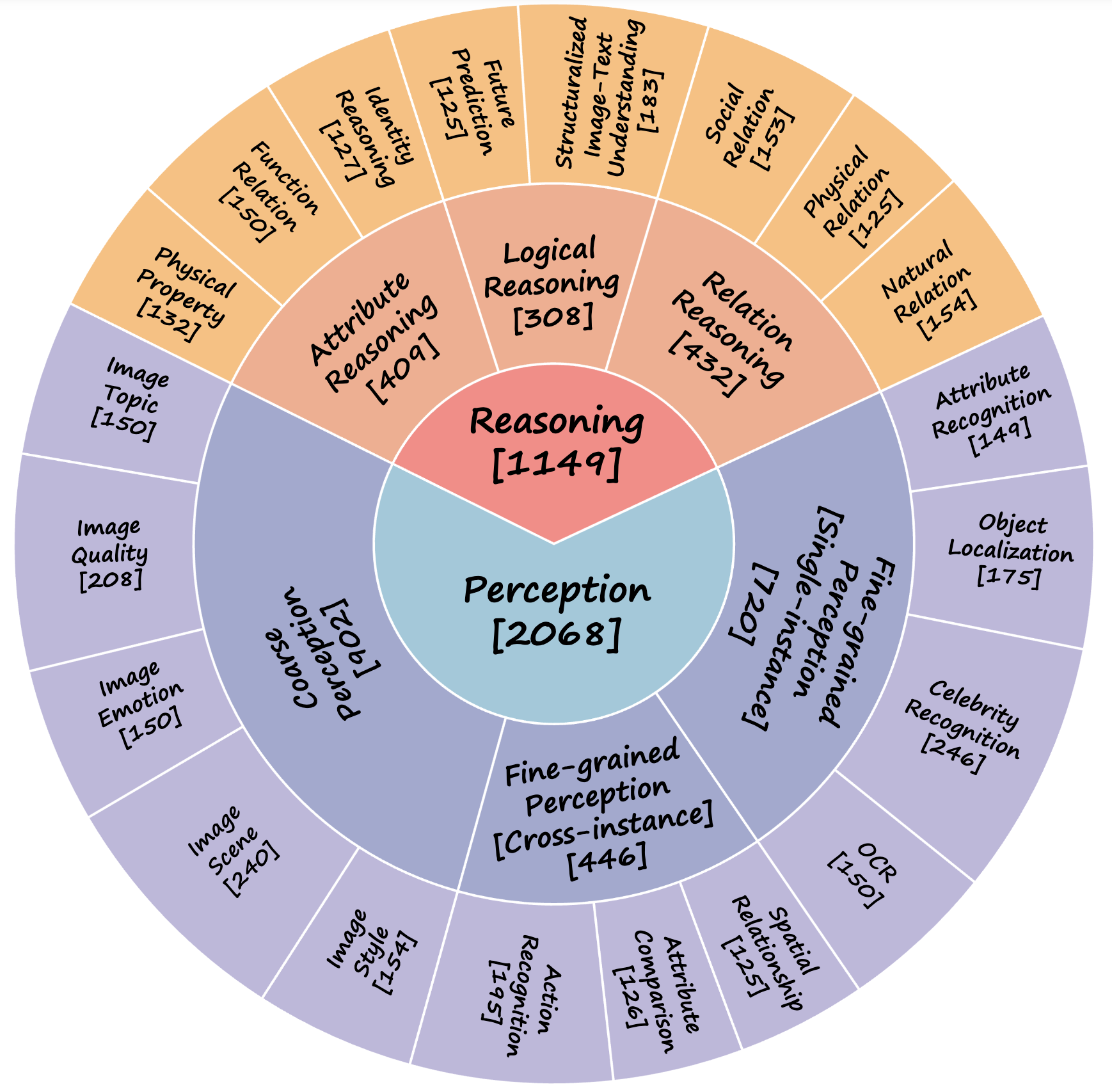}
    \caption{Ability taxonomy in MMBench~\cite{liu2023mmbench}. The
    benchmark organizes visual questions into hierarchical ability
    dimensions, separating broad \emph{Perception} and \emph{Reasoning}
    categories and further decomposing them into fine-grained skills
    (outer ring), such as OCR, spatial relationships, logical reasoning,
    and identity or attribute reasoning.}
    \label{fig:mmbench-abilities}
\end{figure}

\paragraph{MMBench and SEED-Bench.}
MMBench~\cite{liu2023mmbench} and SEED-Bench~\cite{li2023seedbench}
consist primarily of multiple-choice questions spanning ability
dimensions such as coarse and fine-grained perception, OCR, world
knowledge, commonsense and logical reasoning, as well as safety and
instruction-following. In MMBench, for example, questions are organized
into a three-level hierarchy of capabilities covering more than twenty
leaf skills, ranging from object localization and attribute recognition
to structured reasoning and visual text understanding
(see Figure~\ref{fig:mmbench-abilities}). Images and questions are
carefully curated to reduce annotation artifacts and language priors,
and human-verified answer keys enable automatic evaluation via accuracy.
Public leaderboards report model performance across all abilities,
providing a standardized snapshot of LVLM progress.

\paragraph{LVLM-eHub and related suites.}
LVLM-eHub~\cite{xu2024lvlmehub} aggregates a large collection of existing
benchmarks (captioning, VQA, grounding, chart understanding, etc.) and
provides standardized evaluation scripts and prompts. Models are scored
on each sub-task using the appropriate metrics (e.g., CIDEr for COCO,
VQA accuracy, IoU/AP for grounding), and an overall score is computed as
a weighted or unweighted average, giving a more holistic picture of
capabilities.

\begin{figure}[t]
    \centering
    \includegraphics[width=\linewidth]{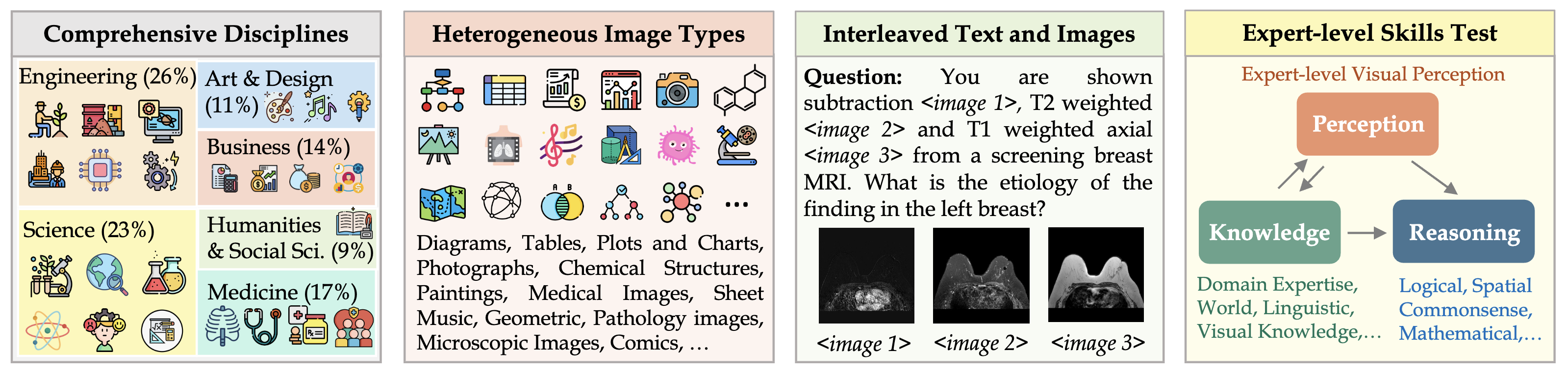}
    \caption{Overview of the MMMU benchmark~\cite{yue2024mmmu}. The
    dataset comprises $\sim$11.5K college-level, multiple-choice
    problems drawn from six broad disciplines and 30 subjects,
    featuring heterogeneous image types (e.g., diagrams, charts,
    photographs), interleaved text and images, and questions designed
    to probe expert-level perception, knowledge, and reasoning.}
    \label{fig:mmmu-overview}
\end{figure}

\paragraph{MMMU and exam-style evaluations.}
MMMU~\cite{yue2024mmmu} and related benchmarks cast evaluation as
multimodal standardized exams spanning many academic disciplines.
Each problem may include diagrams, tables, or photographs, and is
typically posed as a multiple-choice question. Tasks cover six broad
discipline areas (e.g., engineering, science, medicine, humanities)
and dozens of college-level subjects, with questions that require both
expert-level visual perception and domain knowledge. An overview of
MMMU’s coverage, heterogeneous image types, and interleaved
image-text format is provided in Figure~\ref{fig:mmmu-overview}.
Models are scored by answer accuracy, but success on MMMU is widely
interpreted as evidence of stronger general reasoning and subject-matter
understanding, rather than narrow pattern matching.

\begin{didyouknow}
Holistic benchmarks like MMBench and MMMU sometimes reveal ``hidden''
strengths and weaknesses that single-task scores never show. A model can
look strong on classic captioning and VQA, yet collapse on simple chart
questions, textbook diagrams, or exam-style word problems. Conversely,
some LVLMs with only mid-range COCO scores turn out to be remarkably
good at OCR-heavy or math-heavy items. This is one reason serious
evaluations increasingly rely on \emph{suites} of benchmarks: no single
score can capture what it actually feels like to use a model across
real-world multimodal tasks.
\end{didyouknow}

\section{Challenges in Dataset and Benchmark Design}
\label{sec:dataset-challenges}

Despite impressive progress, current datasets and benchmarks only
partially capture the capabilities and failure modes of modern VLMs.
Designing reliable evaluations is itself a challenging research problem.

\subsection{Biases, Shortcuts, and Data Contamination}
\label{subsec:bias-shortcuts}

Vision-language datasets inherit systematic biases from both the visual
sources (e.g., web photographs, stock images) and the annotation
process. In VQA, for example, strong \emph{language priors} allow models
to answer many questions without looking at the image at all: questions
beginning with ``Do you see a \dots'' may be answered ``yes'' a large
fraction of the time, and color questions have heavily skewed answer
distributions. Balancing strategies such as those used in
VQA v2 reduce-but do not eliminate-these effects by collecting
complementary examples where the same question has different answers.

A related issue is \emph{annotation artifacts}. Human annotators often
reuse stock phrases, exhibit consistent stylistic quirks, or respond
differently under time pressure. Models can latch onto these superficial
signals as ``shortcuts'' rather than learning genuine visual grounding
or reasoning. Shortcut learning has been documented extensively in
image recognition and NLP, and similar phenomena have been observed in
multimodal settings: for instance, models may key on the presence of
certain words in captions (``playing'', ``riding'') to infer activity
labels, even when the visual content is ambiguous.

\emph{Data contamination} further complicates evaluation. Large-scale
pretraining on web corpora makes it increasingly likely that test
images or captions (or near-duplicates) appear in the training data. If
a benchmark overlaps heavily with the pretraining corpus, reported
performance may partly reflect memorization rather than generalization.
This is particularly acute for popular datasets such as COCO, Visual
Genome, and MS-COCO-based VQA benchmarks, which have been recycled as
both supervised training data and evaluation targets. Careful deduping,
release of training URL lists, and \emph{de novo} benchmark construction
are active areas of work but are not yet standard practice.

Finally, subtle forms of leakage arise from \emph{near-duplicate} or
template-based test items: if many evaluation questions share almost
identical wording with training examples, models can generalize by
pattern matching over text alone. For LVLMs, disentangling genuine
multimodal competence from such shortcuts remains an open methodological
challenge.

\subsection{Robustness and Out-of-Distribution Evaluation}
\label{subsec:robustness-ood}

Most benchmarks evaluate models on data drawn from the same distribution
as their supervised training sets: everyday photographs, short English
captions, and generic questions. However, real-world deployment often
involves substantial \emph{domain shift}: medical images, charts,
documents, UI screenshots, or images taken under unusual lighting and
viewpoints. Models that perform strongly on in-distribution tests may
degrade sharply when exposed to these settings.

Robustness-oriented evaluations attempt to probe this gap. Some
benchmarks construct explicit \emph{distribution shifts} by corrupting
images (e.g., noise, blur, weather artifacts), altering object textures,
or changing rendering styles. Others focus on \emph{adversarial or
counterfactual} examples in which small changes to the image or
question should flip the correct answer but often do not-for instance,
swapping object colors, moving an object to the opposite side of the
scene, or editing numerical quantities in charts. Compositionality
tests, such as those derived from synthetic scenes or scene graphs,
stress whether a model can correctly answer questions about novel
combinations of familiar attributes and relations.

For LVLMs, a further dimension is \emph{prompt robustness}. Slightly
rephrasing a question, adding irrelevant context, or interleaving
multiple images can lead to qualitatively different behaviors. Current
benchmarks rarely vary prompts systematically, making it difficult to
tell whether models are robust to natural paraphrases or conversational
noise. Developing evaluation suites that combine visual OOD stress
tests with linguistic variation is an important direction for future
work.

\subsection{Human Evaluation and Safety Considerations}
\label{subsec:human-eval-safety}

Automatic metrics are indispensable for large-scale experimentation, but
they only imperfectly capture the qualities that matter in practice. For
open-ended generation (e.g., detailed image descriptions, explanations,
or multi-turn dialogues), metrics such as BLEU or CIDEr correlate only
moderately with human judgment and are insensitive to factual accuracy,
politeness, or helpfulness. As LVLMs are increasingly deployed as
interactive assistants, \emph{human evaluation} becomes essential:
expert raters or crowd workers are asked to compare model outputs, rate
helpfulness and correctness, or judge whether a response is appropriate
for a given user query and image.

Human-in-the-loop evaluation is also central to \emph{safety} and
\emph{fairness}. Images may contain sensitive content, including faces,
license plates, medical scans, or scenes of violence. Benchmarks must
therefore address privacy (e.g., by blurring identifiers or using
synthetic faces), and evaluation protocols should check whether models
respect safety guidelines when asked to describe or manipulate such
content. In addition, dataset composition can introduce demographic and
cultural biases: if people from certain groups are systematically
underrepresented or depicted in stereotypical roles, LVLMs may learn and
amplify those stereotypes. Fairness-aware benchmarking requires both
auditing existing datasets for such patterns and designing targeted
stress tests that probe for biased behavior.

Finally, as models are optimized to perform well on existing automatic
benchmarks, there is a risk of \emph{overfitting to the metric} rather
than to human preferences. For example, captioning systems can inflate
n-gram overlap scores by producing safe, generic sentences that mention
many common objects but fail to capture what is distinctive in an image.
Incorporating human preference data into training and evaluation, and
reporting a mix of automatic and human-centered metrics, are key steps
toward more trustworthy assessment of multimodal systems.
\chapter{Applications of Vision-Language Models}
\label{chap:applications-future}

\section{Overview}

Vision-Language Models (VLMs) have moved from research prototypes to
tools that people interact with every day. They power assistants that
can describe photos, read documents, help debug visual user interfaces,
and even support scientific and medical workflows. At the core, a VLM
takes in pixels and text, builds a joint representation, and then uses a
large language model to reason, plan, and respond.

\section{Application Domains}
\label{sec:application-domains}

\subsection{Assistive Technologies and Accessibility}

A natural home for VLMs is assistive technology. Here the model acts as
an extra pair of eyes and a narrator, helping users make sense of the
visual world.

Modern LVLM-based assistants can describe photographs and real-time
camera streams, answer targeted questions such as ``Where is the exit?''
or ``What does this label say?'', and read or summarize documents and
signage. Compared with traditional screen readers or OCR tools, they can
cope with cluttered scenes, unusual viewpoints, and follow-up questions
in natural language. Datasets such as VizWiz, TextVQA, and TextCaps,
which contain user-taken images with imperfect framing and text-rich
content, have been especially important for this use case.

At the same time, accessibility applications make reliability issues
very concrete. Misreading medication instructions, for example, is not a
harmless error. Deployed systems therefore often combine a VLM with:

\begin{itemize}
  \item mechanisms for expressing uncertainty or saying ``I am not
        confident'' when the image is blurry or ambiguous;
  \item conservative prompting and safety filters that avoid giving
        medical, financial, or other high-risk advice directly; and
  \item human-in-the-loop workflows, where a support agent can step in
        when the model is unsure or when the stakes are high.
\end{itemize}

\begin{didyouknow}
For many blind and low-vision users, VLM-based apps are not only a
convenience but a primary way of ``seeing'' unfamiliar places, documents,
or objects. This is why small design choices - like encouraging the model
to admit uncertainty, or to ask the user for a clearer photo - can matter
as much as another point of benchmark accuracy.
\end{didyouknow}

\subsection{Productivity, Information Access, and Content Creation}

VLMs are also increasingly woven into productivity tools that operate on
documents, slides, screenshots, and media archives.

\paragraph{Document and chart understanding.}
Given a scanned contract or a multi-page report, a VLM can identify
entities (names, dates, totals), describe layout, and answer questions
about specific passages or figures. For charts and dashboards, models
trained on ChartQA-like data can interpret axes,
legends, and visual encodings, then answer questions such as ``What is
the peak value of the orange line?'' or ``How did revenue change after
2019?'' This is particularly powerful when combined with retrieval over
enterprise document stores.

\paragraph{Multimodal search and retrieval.}
Contrastively trained dual-encoder backbones (CLIP-style) power
cross-modal search: users can type a description (``sunset over a
mountain lake'') and retrieve matching photos, or upload an image and
search for similar products or styles. Such systems support applications
in media asset management, stock photography, and e-commerce
recommendation.

\paragraph{Content generation.}
On the generative side, VLMs can write captions, alt-text for
accessibility, social-media blurbs, and slide notes from images or
screenshots. They can also serve as the text interface for
image-generation models, helping refine prompts (``Make the lighting
warmer and add a second person in the background'') or critique and
iterate on designs.

In many of these settings, architectures that reuse a strong frozen LM
with a relatively thin visual bridge (e.g., BLIP-2- or LLaVA-style) are
attractive: they can be dropped into existing text-centric products with
moderate engineering effort.

\begin{didyouknow}
In many companies, the majority of ``images'' that knowledge workers see
each day are not photos but screenshots, slides, scans, and charts.  
That means a well-tuned VLM for productivity often needs \emph{better}
OCR, layout understanding, and retrieval over PDF pages than pure
photographic skills - otherwise most of the visual information in an
enterprise simply remains invisible to the model.
\end{didyouknow}

\subsection{Interactive Agents, Tools, and User Interfaces}

Beyond passive description, VLMs can drive agents that act on digital
interfaces or coordinate multiple tools.

\paragraph{Screen understanding and UI grounding.}
High-resolution LVLMs such as Qwen-VL-style architectures can parse
screenshots and web pages, identify interactive elements (buttons,
menus, input boxes), and read on-screen text. This enables agents that
can:

\begin{itemize}
  \item automate multi-step GUI workflows (e.g., downloading reports,
        exporting data, filling forms);
  \item guide users through complex applications by answering questions
        like ``Where do I click to change the chart type?''; and
  \item align visual UI elements with underlying APIs or scripts for
        low-code automation.
\end{itemize}

\paragraph{Tool-augmented multimodal agents.}
When the VLM is allowed to call external tools (search engines,
calculators, databases), it can use vision as a front end and tools as a
back end. A single system might read numbers from a plot, run a
regression in a code interpreter, and then explain the results; or it
might inspect an inventory photo, query a database of SKUs, and suggest
replenishment actions. In these architectures, the LVLM is primarily a
planner and explainer.

\subsection{Robotics and Embodied AI}

In embodied settings, VLMs connect language instructions with
perception and control. A robot equipped with cameras and a VLM can
understand commands such as ``Pick up the red mug on the left table''
or ``Open the drawer under the sink and look for a sponge,'' grounding
referring expressions in the observed scene.

Narrated video datasets (HowTo100M, WebVid, and related corpora) supply
a bridge between visual trajectories and language describing actions.
These data can be used to pretrain video-language encoders, which then
inform downstream policies via imitation learning or high-level
planning. In many cases the LVLM is not directly producing motor torques
but rather suggesting waypoints, object choices, or decomposed steps
that a separate controller executes.

Real-world robotics introduces strict requirements: latency, robustness
to viewpoint change, calibrated uncertainty, and adherence to safety
constraints on actions. Nonetheless, multimodal foundation models are
rapidly becoming a standard component in embodied AI stacks.

\subsection{Scientific, Medical, and Industrial Applications}

Finally, there is a growing ecosystem of domain-specific VLMs tailored to
scientific, medical, and industrial imagery.

In medical imaging, models trained on radiology scans paired with
reports can answer structured questions, propose candidate findings, or draft preliminary
reports for radiologist review. The constraints here are tight:
high-resolution inputs, specialized terminology, and the need for
carefully controlled deployment where model suggestions complement - but
never replace - expert judgment.

In scientific publishing, VLMs are being explored as assistants that can
interpret plots, molecular diagrams, and microscopy images, then
summarize or cross-reference them with textual descriptions in the
paper. For industrial inspection, similar architectures are trained on
product photos and defect labels, so that the model can both flag
anomalies and explain them in everyday language (``A crack is visible on
the left side of the housing'').

These specialized systems typically reuse general-purpose VLM
architectures but rely on domain-curated datasets, expert evaluation,
and strict governance around how outputs are used.

\begin{didyouknow}
In many high-stakes domains, the hardest part is not getting a VLM to
\emph{say something sensible}, but deciding when it should say
\emph{nothing at all}. A medical or industrial assistant that can
confidently answer easy cases \emph{and} reliably admit uncertainty on
borderline ones is often far more valuable in practice than a model that
achieves a slightly higher benchmark score but never says ``I don’t
know''.
\end{didyouknow}

\section{Future Directions}
\label{sec:future-directions}

Although current VLMs are impressive, they are still far from human
visual understanding. Several directions stand out as particularly
important.

\subsection{Richer World Models and Causal Understanding}

Most existing LVLMs excel at correlational pattern matching: they know
that ``umbrellas'' often co-occur with ``rain'' and ``sidewalks'', but
they do not necessarily understand cause and effect. Future work will
need to push toward:

\begin{itemize}
  \item better models of physical dynamics and affordances (e.g.,
        predicting what will happen if an object is pushed or dropped);
  \item reasoning about human goals, intentions, and social interactions
        in images and videos;
  \item answering counterfactual questions: ``What might happen if the
        chair were removed?'' or ``How would the scene change if the
        light were off?''.
\end{itemize}

Achieving this likely requires training signals that tie vision and
actions together: interactive environments, simulation-based data, or
video corpora with explicit annotations of activities and outcomes.

\subsection{Grounded, Faithful, and Controllable Generation}

Hallucination is still a major pain point. Models can produce fluent
descriptions that mention objects or attributes not present in the
image. Emerging benchmarks such as CHAIR/CHAIRs and POPE make these
errors more visible, but solving them will require architectural and
training innovations.

Promising directions include stronger grounding mechanisms (e.g.,
generating region pointers or masks alongside text), losses that penalize
unsupported claims, and interaction modes where users can request
strictly grounded answers (for example, restricting the model to choose
from detected objects or OCR tokens). Over time, we may see VLMs
explicitly separating ``what I see'' from ``what I infer'' in their
outputs.

\subsection{Scalable, Diverse, and Responsible Data Curation}

The pretraining data used today is heavily skewed toward English-language
web photos and alt-text. To support broader and fairer applications,
future corpora will need:

\begin{itemize}
  \item wider linguistic and cultural coverage, particularly for
        underrepresented languages and regions;
  \item more domain-specific imagery (scientific, medical, industrial)
        collected under clear consent and governance frameworks;
  \item better deduplication and contamination checks to avoid training
        on test benchmarks or private material; and
  \item richer metadata and documentation about provenance, licenses,
        and potential risks.
\end{itemize}

Semi-automatic pipelines that combine heuristic filtering, model-based
screening, and human review will be essential for curating such data at
scale.

\subsection{Evaluation Beyond Static Benchmarks}

Static benchmarks are useful but inevitably incomplete. They rarely
capture how models behave in prolonged, real-world interactions.

Future evaluation may rely more on:

\begin{itemize}
  \item \textbf{Interactive studies}, where humans work with LVLMs on
        realistic tasks and provide feedback on helpfulness, clarity,
        and trustworthiness;
  \item \textbf{Task-completion benchmarks}, in which success is defined
        by achieving an end goal (e.g., completing a form, navigating a
        UI, solving a science problem) rather than getting a single
        answer right; and
  \item \textbf{Continuous auditing}, where deployed systems are
        monitored for failures, regressions, and emergent biases, and
        evaluation suites evolve accordingly.
\end{itemize}

Such setups will require closer collaboration between model developers,
domain experts, and end users.

\subsection{Personalization, Collaboration, and Human-AI Interaction}

Finally, as VLMs become everyday tools, interaction style matters almost
as much as raw accuracy. People differ in how much detail they want, how
they prefer information to be structured, and what they consider a
satisfactory explanation.

Future systems are likely to:

\begin{itemize}
  \item adapt to user-specific preferences for verbosity, formality,
        language, and visual focus (e.g., emphasizing text, layout, or
        high-level composition);
  \item support collaborative workflows where users iteratively refine
        analyses or designs with the model, and the model learns from
        these corrections over time; and
  \item offer more transparent explanations of visual decisions, for
        instance by highlighting the regions that support a particular
        answer in a way that is intuitive for non-experts.
\end{itemize}

Designing such systems will require insights from human-computer
interaction, cognitive science, and ethics, in addition to core machine
learning research.




\chapter*{Afterword}
\thispagestyle{empty}
\addcontentsline{toc}{chapter}{Afterword}

Writing about Vision-Language Models comes with a quiet paradox: by the
time you finish a book, the field it describes has already changed. New
architectures appear, training recipes shift, benchmarks are replaced,
and yesterday’s “state of the art” becomes today’s baseline. Underneath
that turbulence, though, a few questions keep returning: how to
represent visual information so that language models can genuinely
\emph{reason} with it; what it means for a caption or answer to be
faithfully grounded in an image; and how to evaluate systems that are
meant to help real people solve messy, open-ended tasks, not just pass
carefully curated tests.

This book has tried to stay close to those questions. We started from
pixels, convolutions, and patches; moved through transformers and large
language models; and examined the different ways vision and language are
brought together - contrastive spaces, cross-attention, bridge
transformers, instruction-tuned LVLMs. Along the way, we spent time with
datasets and benchmarks that quietly define what models learn, and with
applications where abstract design choices turn into tools for students,
developers, clinicians, analysts, and assistive technologies. If the
chapters have made architectures feel less opaque, datasets less
mysterious, and evaluation less mechanical, then they have done their
job.

It is normal to feel overwhelmed in a field that sits at the intersection
of vision, language, learning, HCI, and the social sciences; no one can
follow everything. The real goal is not to memorize every model, but to
develop a grounded intuition for which ideas matter and why. If this
book leaves you with a clearer sense of where we are, an honest
appreciation of what we still do not know, and a feeling that there is
room for your own ideas in what comes next, then it has served its
purpose. Thank you for taking the time to read, to question, and to
think about this evolving conversation between pixels, words, and the
people who connect them.

\thispagestyle{empty}

\bibliography{components/ref}  

\begin{titlepage}
  \thispagestyle{empty}
  \begin{tikzpicture}[remember picture,overlay]

    \fill[bgdark]
      (current page.south west) rectangle (current page.north east);

    \fill[accentA!55!black,opacity=0.65]
      ($(current page.north east)+(-0.5cm,1cm)$) --
      ($(current page.south east)+(-1cm,-1cm)$) --
      ($(current page.south west)+(-2cm,-3cm)$) --
      ($(current page.north west)+(-1cm,-1cm)$) -- cycle;

    \fill[accentB!70!black,opacity=0.75]
      ($(current page.north east)+(-0.5cm,3.5cm)$) --
      ($(current page.south east)+(-0.5cm,-2.0cm)$) --
      ($(current page.south west)+(-3cm,-3.5cm)$) --
      ($(current page.north west)+(-2cm,0.5cm)$) -- cycle;

    \fill[accentC!75!black,opacity=0.6]
      ($(current page.north east)+(-1cm,4.8cm)$) --
      ($(current page.south east)+(-2.5cm,0.5cm)$) --
      ($(current page.south west)+(-3.5cm,-0.5cm)$) --
      ($(current page.north west)+(-2cm,2.5cm)$) -- cycle;

    \draw[accentA!65, line width=0.8pt]
      ($(current page.south west)+(3.0cm,2.5cm)$) circle (2.0cm);
    \draw[accentB!55, line width=0.6pt]
      ($(current page.south west)+(3.0cm,2.5cm)$) circle (1.3cm);
    \fill[accentC!80!white,opacity=0.9]
      ($(current page.south west)+(3.0cm,2.5cm)$) circle (0.12cm);

    \fill[white,opacity=0.9]
      ($(current page.south west)+(5.2cm,3.6cm)$) circle (0.06cm);
    \fill[white,opacity=0.75]
      ($(current page.south west)+(4.0cm,4.4cm)$) circle (0.06cm);
    \fill[white,opacity=0.6]
      ($(current page.south west)+(2.1cm,3.9cm)$) circle (0.06cm);


    \node[anchor=north west]
      at ($(current page.north west)+(1.8cm,-3.2cm)$) {%
        \begin{minipage}[t]{0.5\textwidth}
          \setlength{\parskip}{0.45em}
          {\TitleFont\large\bfseries\color{textlight}
            About this book}\\[0.3em]
          {\TitleFont\small\color{white!82}
            This volume grew out of a simple question: \emph{How do we
            help a language model truly see?} It follows that question
            from the level of pixels and convolutional filters, through
            transformers and multimodal alignment mechanisms, to the
            datasets, benchmarks, and real applications that shape
            modern Vision--Language Models.

            Rather than chasing every transient leaderboard, the book
            focuses on the ideas that persist: how visual encoders
            tokenize the world, how language models act as reasoning
            engines, how the two are connected by contrastive,
            generative, and instruction-tuning objectives, and how we
            might evaluate these systems with a bit more care and
            humility. It is intended as a companion for researchers,
            practitioners, and students who want not only to use VLMs,
            but to understand the design choices behind them.}
        \end{minipage}
      };

    \node[anchor=north east]
      at ($(current page.north east)+(-1.8cm,-3.2cm)$) {%
        \begin{minipage}[t]{0.5\textwidth}
          \setlength{\parskip}{0.45em}
          {\TitleFont\large\bfseries\color{textlight}
            About the author}\\[0.3em]
          {\TitleFont\small\color{white!80}
            VO HOANG NHAT KHANG is a Ph.D.\ student in Natural Language
            Processing at MBZUAI, working at the intersection of
            language, vision, and learning from large-scale data. His
            research interests include multimodal representation
            learning, vision-language model, uncertainty quantification, and large language models.

            This book reflects a mixture of theory, practice, and
            curiosity about how models actually behave when confronted
            with messy, real-world inputs. It is written for readers who
            enjoy both equations and experiments, and who care about
            the people on the other side of the interface.}
        \end{minipage}
      };

    \draw[white!30, line width=0.4pt]
      ($(current page.west)+(1.8cm,-20.4cm)$) --
      ($(current page.east)+(-1.8cm,-20.4cm)$);

    \node[anchor=west]
      at ($(current page.west)+(1.8cm,-21.0cm)$) {%
        \begin{minipage}[t]{0.65\textwidth}
          {\scriptsize\color{white!70}
            Vision--Language Models will keep changing, but the core
            questions about representation, grounding, evaluation, and
            responsibility will remain. If this book has given you a
            clearer map of the current landscape---and a few ideas for
            where to explore next---then it has served its purpose.}
        \end{minipage}
      };

    \node[anchor=south east]
      at ($(current page.south east)+(-1.4cm,1.2cm)$) {%
        {\scriptsize\color{white!45}
          First edition \quad\textbullet\quad \the\year}
      };

  \end{tikzpicture}
\end{titlepage}
\end{document}